\PassOptionsToPackage{unicode}{hyperref}
\PassOptionsToPackage{hyphens}{url}
\documentclass[
  11pt,
  english,
]{article}
\usepackage{lmodern}
\usepackage{amssymb,amsmath}
\usepackage{ifxetex,ifluatex}
\ifnum 0\ifxetex 1\fi\ifluatex 1\fi=0 
  \usepackage[T1]{fontenc}
  \usepackage[utf8]{inputenc}
  \usepackage{textcomp} 
\else 
  \usepackage{unicode-math}
  \defaultfontfeatures{Scale=MatchLowercase}
  \defaultfontfeatures[\rmfamily]{Ligatures=TeX,Scale=1}
\fi
\IfFileExists{upquote.sty}{\usepackage{upquote}}{}
\IfFileExists{microtype.sty}{
  \usepackage[]{microtype}
  \UseMicrotypeSet[protrusion]{basicmath} 
}{}
\makeatletter
\@ifundefined{KOMAClassName}{
  \IfFileExists{parskip.sty}{%
    \usepackage{parskip}
  }{
    \setlength{\parindent}{0pt}
    \setlength{\parskip}{6pt plus 2pt minus 1pt}}
}{
  \KOMAoptions{parskip=half}}
\makeatother
\usepackage{xcolor}
\IfFileExists{xurl.sty}{\usepackage{xurl}}{} 
\IfFileExists{bookmark.sty}{\usepackage{bookmark}}{\usepackage{hyperref}}
\hypersetup{
  pdftitle={VLMProbe: An Open-source Library for VLM Probing and Interpretation},
  pdfauthor={Gexin Huang; Haisheng Fu; Yanting Yang; Chen Zhao; Ruinan Jin; Jiaqi Liu; Xiaoxiao Li},
  pdflang={en},
  hidelinks,
  pdfcreator={LaTeX via pandoc}}
\usepackage[margin=0.76in,headheight=15pt,headsep=12pt,footskip=24pt]{geometry}
\usepackage{fancyvrb}

\DefineVerbatimEnvironment{Highlighting}{Verbatim}{commandchars=\\\{\},fontsize=\footnotesize}
\newenvironment{Shaded}{}{}

\newcommand{\ExtensionTok}[1]{#1}

\newcommand{\ImportTok}[1]{#1}

\newcommand{\KeywordTok}[1]{\textcolor[rgb]{0.00,0.44,0.13}{\textbf{#1}}}
\newcommand{\NormalTok}[1]{#1}
\newcommand{\OperatorTok}[1]{\textcolor[rgb]{0.40,0.40,0.40}{#1}}

\newcommand{\StringTok}[1]{\textcolor[rgb]{0.25,0.44,0.63}{#1}}

\definecolor{probeNavy}{HTML}{10213B}
\definecolor{probeBlue}{HTML}{2456CE}
\definecolor{probePale}{HTML}{F5F8FF}
\definecolor{probeBorder}{HTML}{D7E2FA}
\definecolor{codeGray}{HTML}{F2F3F5}
\usepackage{titlesec}
\titleformat{\section}{\Large\bfseries\color{probeNavy}}{\thesection}{0.7em}{}
\titleformat{\subsection}{\large\bfseries\color{probeNavy}}{\thesubsection}{0.7em}{}
\titlespacing*{\section}{0pt}{15pt plus 3pt minus 2pt}{6pt}
\titlespacing*{\subsection}{0pt}{12pt plus 2pt minus 2pt}{5pt}
\usepackage{caption}
\usepackage{tabularx,booktabs,longtable}

\usepackage{etoolbox}
\usepackage{needspace}
\newcommand{\sectionreserve}{\Needspace{5\baselineskip}}
\newcommand{\subsectionreserve}{\Needspace{4\baselineskip}}
\preto{\section}{\sectionreserve}
\preto{\subsection}{\subsectionreserve}
\AtBeginEnvironment{table}{\renewcommand{\arraystretch}{1.15}\setlength{\tabcolsep}{5pt}}
\usepackage[section]{placeins}
\usepackage{fancyhdr}
\fancypagestyle{plain}{\fancyhf{}\fancyfoot[R]{\sffamily\footnotesize\color{probeNavy!65}\thepage}}
\usepackage[most]{tcolorbox}
\tcolorboxenvironment{Shaded}{colback=codeGray,colframe=codeGray,boxrule=0pt,
  arc=2pt,breakable,left=7pt,right=7pt,top=6pt,bottom=6pt}
\renewenvironment{abstract}{%
  \begin{tcolorbox}[enhanced,colback=probePale,colframe=probeBorder,
    boxrule=0.45pt,sharp corners,borderline west={2.5pt}{0pt}{probeBlue},
    left=12pt,right=12pt,top=10pt,bottom=10pt,
    before skip=10pt,after skip=16pt]
  {\sffamily\bfseries\color{probeBlue}\small ABSTRACT}\par\vspace{4pt}
  \normalfont\fontsize{10}{12.8}\selectfont
  \setlength{\parskip}{0pt}%
}{\end{tcolorbox}}
\hypersetup{colorlinks=true,linkcolor=probeNavy,citecolor=probeNavy,urlcolor=probeNavy}
\AtBeginDocument{\AtEveryBibitem{\ifentrytype{online}{\toggletrue{bbx:url}}{}}}
\AtBeginDocument{\AtBeginBibliography{%
  \raggedright
  \setlength{\bibitemsep}{0pt}\setlength{\bibparsep}{0pt}\setlength{\parskip}{0pt}
  \setcounter{biburlnumpenalty}{100}%
  \setcounter{biburllcpenalty}{100}%
  \setcounter{biburlucpenalty}{100}}}
\ifxetex
  \usepackage{polyglossia}
  \setmainlanguage[]{english}
\else
  \usepackage[shorthands=off,main=english]{babel}
\fi
\usepackage[backend=biber,style=authoryear,maxcitenames=2,maxbibnames=99,doi=true,isbn=false,url=false]{biblatex}
\defbibheading{reportrefs}[References]{\section*{#1}\markboth{References and data resources}{References and data resources}}

\title{VLMProbe: An Open-source Library for VLM Probing and Interpretation}
\author{Gexin Huang \and Haisheng Fu \and Yanting Yang \and Chen Zhao \and Ruinan Jin \and Jiaqi Liu \and Xiaoxiao Li}
\date{}

\begin{document}
\thispagestyle{plain}
\begingroup
\setlength{\parindent}{0pt}
\vspace*{-8pt}
{\sffamily\bfseries\color{probeNavy!75}\fontsize{8.5}{11}\selectfont
  \MakeUppercase{Technical Report}\par}
\vspace{5pt}
{\sffamily\bfseries\color{probeBlue}\fontsize{30}{35}\selectfont
VLMProbe:\par}
\vspace{4pt}
{\sffamily\bfseries\color{probeNavy}\fontsize{19}{23}\selectfont
An Open-source Library for VLM Probing and Interpretation\par}
\vspace{7pt}
{\color{probeNavy!75}\fontsize{13}{16}\selectfont Exploring Model Internals
Beyond the Answer\par}
\vspace{10pt}
{\fontsize{10}{14}\selectfont
  \mbox{Gexin Huang\textsuperscript{1,2}},\quad \mbox{Haisheng Fu\textsuperscript{1,2}},\quad \mbox{Yanting Yang\textsuperscript{1,2}},\quad
  \mbox{Chen Zhao\textsuperscript{1,2}},\quad
  \mbox{Ruinan Jin\textsuperscript{1,2}},\quad \mbox{Jiaqi Liu\textsuperscript{1,2}},\quad
  \mbox{Xiaoxiao Li\textsuperscript{1,2}}\par}
\vspace{3pt}
{\sffamily\color{probeNavy}\fontsize{9.5}{12}\selectfont
  \textsuperscript{1}The University of British
Columbia\hspace{12pt}\textsuperscript{2}Vector Institute\par}
\vspace{5pt}
\endgroup

\begin{abstract}
Understanding how vision-language models use visual information requires
connecting internal representations to model behavior.
We present vlmprobe, an open-source library that integrates lens readouts,
attention attribution, causal interventions, and sparse autoencoder (SAE)
analysis within a common model interface. Shared adapters, calibration
and checkpoint utilities, evaluation pipelines, and visualizations support
systematic analysis across architectures. Using this framework, we compare
methods across vision-language, language, and vision models through
matched case studies and controlled evaluations of pretrained sparse
dictionaries. The comparisons show that representation readouts and
spatial explanations depend on the analysis objective and observation site,
and that local attribution estimates can disagree with measured intervention
effects. SAE evaluations further expose distinct trade-offs in reconstruction
fidelity, concept readout, and spatial localization. Together, these results
support combining multiple forms of evidence to evaluate an interpretation.
vlmprobe provides a reusable basis for selecting methods, comparing their
conclusions, and testing hypotheses about model behavior.
\end{abstract}

\hypertarget{introduction}{%
\section{Introduction}\label{introduction}}

Vision-language models answer questions, describe images, and make
predictions from combined visual and textual inputs. Their responses,
however, do not reveal how visual information is encoded, transported, or
used to select an answer. Understanding these processes is important for
diagnosing failures and assessing whether a model's behavior is grounded
in the input. A useful analysis must connect information in internal
representations with spatial evidence and the effects of changing model
computation.

Lenses such as Tuned Lens and Patchscopes make hidden representations
readable \autocite{tuned,patchscopes}; attention rollout and relevance
propagation characterize token interactions and spatial evidence
\autocite{rollout,generic}. Activation/path patching and attention knockout
examine how changes to activations or information pathways affect behavior
\autocite{ioi,knockout}. SAEs, including ReLU and TopK dictionaries,
provide learned sparse coordinates for feature analysis
\autocite{sae_relu,sae_topk}. Prisma supports vision and video analysis,
and SAEBench provides multidimensional SAE evaluation
\autocite{prisma,saebench}. These advances offer complementary evidence,
but combining them across architectures requires consistent input
conventions, activation sites, calibration or dictionary bindings, and
target scores. Similar output formats do not make the underlying
measurements equivalent.

We present vlmprobe with two objectives: to integrate these complementary
methods into a reusable analysis workflow, and to support systematic
comparisons of the evidence they provide. Users bind a loaded model once,
select compatible methods and observation sites, and retain the resulting
tensors together with their analysis settings. Model adapters preserve
native activation sites and image--text token layouts, while the lens,
attention, causal and SAE interfaces organize method execution.
Calibration and checkpoint-loading utilities, shared evaluation pipelines,
and visualizations support the full workflow. This design keeps
model-specific requirements explicit while reducing the repeated
infrastructure needed to analyze different architectures.

Using this framework, we conduct matched VLM case studies and complementary
language and vision studies. Within compatible comparisons, we hold source
inputs and semantic targets fixed and report the controls needed to
interpret each result. Evaluations of pretrained sparse dictionaries
additionally examine reconstruction, concept readout, spatial localization,
and decoder answer evidence. The report makes three contributions:

\begin{itemize}
\setlength{\itemsep}{3pt}
\setlength{\parskip}{0pt}
\setlength{\parsep}{0pt}
\item \textbf{An integrated analysis library.} A common model interface
connects four method families with calibration, checkpoint handling,
evaluation, and visualization across supported vision-language, language,
and vision architectures.
\item \textbf{A comparative evaluation workflow.} Matched cases and
controlled pretrained-SAE studies make inputs, analysis units, targets,
and controls explicit, supporting comparisons among compatible methods.
\item \textbf{Evidence about method behavior.} Readouts and spatial
explanations depend on objectives and observation sites; local attribution
estimates can disagree with measured intervention effects; and SAE
reconstruction, concept readout and spatial utility exhibit distinct
trade-offs. These findings motivate combining measurements and validating
explanatory hypotheses against model behavior.
\end{itemize}

\hypertarget{library-and-evaluation-protocol}{%
\section{Library and evaluation
protocol}\label{library-and-evaluation-protocol}}

\hypertarget{model-adapters-and-method-configuration}{%
\subsection{Model adapters and method
configuration}\label{model-adapters-and-method-configuration}}

\begin{figure}[!htbp]
\centering
\includegraphics[width=\linewidth]{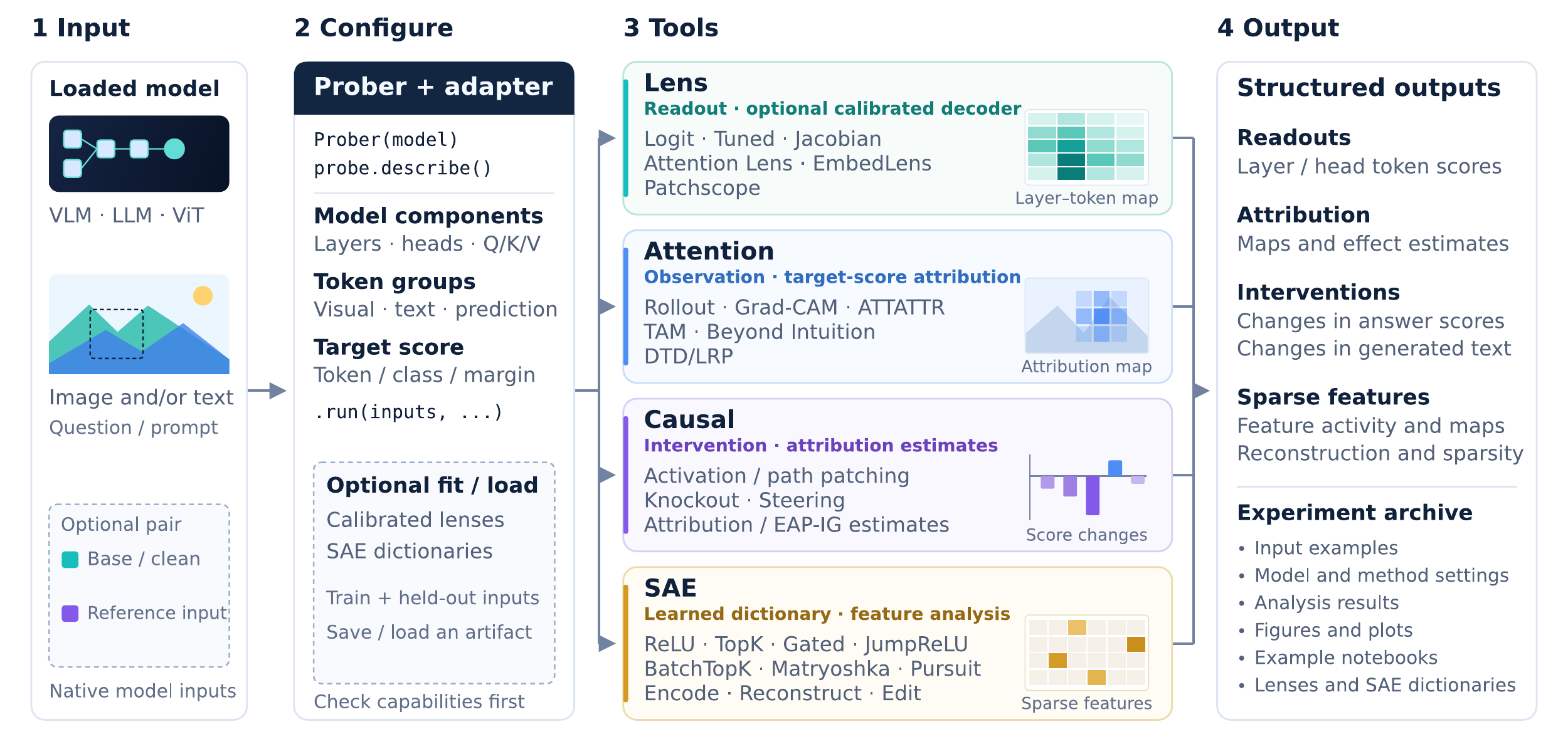}
\caption{The vlmprobe workflow. Native inputs and explicit analysis settings feed lens, attention, causal and sparse-feature methods; their outputs support token readouts, spatial explanations, sparse feature analysis and measured interventions. Miniature plots illustrate output types rather than experimental results.}
\label{fig:protocol}
\end{figure}

Figure \ref{fig:protocol} summarizes the library workflow. An input
contains the native model tensors and their text/image coordinates.
Configuration specifies the method, activation sites, positions, and
target score; paired inputs and calibrated artifacts are supplied where
required. Outputs retain these choices alongside decoded tokens,
attribution maps, or intervention effects. This shared contract allows
methods to reuse an input preparation step and a visualization without
assuming identical internal architectures.

The following example reads a representation, explains an answer with
attention gradients, and compares exact and first-order residual
replacement:

\begin{Shaded}
\begin{Highlighting}[]
\ImportTok{from}\NormalTok{ vlmprobe }\ImportTok{import}\NormalTok{ Prober}
\ImportTok{from}\NormalTok{ vlmprobe.metrics }\ImportTok{import}\NormalTok{ SequenceLogProb}

\NormalTok{probe }\OperatorTok{=}\NormalTok{ Prober(model, processor)}
\NormalTok{metric }\OperatorTok{=}\NormalTok{ SequenceLogProb(answer\_mask)}

\NormalTok{readout }\OperatorTok{=}\NormalTok{ probe.lens.logit(}
\NormalTok{    layers}\OperatorTok{=}\NormalTok{layers, tokens}\OperatorTok{=}\NormalTok{prediction\_positions}
\NormalTok{).run(inputs)}
\NormalTok{attention }\OperatorTok{=}\NormalTok{ probe.attention.grad\_cam(}
\NormalTok{    layers}\OperatorTok{=}\NormalTok{attention\_layers, queries}\OperatorTok{=}\NormalTok{prediction\_mask, keys}\OperatorTok{=}\StringTok{"visual"}
\NormalTok{).run(inputs, metric}\OperatorTok{=}\NormalTok{metric)}
\NormalTok{exact }\OperatorTok{=}\NormalTok{ probe.causal.patch(layers}\OperatorTok{=}\NormalTok{layers, tokens}\OperatorTok{=}\StringTok{"visual"}\NormalTok{).run(}
\NormalTok{    inputs, source}\OperatorTok{=}\NormalTok{reference, metric}\OperatorTok{=}\NormalTok{metric}
\NormalTok{)}
\NormalTok{estimate }\OperatorTok{=}\NormalTok{ probe.causal.attribute(layers}\OperatorTok{=}\NormalTok{layers, tokens}\OperatorTok{=}\StringTok{"visual"}\NormalTok{).run(}
\NormalTok{    inputs, source}\OperatorTok{=}\NormalTok{reference, metric}\OperatorTok{=}\NormalTok{metric}
\NormalTok{)}
\end{Highlighting}
\end{Shaded}

Here \texttt{inputs} includes the fixed answer for teacher-forced
scoring; \texttt{answer\_mask} selects its token pieces, and
\texttt{reference} supplies aligned source activations. The method
implementations share the adapter interface, while
\texttt{probe.describe()} exposes capabilities relevant to choosing
sites and tools. The executed notebooks supply the native preprocessing
and position-selection steps omitted from this compact example.

\hypertarget{models-inputs-and-comparison-groups}{%
\subsection{Models, inputs, and comparison
groups}\label{models-inputs-and-comparison-groups}}

Table \ref{tab:models} lists the five pretrained VLMs used for the
shared case studies. They vary in decoder architecture, visual-token
compression, and model size. Table \ref{tab:data} defines the eight
fixed inputs, chosen before inference. GQA supplies a common attribute
question and target boxes \autocite{gqa}; MMVP contrasts visually
similar images \autocite{mmvp}; VSR tests spatial propositions
\autocite{vsr}; and VisOnlyQA supplies diagrams and charts
\autocite{visonly}.

\begin{table}[htbp]
\centering\small
\caption{Models in the shared-case comparison. Sizes follow public checkpoint names. Native adapters retain each model's decoder sites and visual-token grid.}
\label{tab:models}
\begin{tabularx}{\linewidth}{l c X}
\toprule Model & Image grid & Architectural feature \\
\midrule
SmolVLM-500M-Instruct & $8\times8$ & Compressed image tokens with a Llama decoder \\
InternVL3-1B-hf & $16\times16$ & InternVL vision path with a Qwen2 decoder \\
Qwen3-VL-2B-Instruct & $14\times14$ & Normalized Q/K and DeepStack visual additions \\
Pixtral-12B & $16\times16$ & Pixtral patch encoder with a Mistral decoder \\
LLaVA-1.5-7B & $24\times24$ & CLIP visual features projected into a Llama decoder \\
\bottomrule
\end{tabularx}
\end{table}

\begin{table}[htbp]
\centering\small
\caption{Predetermined shared inputs. Each source contributes two cases; image captions from a separate COCO split calibrate the VLM Tuned Lenses.}
\label{tab:data}
\begin{tabularx}{\linewidth}{l l X}
\toprule Source & Released identifiers & Analysis question \\
\midrule
GQA & 07302654 / 19225230 & Is the shirt black or yellow? \\
MMVP & Images 1 / 2 & Are the butterfly's wings open or closed? \\
VSR zero-shot & Test rows 0 / 1 & Car behind suitcase; cat under backpack \\
VisOnlyQA real & Row 0 of two splits & Connected triangle; chart-value extraction \\
\bottomrule
\end{tabularx}
\end{table}

For these shared cases, the same photograph and semantic question are supplied to each VLM.
Native chat templates and tokenizers remain model-specific. Images are
padded to a square before preprocessing; displayed maps are transformed
back to the original photograph, including its aspect ratio and
bounding-box coordinates. Comparisons within a model hold token
positions fixed. Comparisons across models preserve semantic targets,
since native subwords and visual grids differ. The generation-steering
comparison separately uses COCO validation image 310196 across three
VLMs, retaining the original image geometry (Figure \ref{fig:vsv}).

For a fixed answer \(a=(a_1,\ldots,a_m)\), the shared score is

\[
S(x,a)=\sum_{j=1}^{m}\log P(a_j\mid x,a_{<j}),\qquad
\Delta S=S(x\text{ with intervention},a)-S(x,a).
\]

The score includes every native answer piece and excludes EOS. A
negative \(\Delta S\) means reduced support for that answer. Native
generation is shown separately so that an explanation of a forced answer
is distinguishable from an explanation of the model's actual response.
Different tokenizations make absolute scores unsuitable as a cross-model
calibration ranking.

The comparison groups also depend on the method's analysis unit. Table
\ref{tab:groups} outlines these groups. GPT-2 and ViT studies retain the
original text and vision settings for tools whose calibrated artifacts
or audited computational graphs are not yet available on all five VLMs.
These studies complement the shared cases rather than forming one
universal method leaderboard. Training, checkpoint revisions, literal
prompts, and validation checks are provided in Appendix
\ref{sec:details}.

The SAE studies extend this design with frozen language and spatial
dictionaries, independent calibration and test data, and reconstruction and
spatial-shuffle controls. Visual SAE comparisons retain source-native
transforms and patch coordinates across CLIP, DINO, SigLIP and Qwen variants.
These coordinates are distinct from the square-padded shared-case displays.

The comparisons cover the paper-linked analysis and intervention
methods. Attention profiling, head-logit inspection, temperature
scaling, and reweighting are also available as configurable utilities;
they are not counted as separate paper-reproduction studies.

\begin{table}[htbp]
\centering\small
\caption{Comparison groups and the evidence each provides. Groups organize matched inputs and compatible measures; distinct analysis units remain separately interpreted.}
\label{tab:groups}
\begin{tabularx}{\linewidth}{l X X}
\toprule Family & Methods demonstrated & Matched comparison \\
\midrule
Lens & Logit, Tuned, Jacobian, Attention Lens; EmbedLens readout; Patchscope & Residual/head decoding on IOI; visual-token and transfer readouts on GQA \\
Attention & Rollout, Grad-CAM, ATTATTR, TAM, Beyond-H/C, generic relevance & Seven attribution methods on GQA across five VLMs; native ViT maps and deletion tests \\
Causal & Activation/attribution patching, knockout, path patching, EAP/EAP-IG, VSV steering & Exact versus approximate replacements; attention knockout; controlled sender paths; edge retention; generation changes across three VLMs \\
SAE & ReLU, TopK, Gated, JumpReLU, BatchTopK, Matryoshka; RA-SAE and SAE++ readouts & Frozen Pythia dictionaries; source-native spatial readouts across nine vision/VLM backbones; image--question attribution and feature removal in five decoder configurations \\
\bottomrule
\end{tabularx}
\end{table}

\FloatBarrier

\hypertarget{lens-methods}{%
\section{Lens methods}\label{lens-methods}}

\hypertarget{residual-and-head-readouts-expose-different-levels-of-a-decision}{%
\subsection{Residual and head readouts expose different levels of a
decision}\label{residual-and-head-readouts-expose-different-levels-of-a-decision}}

Logit Lens applies the model's final normalization and unembedding to an
intermediate residual \autocite{logit}. Tuned Lens first applies a
fitted affine translator, aligning that representation with the final
predictive basis \autocite{tuned}. At layer \(\ell\) and position \(t\),

\[
p^{\mathrm{logit}}_{\ell,t}
=\mathrm{softmax}(W_U\,\mathrm{Norm}(h_{\ell,t})),\quad
p^{\mathrm{tuned}}_{\ell,t}
=\mathrm{softmax}(W_U\,\mathrm{Norm}(A_\ell h_{\ell,t}+b_\ell)).
\]

\begin{figure}[!htbp]
\centering
\includegraphics[width=\linewidth,height=.74\textheight,keepaspectratio]{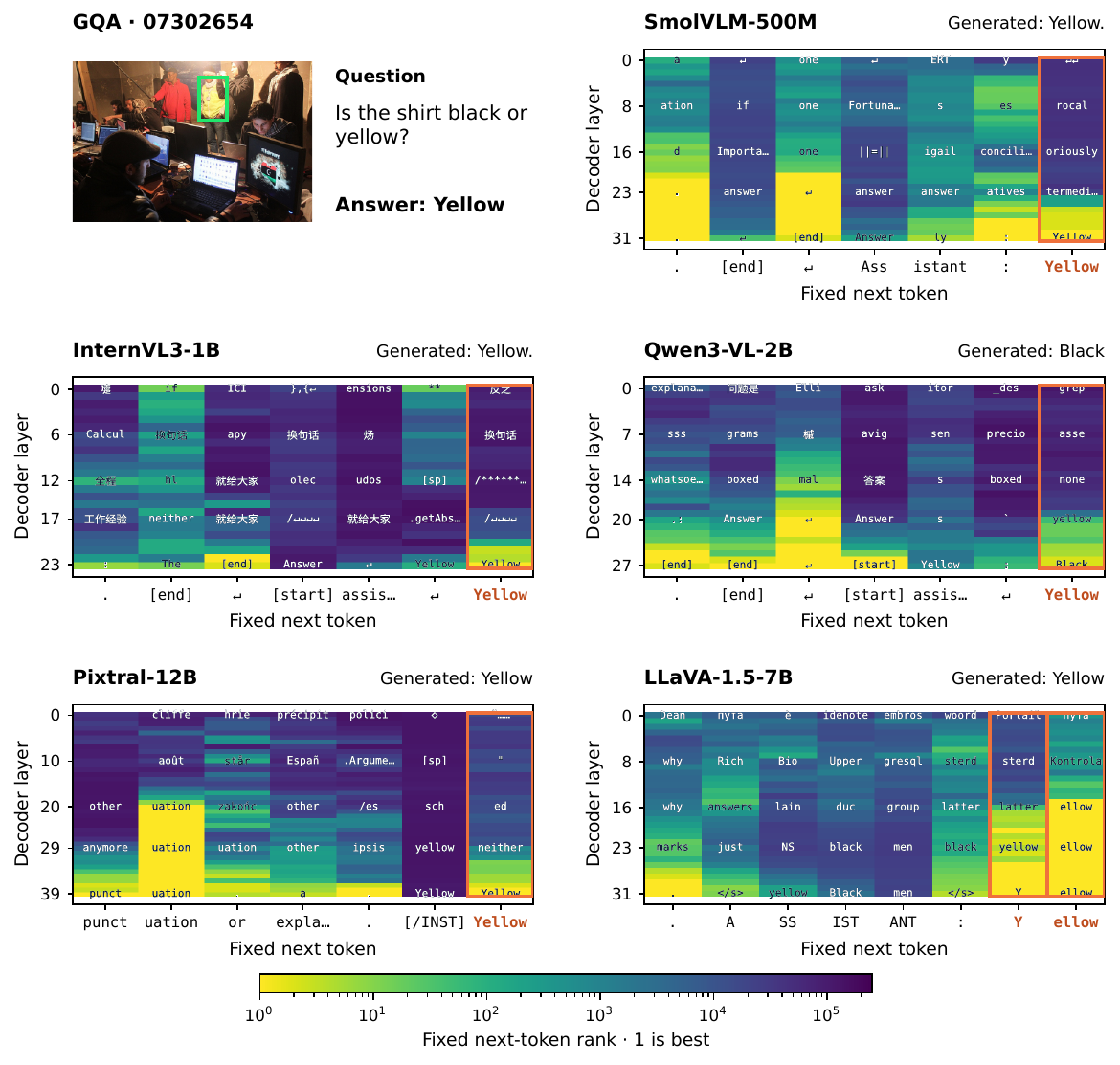}
\caption{Layer-wise vocabulary readouts for one GQA decision across five VLMs. The input panel shows the question, answer and annotated shirt. All decoder layers retain their measured colors; selected rows show shortened top-1 fragments after native tokenizer decoding. Top-1 search includes all vocabulary languages, regardless of the English input. Undecodable or unrenderable fragments remain unlabeled. Yellow indicates a better rank for the fixed next token, using one shared logarithmic scale. Orange outlines identify answer predictions: Qwen3-VL generates Black while the other models generate Yellow. LLaVA predicts Yellow in two pieces, with the second conditioned on the first. Horizontal labels shorten chat markers to [start]/[end], omit whitespace markers and display line breaks as arrows; columns follow each model's native template.}
\label{fig:logit-vlm}
\end{figure}

Figure \ref{fig:logit-vlm} shows the same yellow-shirt GQA question
across five VLMs. Rows are decoder layers; columns are selected native
prompt and answer positions. Cell text gives the most likely vocabulary
fragment after native tokenizer decoding, while color gives the rank of
the fixed next token. The top-1 search covers the full vocabulary
without restricting the language, so English inputs can yield
non-English fragments. These are intermediate vocabulary projections,
not generated answers or evidence of a particular internal reasoning
language. Most context columns are native chat-template suffixes, rather
than aligned question words across models. The outlined answer columns
expose a concrete disagreement: Qwen3-VL ranks Yellow second at the
final readout and generates Black; the other models generate Yellow.
Early decoded fragments describe an evolving vocabulary projection, but
cannot by themselves establish that the shirt was recognized or locate a
causal loss of visual information.

Two other lenses change the readout question. Jacobian Lens uses a
calibrated causal transport to the final residual basis to expose
verbalizable content \autocite{jacobian}. Attention Lens learns a
vocabulary decoder for an individual head's projected output
\autocite{attentionlens}. These are different units and objectives: head
readouts resolve components combined in a residual, whereas Jacobian
readouts need not represent the next-token predictive distribution.

\begin{figure}[!htbp]
\centering
\includegraphics[width=\linewidth]{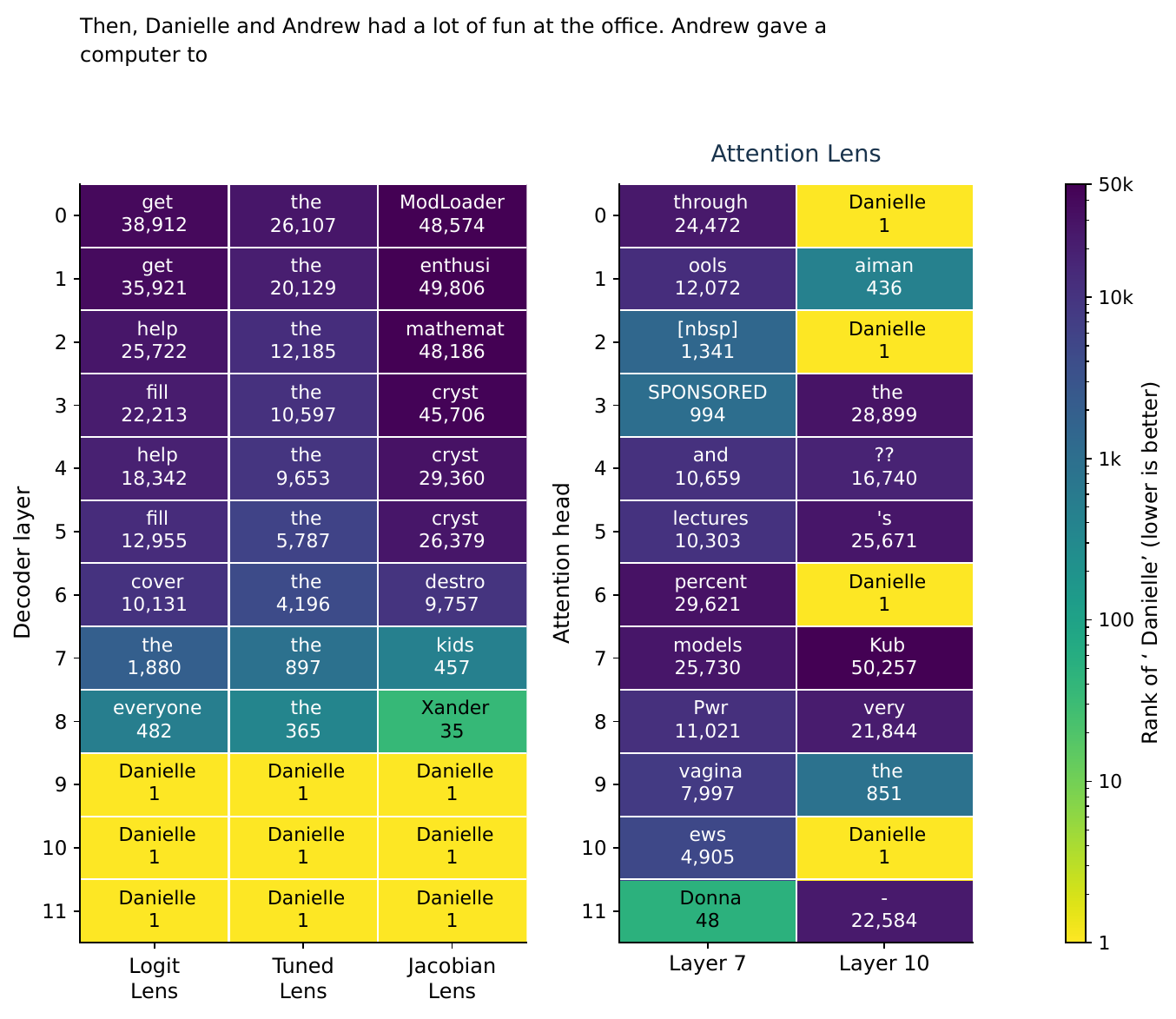}
\caption{Residual and head readouts of the same IOI decision in GPT-2 small. Left: Logit, Tuned, and Jacobian Lens track Danielle at the final prediction position. Right: Attention Lens distinguishes heads within layers 7 and 10. Top-token labels and target ranks reveal agreement at late residual sites alongside substantial head-level variation.}
\label{fig:text-lenses}
\end{figure}

We examine all four lenses on a GPT-2 IOI prompt ending ``Andrew gave a
computer to,'' whose correct next name is Danielle. The native
next-token probabilities are 0.842 for Danielle and 0.003 for Andrew.
Figure \ref{fig:text-lenses} holds the final prediction position and
target token fixed. Logit and Tuned Lens first select Danielle at layer
9. At layer 7, calibration improves her rank from 1,880 to 897, while
Jacobian Lens ranks her 457th and instead decodes ``kids.'' At layer 8,
Jacobian Lens places Danielle at rank 35, compared with 482 for Logit
Lens and 365 for Tuned Lens. All three select Danielle from layer 9.
Attention Lens reveals variation within layer 10: heads 0, 2, 6, and 10
decode Danielle, while the other eight do not.

This distinction is useful when selecting subsequent interventions. A
residual lens traces how a prediction emerges; a head lens nominates
finer components for testing. A decoded name alone does not establish
that a head is necessary or identify its functional role. Comparing
calibrated head and residual probabilities as if they were additive
causal contributions would obscure this difference.

Tuned Lens is calibrated against the final next-token distribution,
whereas Jacobian Lens estimates downstream causal transport and
Attention Lens decodes individual head messages. Those objectives do not
define one shared fidelity metric. We therefore compare their visible
readouts on matched inputs, rather than treating small-case distribution
divergences as a method ranking. The separately fitted VLM translators
and their independent validation settings are retained in Appendix
\ref{sec:details}.

\hypertarget{lexical-neighbors-reveal-associations-rather-than-object-labels}{%
\subsection{Lexical neighbors reveal associations rather than object
labels}\label{lexical-neighbors-reveal-associations-rather-than-object-labels}}

\begin{figure}[!htbp]
\centering
\includegraphics[width=\linewidth]{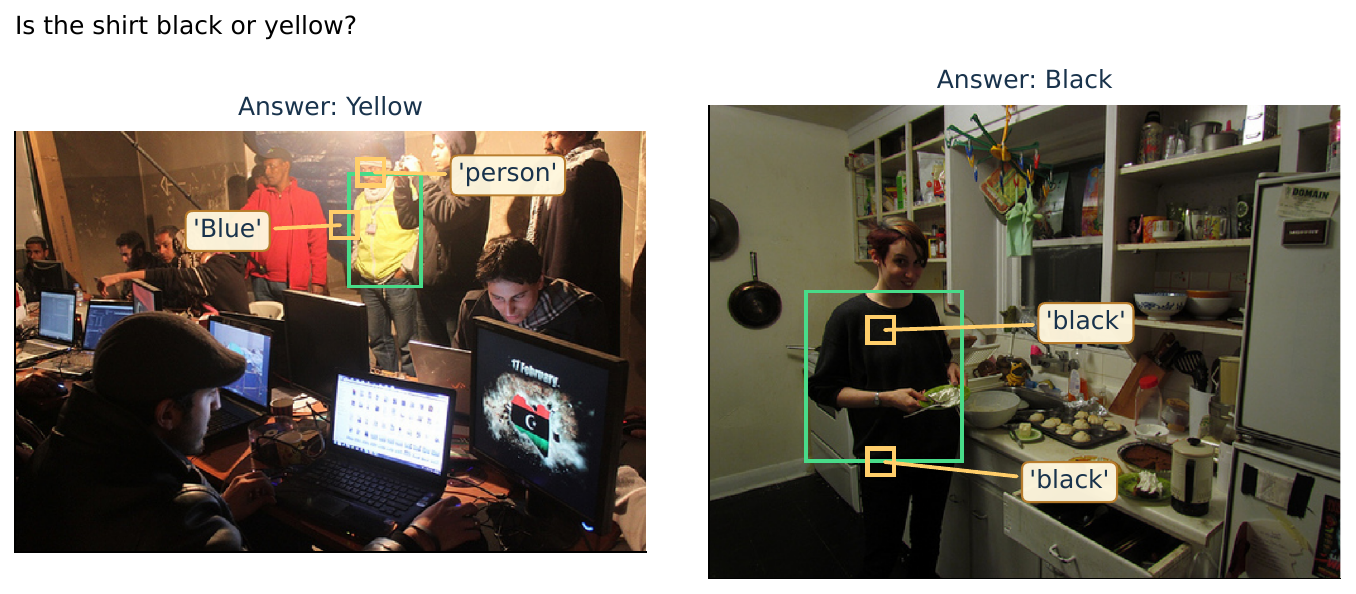}
\caption{EmbedLens attaches lexical readouts to visual patches in LLaVA-1.5-7B. Amber patch boxes and leader lines connect the nearest vocabulary fragments to their projected visual tokens; green boxes mark the annotated shirt. These are embedding-space neighbors rather than independently detected object labels. The black-shirt example associates with “black,” while yellow-shirt patches also yield unrelated fragments such as “Blue.”}
\label{fig:embed}
\end{figure}

The implemented EmbedLens readout examines projected visual-token
embeddings before the decoder \autocite{embed}. For visual token \(v_i\)
and vocabulary embedding \(e_w\), it reports neighbors ranked by
\(\cos(v_i,e_w)=v_i^\top e_w/(\lVert v_i\rVert\lVert e_w\rVert)\), with
token norms available in the saved outputs. Figure \ref{fig:embed}
applies this readout to both GQA shirt images. The image annotations
attach decoded lexical fragments to the selected visual patches.

A black-shirt patch associates with ``black,'' while a yellow-shirt
patch instead associates with ``Blue.'' Successful and ambiguous
neighbors therefore coexist under the same readout. This tool exposes
associations cheaply and spatially, but neither cosine similarity nor a
large norm is a calibrated object prediction. The library currently
implements this semantic readout; the paper's complete sink/dead/alive
classification and pruning analysis are outside the demonstrated
implementation.

\hypertarget{transfer-readouts-depend-on-the-decoding-prompt}{%
\subsection{Transfer readouts depend on the decoding
prompt}\label{transfer-readouts-depend-on-the-decoding-prompt}}

\begin{figure}[!htbp]
\centering
\includegraphics[width=\linewidth]{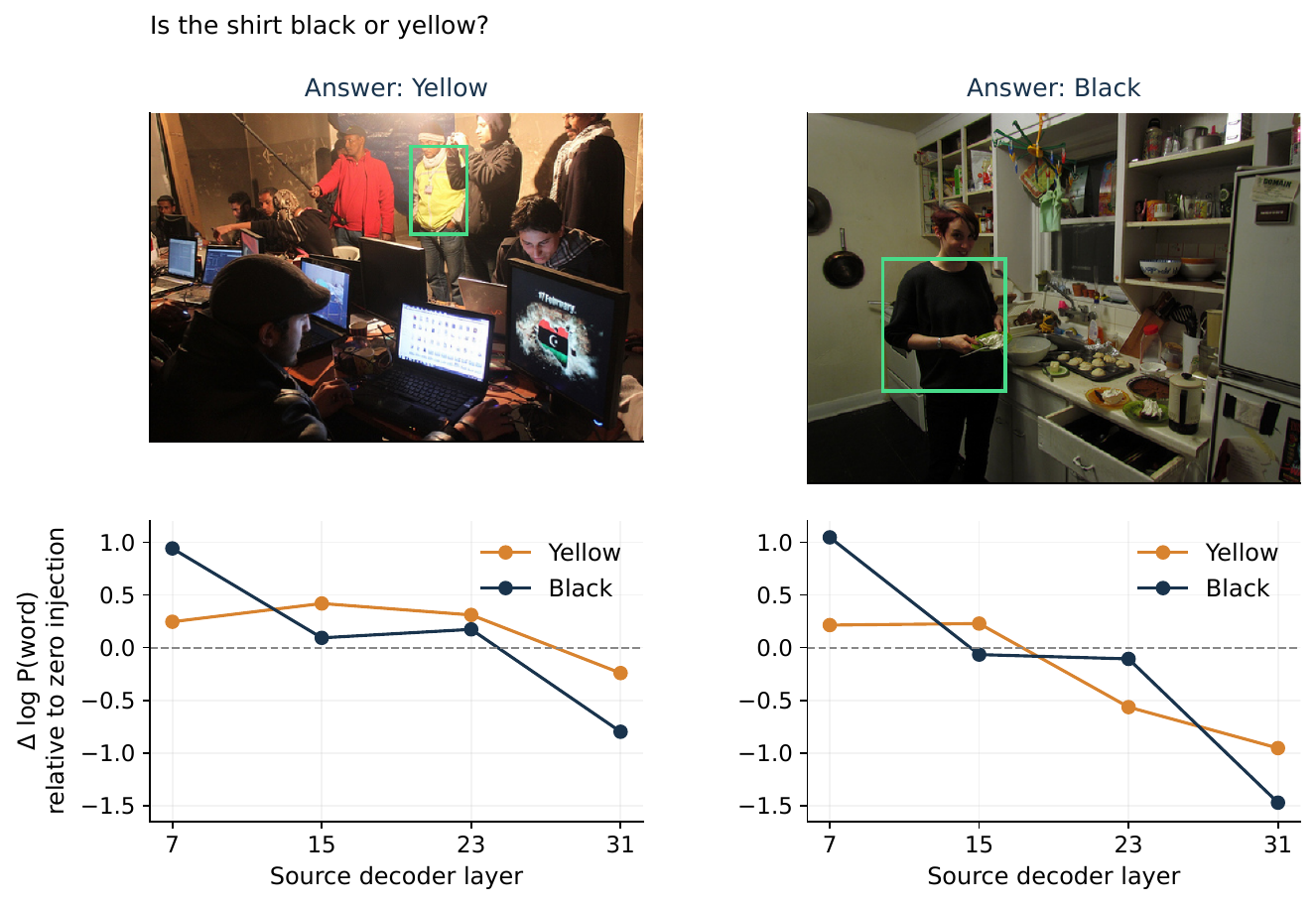}
\caption{Patchscope reads a source residual through a new text context. In LLaVA-1.5-7B, the source answer-position residual replaces the placeholder after target layer 0 in a four-line color-copy prompt, \texttt{red -> red}, \texttt{blue -> blue}, \texttt{green -> green}, then \texttt{x ->}. Curves show changes in complete-word log probability relative to a same-site zero-vector injection; positive values indicate greater support for the plotted color. This configuration changes word likelihoods but fails to recover the shirt color as its top continuation.}
\label{fig:patchscope}
\end{figure}

Patchscopes moves a source representation into a specified target
computation to make its content readable \autocite{patchscopes}. We use
an identity mapping in LLaVA: an answer-position residual from a GQA
source replaces a placeholder residual in a color-copy prompt. Figure
\ref{fig:patchscope} shows the source images and word-likelihood changes
under a fixed target decoding context, relative to a zero-vector
injection at that same target site.

The yellow source at layer 15 increases the complete-word Yellow log
probability by 0.419 relative to zero injection, yet its top prediction
is ``is.'' Both sources produce ``is'' at layers 7, 15, and 23, and
``-\textgreater{}'' at layer 31. This configuration fails to verbalize
the color even though the source computation answers correctly. Unlike
direct projection, a transfer readout can use a new decoding context,
but its source site, mapping, and target prompt need validation.
Likelihood changes alone do not establish successful information
transfer.

\FloatBarrier

\section{Attention explanations}
\label{attention-explanations}

\subsection{Propagation rules change the map for a fixed answer}
\label{propagation-rules-change-the-map-for-a-fixed-answer}

\begin{figure}[!htbp]
\centering
\includegraphics[width=\linewidth]{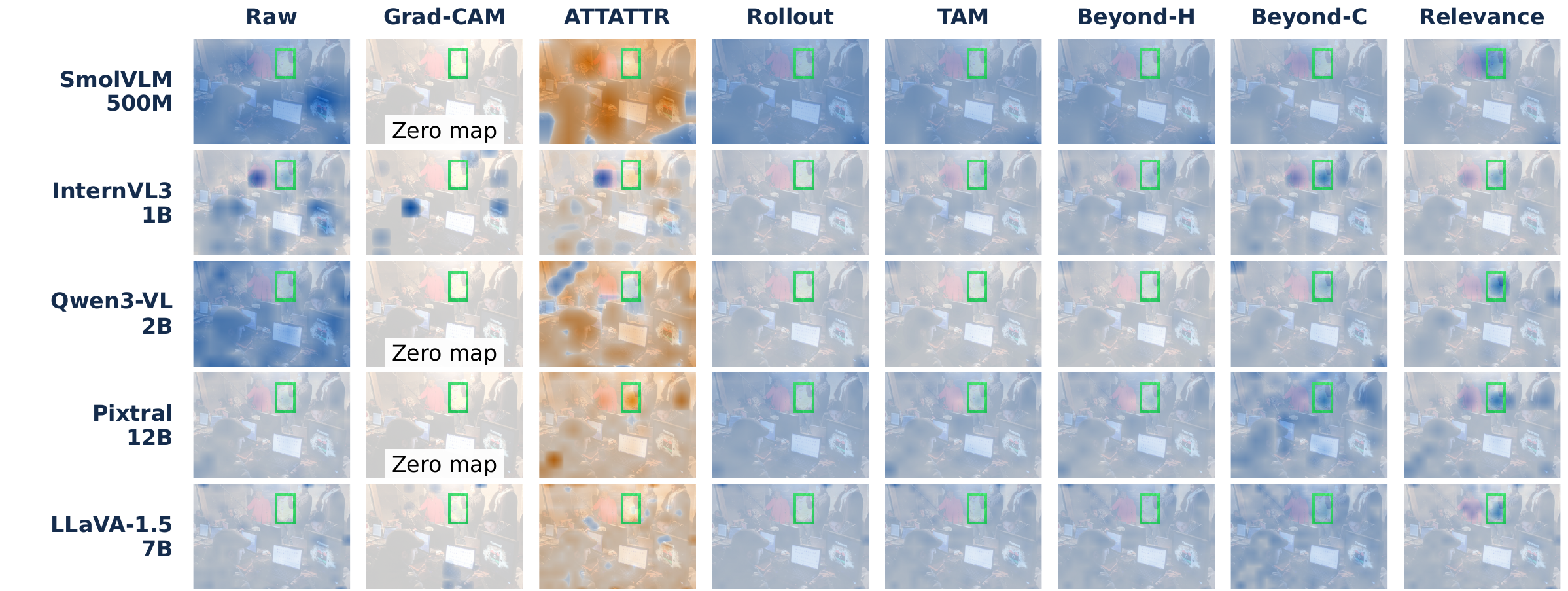}
\caption{Attention explanations assign different spatial emphasis to the same yellow-shirt input. Rows compare five VLMs; columns compare raw attention and seven explanation methods. Target-conditioned methods explain Yellow. Blue indicates positive values and orange negative ATTATTR values; green boxes mark the shirt. Each map is independently normalized, so intensity compares positions within a map rather than score magnitude across maps. Original photographs are displayed at 20\% opacity over opaque heatmaps. Raw attention, Grad-CAM, and ATTATTR inspect the last decoder attention layer; propagation methods span the decoder attention stack. ``Zero map'' denotes a measured null attribution.}
\label{fig:attention-comparison}
\end{figure}

Figure \ref{fig:attention-comparison} compares raw attention and seven
explanation methods on the same yellow-shirt GQA input across five VLMs.
Reading across a row isolates the method; reading down a column isolates
the model. Target-conditioned maps explain Yellow, including Qwen3-VL,
whose native answer is Black.

The seven methods use different attribution and propagation rules:
Rollout \autocite{rollout}, attention Grad-CAM \autocite{gradcam},
ATTATTR \autocite{attattr}, TAM \autocite{tam}, Beyond Intuition-H/C
\autocite{beyond}, and generic relevance \autocite{generic}. Raw
attention provides an additional reference. Rollout follows
residual-augmented attention connections; Grad-CAM weights attention
with a local target gradient; ATTATTR integrates sensitivity along an
attention-probability path. The remaining methods use different
gradient, value, or relevance-propagation rules. A common output shape
makes their maps comparable visually, but does not make their scores
equivalent.

The spatial differences are substantial even when the input and target
remain fixed. Generic relevance concentrates more strongly within the
annotated shirt region than rollout in each displayed model, whereas
rollout spreads much of its map outside that object. A sharper map still
does not establish a more faithful explanation: the box fractions
describe this one input and each map has its own scale.

ATTATTR adds a different distinction. Its signed results are predominantly negative in
four models and more mixed in InternVL; discarding the sign would
conceal that difference from positive-only maps.

\hypertarget{a-zero-local-map-can-coexist-with-visual-dependence}{%
\subsection{A zero local map can coexist with visual
dependence}\label{a-zero-local-map-can-coexist-with-visual-dependence}}

SmolVLM, Qwen3-VL, and Pixtral have zero last-layer Grad-CAM maps, yet
their TAM, Beyond Intuition, and generic relevance maps are nonzero.
These methods inspect different paths or aggregate connections under
different clipping rules. The actual residual replacement in Section
\ref{sec:patching} also changes SmolVLM's answer score. A zero local,
rectified heatmap therefore does not establish that the image is
irrelevant to the model. Local Grad-CAM offers a focused and inexpensive
view; propagation examines broader dependencies while introducing
additional modeling assumptions.

The adapters now expose the attention probabilities actually multiplied
by values in the supported eager decoders. ATTATTR therefore uses the
same public call across Llama, Qwen2, Qwen3-VL, and Mistral, without an
architecture- specific user hook. A measured zero remains a result
rather than an unavailable method. Qwen3.5's separate hybrid-stack
limitations are discussed in Appendix \ref{sec:qwen35}.

\hypertarget{deletion-tests-reveal-disagreements-hidden-by-plausible-heatmaps}{%
\subsection{Deletion tests reveal disagreements hidden by plausible
heatmaps}\label{deletion-tests-reveal-disagreements-hidden-by-plausible-heatmaps}}

\begin{figure}[!htbp]
\centering
\includegraphics[width=\linewidth,height=0.64\textheight,keepaspectratio]{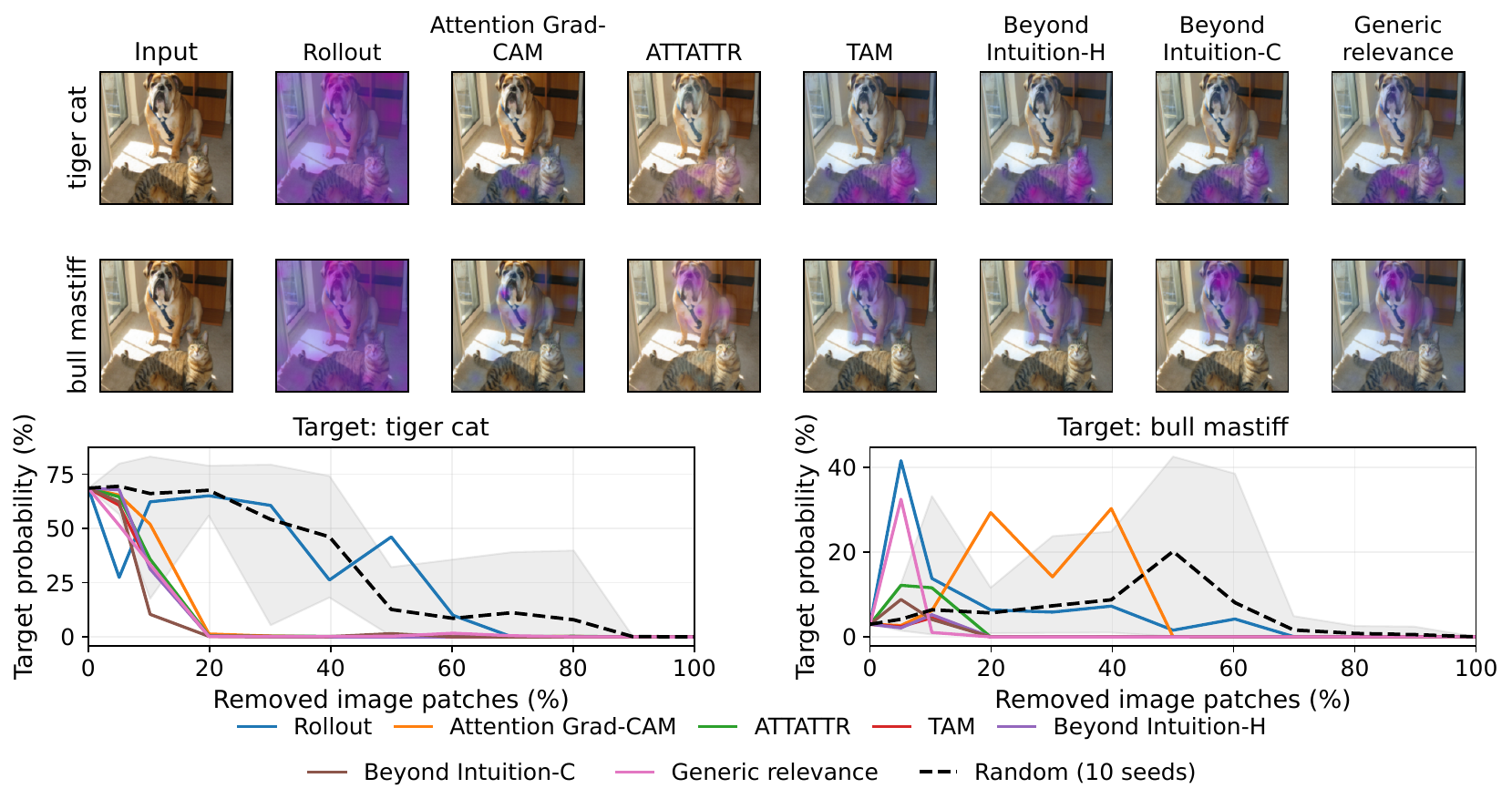}
\caption{Object-conditioned attribution and deletion on ViT-B/16. Seven maps explain cat and dog in the same author-provided image. Curves measure class probability after replacing top-ranked patches with training-mean RGB. Rollout combines target-independent connections, while other tools can separate the classes; deletion exposes differences that map appearance alone does not resolve.}
\label{fig:vit}
\end{figure}

To check these methods on their original vision structure, we use
pretrained ViT-B/16 and an author-released cat--dog image. This control
is a vision classifier, not a VLM column omitted from the preceding
comparison. Figure \ref{fig:vit} combines the seven maps and their
matched deletion curves in one figure. The image is unchanged between
rows; the explained class changes from cat to dog. Target-independent
rollout consequently stays the same, whereas target-conditioned tools
can separate the objects.

Deleting 39 of 196 patches leaves cat probability at 65.1\% under
rollout, compared with 0.27\% under Beyond-C; the original value is
68.6\%. Conversely, the dog Grad-CAM map looks recognizable, but its
deletion raises dog probability from 2.99\% to 29.3\%. A visual
explanation and the effect of removing its selected regions can thus
disagree. Deletion supplies an intervention-based diagnostic for
choosing among maps, while remaining sensitive to the replacement
baseline and removal of competing objects. The library supports both
parts of this comparison rather than treating visual sharpness as
sufficient evidence of faithfulness.

\FloatBarrier

\hypertarget{causal-interventions}{%
\section{Causal interventions}\label{causal-interventions}}

\hypertarget{attribution-patching-can-miss-large-replacement-effects}{%
\subsection{Attribution patching can miss large replacement
effects}\label{attribution-patching-can-miss-large-replacement-effects}}

\label{sec:patching}

Activation patching replaces a selected activation with its value from a
reference input and recomputes downstream behavior \autocite{rome}.
Attribution patching estimates the same replacement from a local
gradient \autocite{attribution}. For a residual \(h_{\ell,S}\) at token
set \(S\) and aligned reference value \(d_{\ell,S}\),

\[
\Delta_{\ell,S}=S(x;h_{\ell,S}\leftarrow d_{\ell,S})-S(x),\qquad
\widehat\Delta_{\ell,S}=\left\langle
\nabla_{h_{\ell,S}}S(x),\ d_{\ell,S}-h_{\ell,S}\right\rangle.
\]

\begin{figure}[!htbp]
\centering
\includegraphics[width=\linewidth,height=0.80\textheight,keepaspectratio]{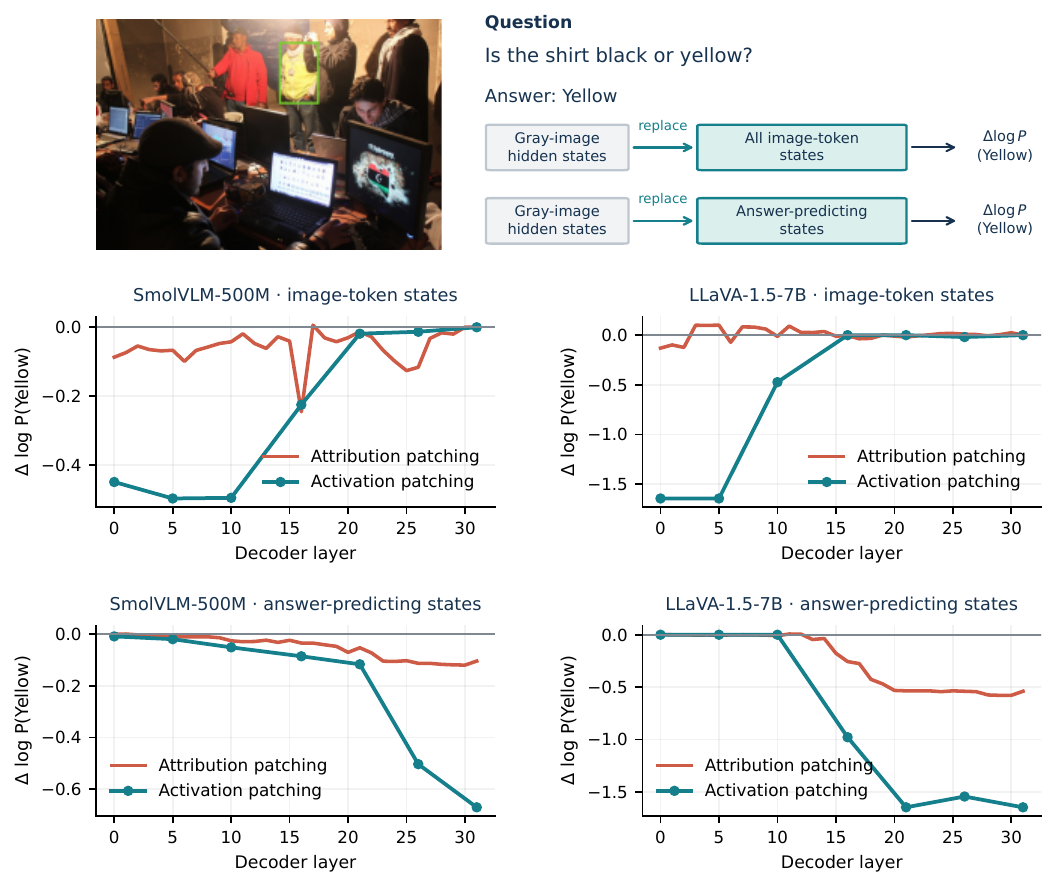}
\caption{Attribution patching can miss effects measured by activation patching. Gray-image hidden states replace selected states in the real-image run at one decoder layer. The schematic rows are separate interventions: all image-token states or the states that predict the answer's token pieces. Teal markers measure changes in the complete-answer log probability; red curves estimate the same changes from gradients. Negative values reduce support for Yellow. The estimates miss large early image-state and late answer-state effects.}
\label{fig:patch-comparison}
\end{figure}

We compare these effects on the same GQA input in SmolVLM and LLaVA. The
reference input is a gray image with identical dimensions, prompt, and
answer prefixes. Figure \ref{fig:patch-comparison} distinguishes
replacement of all visual states from replacement of the
answer-predicting states. These are hidden-state interventions: the
image and text of the patched run stay fixed. Answer-predicting states
are the residual vectors used to predict each piece of the scored
answer; a multi-piece answer can involve more than one position. Actual
effects are measured at seven layers; first-order estimates cover every
layer. The diagram shows the two alternatives, tested in separate runs.

SmolVLM's layer-5 visual replacement changes \(\log P(\mathrm{Yellow})\)
by \(-0.498\), compared with an estimate of \(-0.067\). LLaVA's
corresponding effect is \(-1.646\), while the estimate is \(+0.100\),
reversing the sign. Large blank-image replacements leave the
neighborhood described by a local gradient. Attribution patching can
cheaply nominate candidate sites, but these results motivate testing
important candidates with actual replacement. They do not imply that
every smaller head- or token-level estimate will fail.

\hypertarget{the-intervention-unit-determines-the-apparent-locus-of-dependence}{%
\subsection{The intervention unit determines the apparent locus of
dependence}\label{the-intervention-unit-determines-the-apparent-locus-of-dependence}}

Image-slot replacement is strongest early in the yellow-shirt case;
replacing prediction slots has larger late effects. The final
visual-slot effect is zero because no remaining decoder layer transports
the change into an answer position. That zero is a consequence of the
intervention location, rather than evidence that the model ignored the
image.

\begin{figure}[!htbp]
\centering
\includegraphics[width=\linewidth]{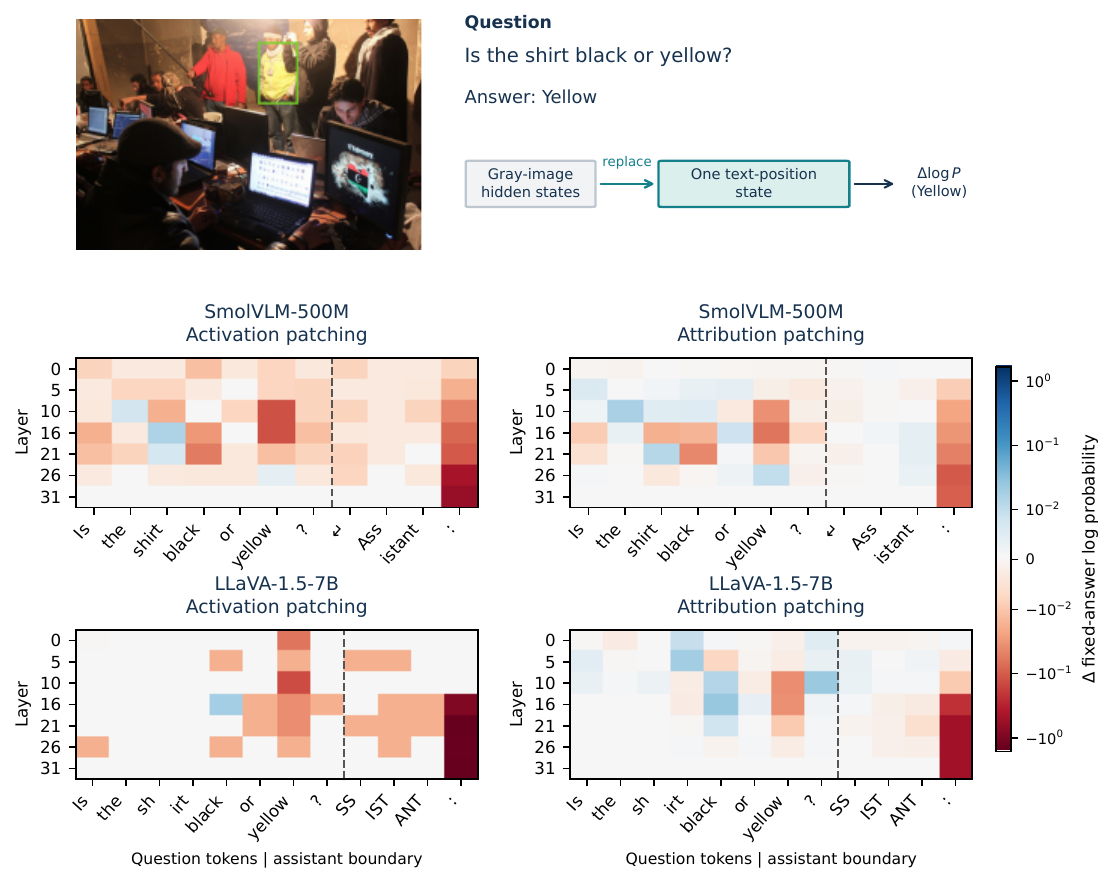}
\caption{Single-position replacement separates question-word effects from prediction-site sensitivity. Each cell replaces one native text-position state with its corresponding gray-image state at the indicated layer, as depicted in the schematic. Columns cover question and assistant-boundary subwords. Red reduces the complete-answer log probability of Yellow; blue increases it. Activation patching measures the replacement effect and attribution patching estimates it, using the same signed scale. Contrast with the joint replacements in Figure \ref{fig:patch-comparison}.}
\label{fig:patch-tokens}
\end{figure}

Figure \ref{fig:patch-tokens} refines the experiment to one native text
position at a time. Rows are layers and columns are question or
assistant-boundary subwords; actual and estimated effects share a signed
scale. Replacing the question word ``yellow'' at an intermediate layer
changes the answer score in both models. The largest late effects occur
at the final prediction position, whose role makes this sensitivity
expected. This distinguishes sensitivity at question-token sites from
disruption of the answer readout itself.

The black-shirt input provides a useful contrast (Figures
\ref{fig:patch-black} and \ref{fig:tokens-black}). LLaVA's largest
prediction-slot effect is only \(-0.172\), compared with \(-1.646\) on
Yellow. The maps therefore characterize a specified input and
corruption. Neither two examples nor a visually similar layer pattern
establishes a universal pathway or model-level robustness ranking.

\hypertarget{knockout-tests-information-routes-while-path-patching-isolates-senders}{%
\subsection{Knockout tests information routes, while path patching
isolates
senders}\label{knockout-tests-information-routes-while-path-patching-isolates-senders}}

\begin{figure}[!htbp]
\centering
\includegraphics[width=\linewidth]{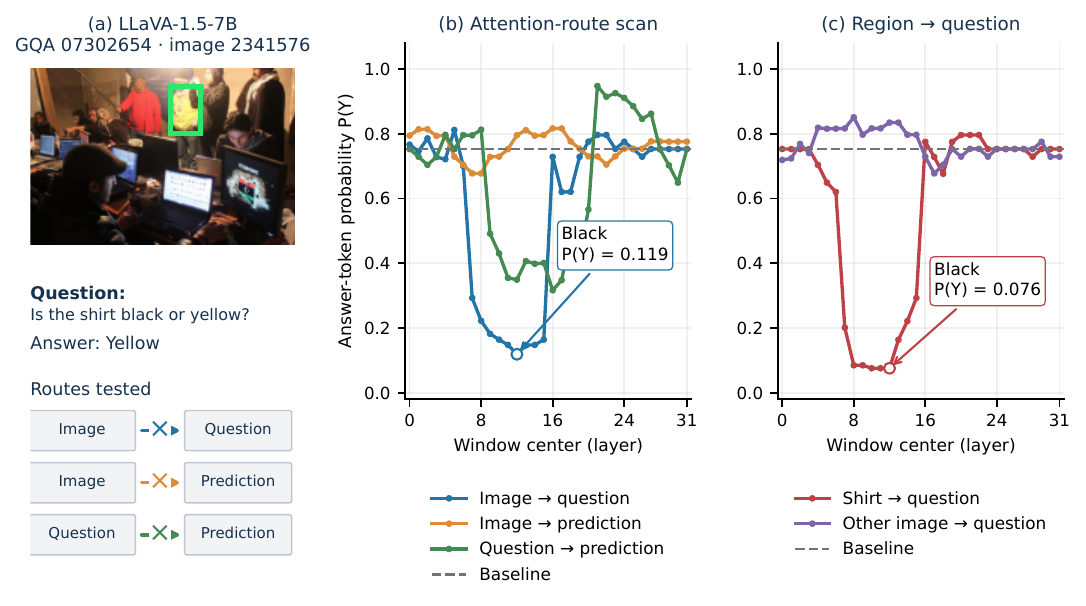}
\caption{Attention Knockout reveals route-specific dependence in LLaVA-1.5-7B. Blocking image-to-question connections at layers 8--16 lowers the fixed Yellow first-subword probability from 0.753 to 0.119 and makes Black the next-token prediction. Blocking shirt keys lowers it to 0.076; region sizes differ, so these curves describe the selected masks rather than controlled localization accuracy.}
\label{fig:knockout}
\end{figure}

Attention knockout blocks selected key-to-query connections before
softmax, allowing downstream computation to recompute
\autocite{knockout}. On the yellow-shirt GQA input, we compare
image-to-question, image-to-prediction, and question-to-prediction
routes over windows of up to nine layers. Figure \ref{fig:knockout}
plots the fixed first-answer-token probability and compares
target-region and other-image keys. The displayed question identifies
the input. This paper-aligned demonstration uses a single-word-or-phrase
instruction and a first-subword metric, rather than the shared
complete-answer score; Appendix \ref{sec:details} records the full
instruction.

The image-to-question route is more disruptive than the direct
image-to-prediction route on this example. Other-region blockade can
instead increase answer support. Knockout tests dependence on a route
without requiring a reference input, but changes attention normalization
and depends on the mask/window. The target region contains 20 keys
versus 556 outside it, so a size-matched control would be required to
attribute the difference specifically to grounding.

\begin{figure}[!htbp]
\centering
\includegraphics[width=\linewidth]{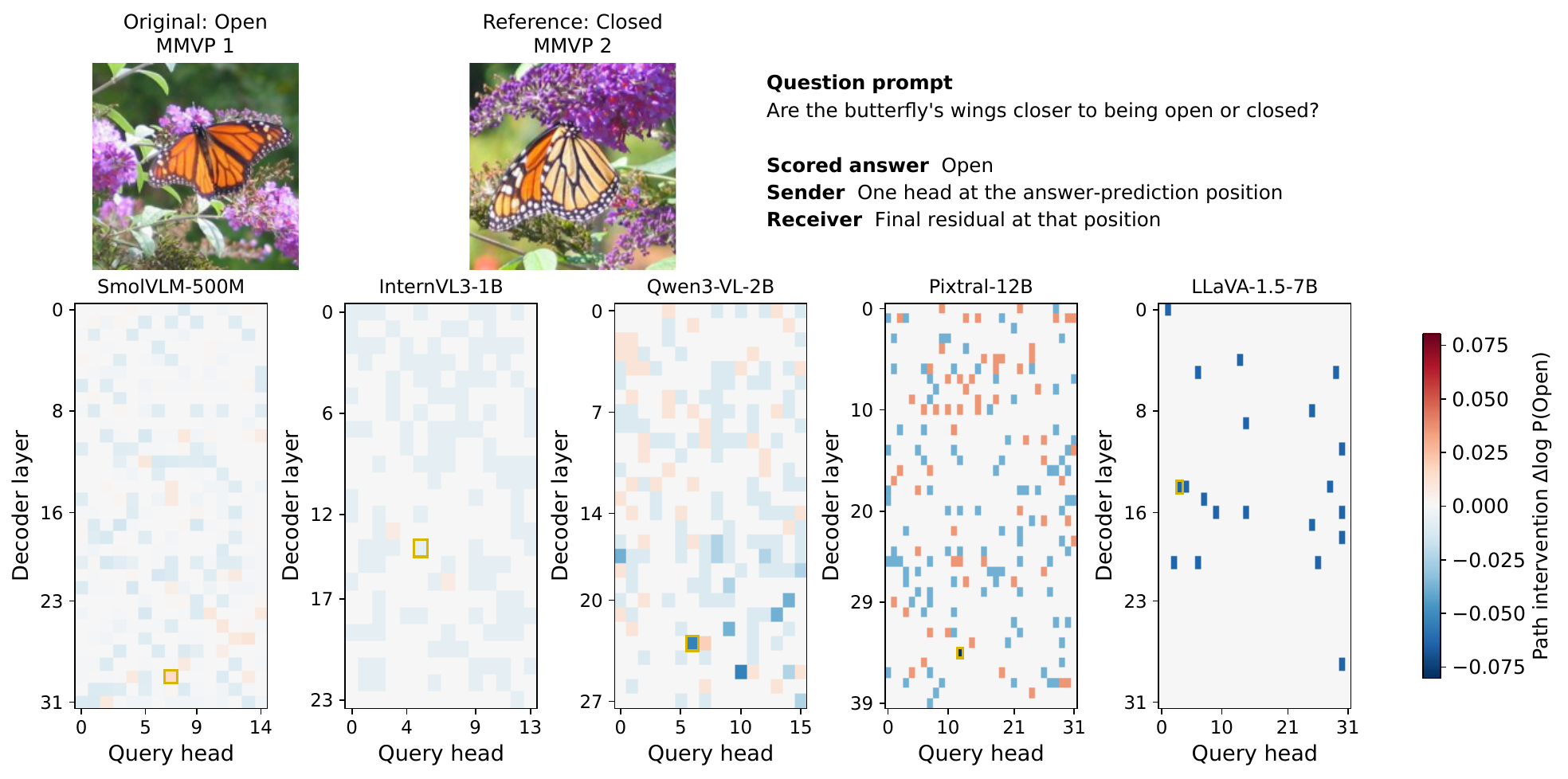}
\caption{Controlled sender-head paths on a paired butterfly-wing input. Each cell replaces one sender using its reference-image value at the prediction position, with other head outputs frozen and only the recomputed final residual replayed. Other head outputs are frozen during the controlled run; MLP and normalization recompute, and only the resulting receiver is injected in a fresh run. Signed effects compare the same configured receiver across five VLMs. Yellow outlines identify the largest absolute effect within each model; blue decreases and red increases support for Open.}
\label{fig:path}
\end{figure}

IOI-style path patching asks a narrower question \autocite{ioi}. A
controlled base run replaces a sender head with its reference-input
output while freezing other attention-head outputs to base values. MLPs
and normalization recompute; a fresh base run then replays only the
resulting receiver activation. Figure \ref{fig:path} sweeps sender heads
on a paired MMVP wing-status input in five VLMs, using the final
prediction-position residual as receiver.

The largest negative effects include Qwen3-VL L23.H6 (\(-0.054\)),
Pixtral L34.H12 (\(-0.080\)), and LLaVA L14.H3 (\(-0.064\)). The
heatmaps identify influential senders under this receiver and freezing
rule. They are not an automatically discovered complete circuit or
functional labels such as ``name mover.'' Path patching offers tighter
control of selected paths than unrestricted residual replacement, at the
cost of more forwards and a receiver-dependent question.

\hypertarget{circuit-attribution-requires-measured-recovery}{%
\subsection{Circuit attribution requires measured
recovery}\label{circuit-attribution-requires-measured-recovery}}

\begin{figure}[!htbp]
\centering
\includegraphics[width=\linewidth]{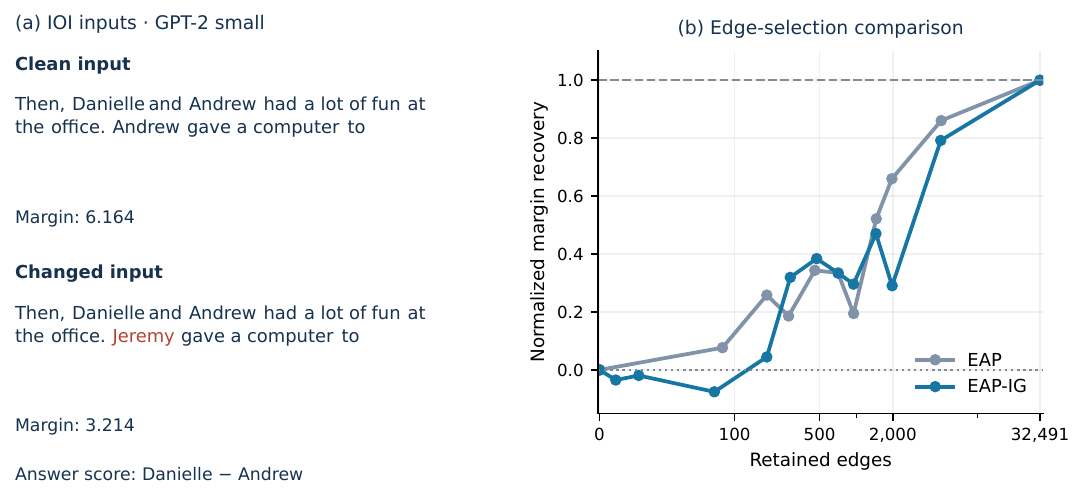}
\caption{Attribution rankings differ in actual IOI circuit recovery. EAP and input-path EAP-IG select connections in the same GPT-2 graph; each curve point is a fresh retained-edge intervention. Both complete graphs recover the clean margin, but EAP-IG does not uniformly outperform EAP on this single prompt pair.}
\label{fig:eap}
\end{figure}

EAP and EAP-IG rank independently replaceable connections in an explicit
computation graph; EAP-IG integrates gradients along the input-embedding
path \autocite{eapig}. We retain the paper-aligned GPT-2 IOI graph,
rather than substituting VLM residual nodes for unimplemented edges.
Figure \ref{fig:eap} evaluates both rankings by actual edge-retention
interventions on the same 32,491-edge graph. Excluded edges receive
reference messages, while retained edges remain live. Recovery is
measured as

\[
R(E)=\frac{m(E)-m(\mathrm{reference})}
{m(\mathrm{clean})-m(\mathrm{reference})},
\]

where \(m\) is the Danielle-minus-Andrew logit margin. This quantity can
fall below zero or exceed one; it is not a probability or a sum of
attribution scores.

At 184 retained edges, EAP recovers 0.258 of the original-reference
margin difference, while EAP-IG recovers 0.045. Integrating gradients
therefore does not guarantee a better selected circuit on each instance.
The practical benefit of this comparison is the separation of ranking
and validation: a high attribution score proposes a connection, and edge
retention tests the proposed subgraph. The graph-specific intervention
contract and single-pair evaluation limit what can be inferred about
general circuit quality.

\hypertarget{steering-depends-on-both-model-and-strength}{%
\subsection{Steering depends on both model and
strength}\label{steering-depends-on-both-model-and-strength}}

VSV, the visual steering component of VISTA, adds an image-derived
residual direction during generation \autocite{vista}. Figure
\ref{fig:vsv} holds a COCO image and its description request fixed
across LLaVA, SmolVLM and Qwen3-VL. Each row compares the model's
baseline with two smaller steering scales. Exact excerpts retain the
changed content, and ellipses indicate omissions. These outputs examine
generation changes directly, rather than necessity under a corruption.

\par\noindent
\begin{minipage}{\linewidth}
\centering
\captionsetup{type=figure}
\includegraphics[width=0.94\linewidth]{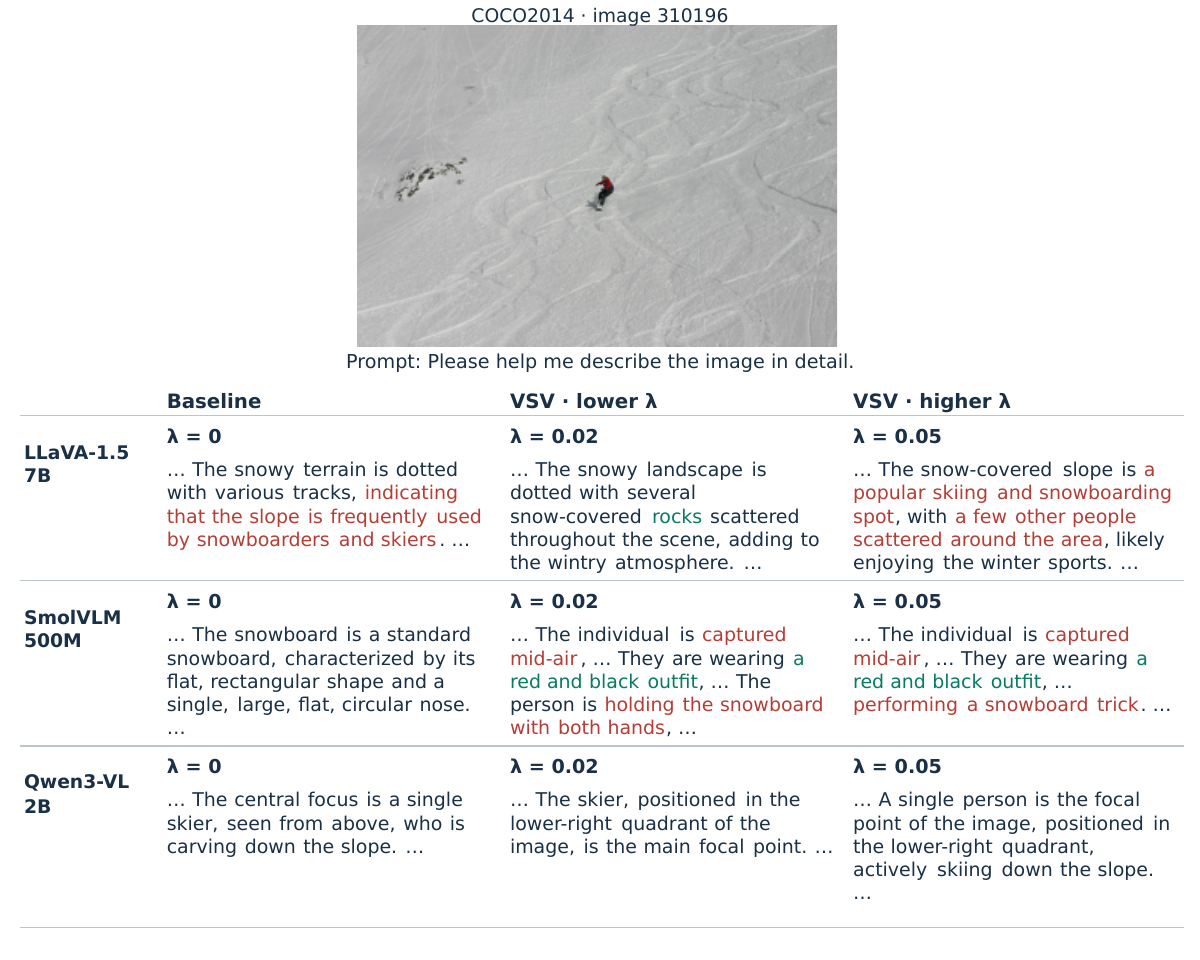}
\caption{The same image-description request under residual VSV in three models. Rows hold the model fixed; columns compare its baseline with two steering strengths, printed in each cell. Excerpts preserve the generated wording, with ellipses marking omitted text. Green highlights newly mentioned details supported by the image; red highlights selected claims without image support. Uncolored text is not fully audited. The common intervention follows the raw residual rule, without SLA.}
\label{fig:vsv}
\end{minipage}
\par\vspace{7pt}

LLaVA illustrates why fluency and grounding need separate inspection.
Its baseline already mentions tracks but infers frequent use of the
slope. At strength 0.02, it instead mentions visible rocks; at 0.05, it
invents other people. All three descriptions are grammatical. The
retained 0.17 stress run repeats snow-related text even after exact
text-token alignment.

SmolVLM changes from a detailed equipment description to a maneuver
description, adding the visible red-and-black clothing but also an
unsupported airborne pose and board-holding claim. Qwen3-VL retains a
single-rider account while changing its spatial framing and track
description. The same raw scale therefore changes different claims in
different models; added visual detail and newly invented content can
occur within the same description.

Our intervention uses the paper's raw block-residual rule, without SLA.
The relative additive displacement depends on
\(\lambda\|v_\ell\|/\|h_\ell\|\), so identical scales need not produce
equivalent perturbations across models. The released VISTA code instead
uses normalized MLP-output directions and additional direction
processing. Its numerical scale cannot be assumed to transfer to this
implementation.

Figure \ref{fig:vsv-ranks} tests local evidence under fixed prefixes
rather than comparing the changing prefixes of those descriptions. The
maps give person/car token ranks at the same 80 baseline prefixes; the
bottom curve holds one prefix and prediction position fixed across
strengths.

\par\noindent
\begin{minipage}{\linewidth}
\centering
\captionsetup{type=figure}
\includegraphics[width=0.92\linewidth,trim=0 0 0 108,clip]{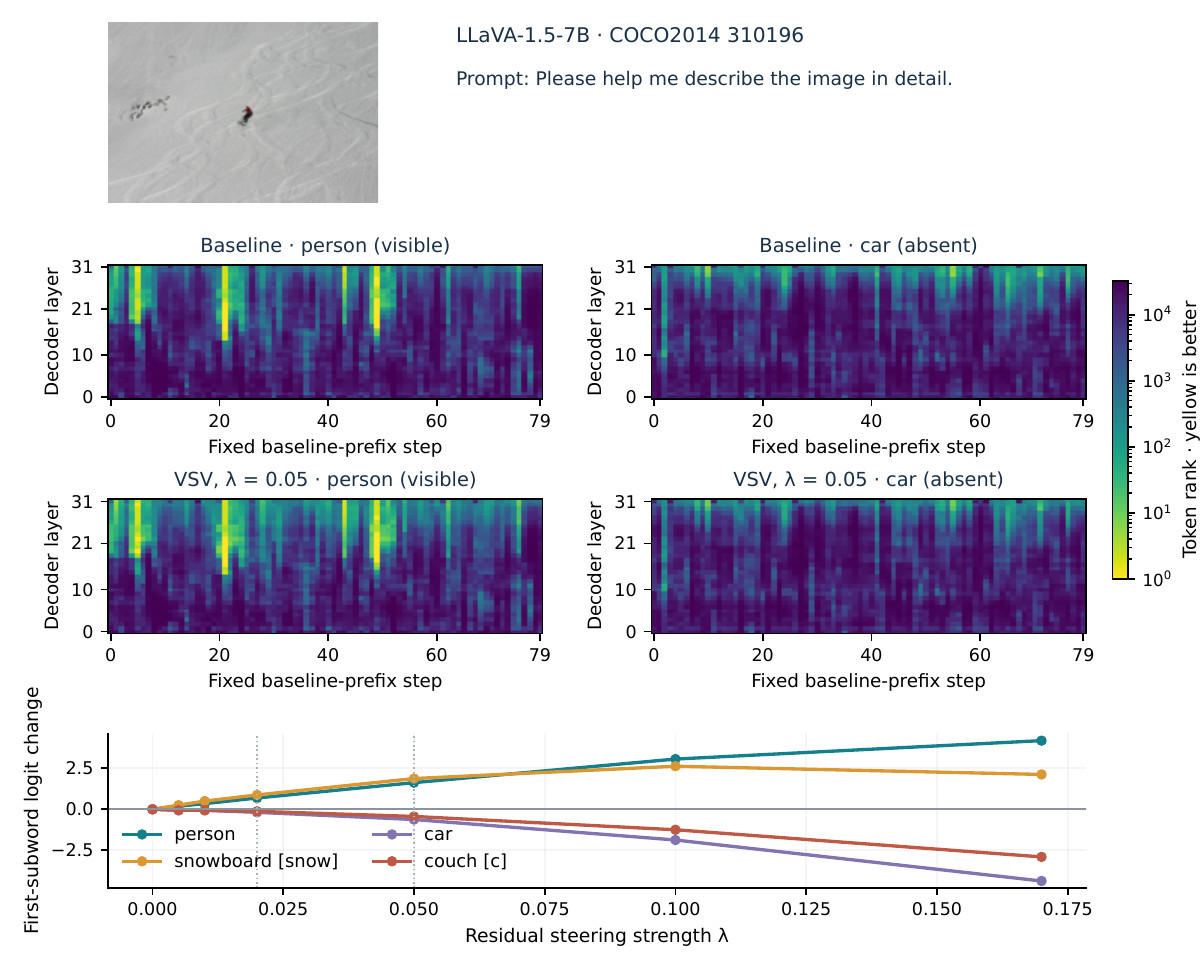}
\caption{Object-token support under fixed prefixes does not guarantee a grounded description. In the matched LLaVA run, baseline and steered person/car heatmaps share the same 80 baseline prefixes and logarithmic rank scale; brighter colors indicate better ranks. The strength curves fix the prefix “The image captures a snowboarder skillfully riding” and measure first-subword logit changes, not complete-word probabilities. At the highest retained strength, favorable person/car responses coexist with repetitive free generation.}
\label{fig:vsv-ranks}
\end{minipage}
\par

At strength 0.17, person's logit increases by 4.19 and car's
decreases by 4.38, despite the repetitive free generation. A favorable
object-token response therefore need not yield a grounded, fluent
description. VSV offers a direct way to alter visual influence, but its
strength requires evaluation against both grounding and decoding
quality. This is a Steering/VSV study; the SLA component is absent, so
it does not represent complete VISTA reproduction.

Across these intervention groups, the library keeps the change and its
question explicit. Residual replacement tests a counterfactual
activation; knockout tests a blocked route; path patching controls
sender influence; edge retention evaluates a proposed circuit; steering
changes generation. Matched comparisons clarify differences within those
questions, while separate studies preserve the meaning of methods with
incompatible interventions.

\FloatBarrier
\section{Sparse features connect reconstruction, concepts and spatial structure}
\label{sec:sae}

A sparse autoencoder (SAE) represents a native activation \(h\) using an
expanded dictionary of features,
\[
z=E(h),\qquad \widehat h=D(z).
\]
The encoder selects a sparse feature vector; the decoder reconstructs the
original representation. This provides feature coordinates for concept
readout and spatial analysis. Feature indices acquire
interpretations through measurements on independent inputs. They are not
predefined concept labels.

The matched language experiment evaluates six pretrained families. ReLU
with an L1 penalty encourages
sparsity by penalizing rectified activations \autocite{sae_relu}; TopK retains
the largest positive activations at each token \autocite{sae_topk}. Gated
SAEs separate feature selection from magnitude estimation
\autocite{sae_gated}, while JumpReLU learns activation thresholds
\autocite{sae_jumprelu}. BatchTopK allocates a sparsity budget across training
examples and uses a saved threshold for inference \autocite{sae_batchtopk}.
Matryoshka BatchTopK additionally trains nested dictionary prefixes to
reconstruct activations at several feature budgets \autocite{sae_matryoshka}.
Our experiments use published weights throughout: neither the backbone nor
any SAE is trained or recalibrated. The same native interface supplies
features and reconstruction to the shared evaluation tools.

\subsection{Matched language dictionaries expose separate trade-offs}
\label{sec:sae-language}

The language comparison uses 18 SAEBench dictionaries: three source settings
for each family, all attached to Pythia-160M block~8, with input width 768
and dictionary width 4,096 \autocite{saebench}. All checkpoints receive the
same 524,288 activations from 512 separate, 1,024-token Pile test-file
contexts \autocite{pile}. An independent 128-context calibration subset
supplies a fixed reference mean \(\mu_{\mathrm{cal}}\). We report global
reconstruction and realized activity as
\[
\mathrm{FVU}_{\mu_{\mathrm{cal}}}
 =\frac{\sum_i\lVert h_i-D(E(h_i))\rVert_2^2}
 {\sum_i\lVert h_i-\mu_{\mathrm{cal}}\rVert_2^2},
\qquad
\overline L_0=\frac{1}{N}\sum_i\sum_j\mathbf{1}[z_{ij}\ne0].
\]
The ratio is computed after aggregating all selected positions. Averaging
per-batch FVU would change the metric with batch partitioning. Behavioral
fidelity is measured on 64 fixed test contexts, containing 65,472 shifted
next-token predictions: the SAE reconstruction replaces the selected native
residual, and the remaining model recomputes. We record the increase in
negative log-likelihood (\(\Delta\mathrm{NLL}\)), distributional KL and loss
recovered relative to a separate zero-residual control.

\begin{figure}[!htbp]
\centering
\includegraphics[width=\linewidth]{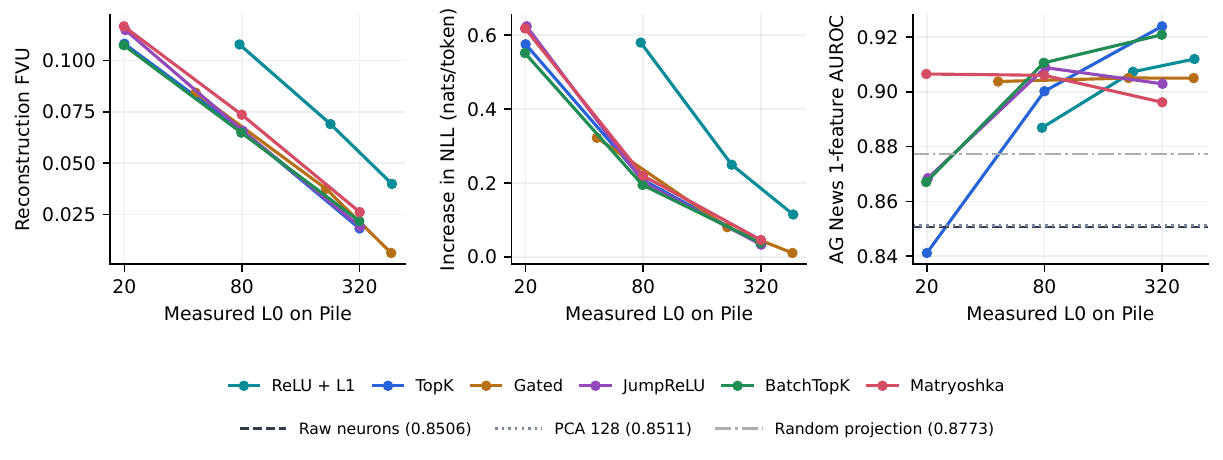}
\caption{Pretrained language SAE trade-offs under a common backbone, layer
and dictionary width. Each point is one of 18 fixed checkpoints. Left:
Pile reconstruction FVU with an independent calibration mean. Middle:
next-token NLL increase after native residual reconstruction. Right:
held-out AG News macro AUROC using one selected feature per one-versus-rest
class; horizontal lines show matched raw-neuron, PCA and random-projection
controls. All panels use measured Pile \(\overline L_0\) on the horizontal
axis. All SAE weights remain frozen during evaluation.}
\label{fig:sae-language}
\end{figure}

Concept readout uses AG News with 4,000 training, 1,000 development and
1,000 official-test documents \autocite{agnews}. Nonpadding token features
are averaged within each document, using at most 128 tokens. For each
one-versus-rest class, the probe selects one, two or five features using
training labels only. Training statistics define standardization;
development AUROC selects logistic-regression regularization. Raw residual
neurons, training-fitted PCA and a fixed random projection use the same
splits and feature budgets. Figure~\ref{fig:sae-language} places concept
readout beside reconstruction and model fidelity. Its horizontal axis is
realized Pile activity, rather than a nominal hyperparameter shared across
incompatible sparsity objectives.

At approximately 80 active features per token, BatchTopK gives FVU 0.0648
and \(\Delta\mathrm{NLL}=0.1945\) nats, while the ReLU checkpoint at
\(\overline L_0=77.71\) gives 0.1078 and 0.5796 nats. The paired
NLL-increase difference is \(-0.3851\) nats, with a 95\% document-bootstrap
interval of \([-0.4192,-0.3569]\). TopK at 320 active
features reaches FVU 0.0182 and an NLL increase of 0.0341 nats. These results
show how activation reconstruction and downstream predictions change along
the available sparsity settings; L1 penalties do not impose the same
activity counts as TopK budgets.

Concept readout gives a different ordering. Near 20 active features,
Matryoshka's one-feature AUROC is 0.9065 versus TopK's 0.8412, although its
FVU is higher, 0.1166 versus 0.1081. Increasing the number of classifier
features also changes the comparison: five PCA components per class reach
AUROC 0.9481, above every evaluated five-feature SAE probe, whose highest
AUROC is 0.9380. A reconstruction advantage therefore does not establish a
concept-readout advantage, and a useful sparse feature does not establish
that the dictionary is the best representation for every classifier budget.
This separation motivates reporting several measured behaviors rather than
collapsing SAE quality into one benchmark score \autocite{sae_reliability}.
Appendix~\ref{sec:sae-details} lists every checkpoint and the probe controls.

\subsection{Visual features expose spatial structure across models and methods}
\label{sec:sae-vision}

We ask whether compact SAE features retain useful object-associated spatial
structure, and whether good reconstruction also gives reliable localization.
Twelve pretrained configurations cover nine backbones and seven recipes
\autocite{prisma,sae_vlm,visaebench,saepp,sae_archetypal}.
ReLU and TopK share block-11 residual observations within CLIP-B/32 and
DINO-B/16, providing the closest within-backbone comparisons. The other
rows evaluate published configurations at their native sites. Eleven observe
visual encoders; Qwen2.5 observes decoder-layer-34 visual tokens.
CLIP's Gated SAE uses block-9 MLP outputs, Qwen3 uses vision-layer-23
SAE++ features, and RA-SAE uses DINOv2 with registers. These site and
backbone differences are retained when interpreting the comparison.

Every model receives the same fixed 500 COCO2017 training and 1,000
validation source images. A fixed category list and training-only
inside-versus-outside box contrast, divided by feature RMS, select one
feature per category and dictionary before testing. Native transforms can
change visible category-box unions and eligible pair counts. Backbone
weights, SAEs and inference thresholds remain frozen;
Appendix~\ref{sec:sae-details} records their source contracts.

\begin{figure}[!htbp]
\centering
\includegraphics[width=4.75in]{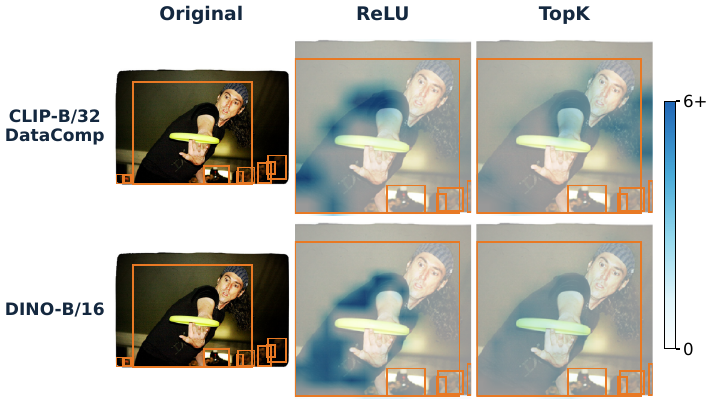}
\caption{ReLU and TopK SAE features on the same photograph for CLIP-B/32
and DINO-B/16. Rows fix the backbone and observation layer; columns compare
dictionaries. Blue heatmaps show feature activation; orange boxes mark
people.}
\label{fig:sae-matrix}
\end{figure}

For native spatial representation \(h_{u,v}\), the map is
\(M_j(u,v)=[E(h_{u,v})]_j\), reshaped on the actual visual-token grid.
Reconstruction retains the author's full feature vector, including signed
coefficients; selection and localization use \(M_j^+=\max(M_j,0)\).
Figure~\ref{fig:sae-matrix} fixes the source photograph, backbone and
block-11 residual observation within each ReLU--TopK row.

\Needspace{5\baselineskip}
Table~\ref{tab:sae-vision-comparison} separates two questions. Full-dictionary
\(\overline L_0\) and FVU measure activity and reconstruction over every
native spatial position, using a training-only reference mean for FVU.
Localization evaluates the single training-selected
feature for each category. Coverage is the fraction of eligible pairs with
positive response; Point@1 counts a hit when its strongest location falls
inside a category box, treating inactive maps as misses. Mass averages the
fraction of positive evidence inside boxes over responding pairs only.
Shuffled Point@1 preserves feature values while permuting their spatial
locations. These spatial scores are percentages; \(N\) counts eligible
image--category pairs after native cropping. Separately evaluated CLS
readouts remain in Table~\ref{tab:sae-vision}.

\begin{table}[!htbp]
\centering\small
\setlength{\tabcolsep}{3pt}
\caption{Pretrained spatial SAE configurations on COCO. Activity and
reconstruction are evaluated alongside category-feature response and
localization; shuffled maps provide a spatial control.}
\label{tab:sae-vision-comparison}
\begin{tabular}{llrrrrrrr}
\toprule
& & \multicolumn{2}{c}{Activity / fidelity} & \multicolumn{5}{c}{Spatial response} \\
\cmidrule(lr){3-4}\cmidrule(l){5-9}
Model & Recipe & \(\overline L_0\) & FVU\(\downarrow\) & \(N\) & Coverage & Point@1 & Shuffled & Mass \\
\midrule
CLIP B/32 & ReLU & 1,413.14 & 0.0152 & 1,534 & 100.0 & 54.5 & 19.2 & 39.4 \\
CLIP B/32 & TopK & 64.00 & 0.2163 & 1,534 & 66.5 & 44.7 & 16.2 & 58.8 \\
CLIP B/32 & Gated & 107.80 & 0.1261 & 1,534 & 69.4 & 47.1 & 16.3 & 57.8 \\
DINO B/16 & ReLU & 1,083.63 & 0.0063 & 1,534 & 97.9 & 52.7 & 18.6 & 38.6 \\
DINO B/16 & TopK & 1,023.82 & 0.0085 & 1,534 & 65.4 & 35.5 & 15.6 & 44.4 \\
CLIP L/14-336 & Matryoshka & 23.09 & 0.1378 & 1,534 & 83.1 & 62.5 & 18.5 & 64.9 \\
CLIP B/16 & BatchTopK & 17.05 & 0.3525 & 1,534 & 52.7 & 41.1 & 15.1 & 70.8 \\
SigLIP B/16 & BatchTopK & 252.68 & 0.2634 & 1,636 & 90.6 & 51.8 & 17.0 & 44.8 \\
DINOv2 B/14 & BatchTopK & 995.02 & 6.6730 & 1,471 & 98.2 & 66.8 & 23.8 & 52.3 \\
\shortstack[l]{Qwen2.5-VL-3B\\decoder L34} & TopK & 100.00 & 0.2078 & 1,636 & 70.7 & 45.2 & 15.0 & 53.6 \\
\shortstack[l]{Qwen3-VL-4B\\vision L23} & SAE++ L1 & 6.21 & 0.0087 & 1,636 & 86.3 & 55.5 & 16.1 & 54.3 \\
\shortstack[l]{DINOv2 B/14\\+ registers} & RA-SAE & 10.30 & 0.2977 & 1,471 & 36.4 & 29.0 & 9.9 & 70.6 \\
\bottomrule
\end{tabular}
\end{table}

\Needspace{6\baselineskip}
The matched CLIP-B/32 observations show a clear compactness--coverage
trade-off. TopK uses 64 active features per position versus ReLU's 1,413,
but FVU rises from 0.0152 to 0.2163 and Point@1 falls from 54.5\% to
44.7\%. Its greater active-map mass inside boxes (58.8\% versus 39.4\%)
coexists with lower response coverage (66.5\% versus 100\%). For these
configurations, TopK produces more concentrated responses when active;
ReLU reconstructs better and responds more reliably across image--category pairs.
\FloatBarrier

The DINO-B/16 pair has similar activity (1,084 versus 1,024 features),
yet ReLU reaches 52.7\% Point@1 versus TopK's 35.5\%. Their pointing
rates conditional on response are nearly equal (53.8\% and 54.3\%);
the overall difference mainly follows coverage, 97.9\% versus 65.4\%.
Thus an active-only localization score can hide missed responses, even
without a large sparsity difference. Gated's distinct MLP hook prevents
isolating a gating effect from its row.

Compact spatial readouts are nevertheless feasible: CLIP-L/14 Matryoshka
uses 23.09 active features, with 83.1\% coverage and 62.5\% Point@1.
Across the broader configurations, reconstruction and localization need
not agree. DINOv2 BatchTopK has FVU 6.6730, worse than the fixed
calibration-mean baseline, but reaches 66.8\% Point@1 versus 23.8\%
shuffled. RA-SAE's 70.6\% active-map mass likewise accompanies only
36.4\% coverage and 29.0\% overall Point@1. These results support
choosing configurations using reconstruction, coverage and localization
together; neither low FVU nor concentrated responding maps establish
reliable object-associated features on their own.

\begin{figure}[!htbp]
\centering
\includegraphics[width=6.7in]{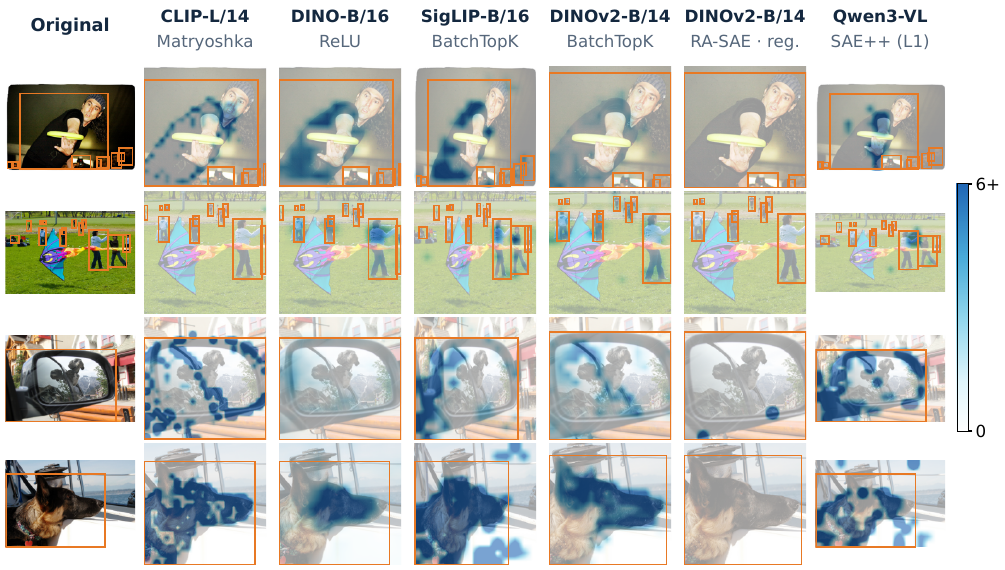}
\caption{Visual-encoder SAE features across six backbones on four
photographs. Each column uses one pretrained dictionary at its native
observation layer. Blue heatmaps show feature activation; orange boxes mark
target objects. All SAE observations precede language decoding.}
\label{fig:sae-atlas}
\end{figure}

Figure~\ref{fig:sae-atlas} compares four fixed photographs across six
visual-encoder backbones, including DINOv2 variants with and without
registers. Its first column gives uncropped originals; each remaining
column uses one fixed dictionary and the backbone's native image transform.
All columns observe visual representations before language decoding.

All twelve configurations exceed their spatially shuffled controls, but
coverage ranges from 36.4\% to 100\%. The atlas retains inactive responses:
compact, spatially associated features do not guarantee broad coverage or
faithful reconstruction.
\FloatBarrier

\subsection{Decoder features probe question-conditioned answer evidence}
\label{sec:sae-decoder}

We probe five decoder configurations: Qwen2.5-VL-3B-Instruct, Qwen3.5-4B, LLaVA-NeXT Llama3-8B,
LLaVA-v1.6-Mistral-7B and PaliGemma2-3B mix448. Each uses a published
SAE with its native encoding equation and activation boundary
\autocite{qwen25_sae,qwen35_sae,llava_decoder_sae,line_of_sight,mmdiff}.
All weights remain frozen. Qwen3.5's and PaliGemma's text-position SAEs
are transferred to image-token states.

\begin{figure}[!htbp]
\centering
\includegraphics[width=\linewidth]{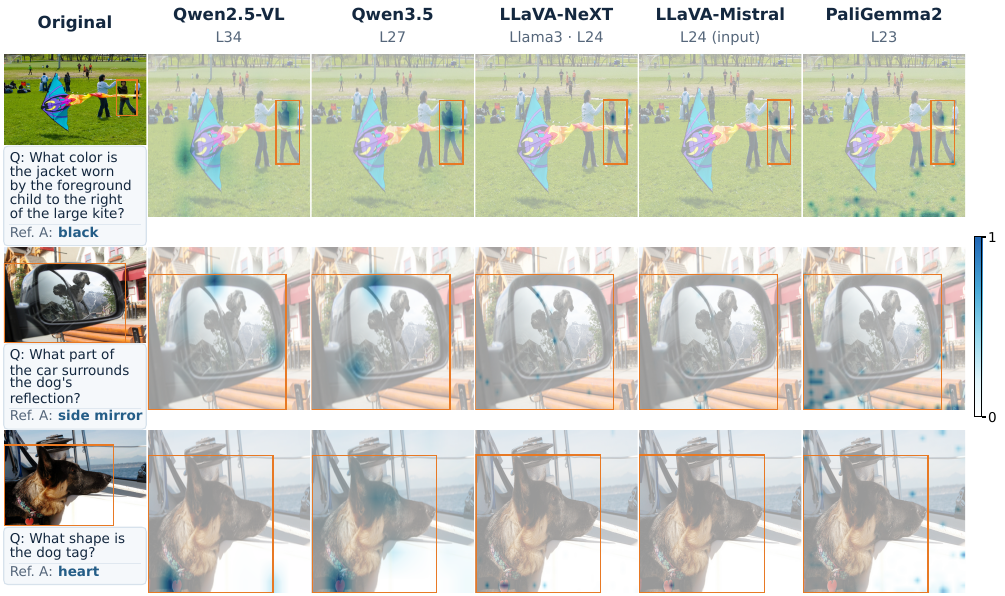}
\caption{Question-conditioned decoder SAE attribution across five model
configurations. Inputs and reference answers appear at left; heatmaps show
positive net contributions from selected SAE feature groups. Orange boxes mark
target objects; maps are normalized independently.}
\label{fig:sae-decoder-qa}
\end{figure}

Three COCO photographs are paired with eight object-focused questions:
two from VQAv2 and six constructed from visible attributes \autocite{vqav2}.
Figure~\ref{fig:sae-decoder-qa} shows one question per photograph. Each
heatmap sums local gradient-times-feature contributions to a fixed
reference-versus-foil contrast of mean answer-token log probabilities.
Up to eight feature IDs are selected by their positive contributions on
that example, without using the object box. Maps retain positive net
contributions and are normalized independently.

The same image supports different spatial patterns across decoder
configurations and answer objectives. Qwen2.5's jacket map has a strong
peak at the kite outside the target child's box, while its dog-tag map
peaks near the collar and tag. These maps locate answer-sensitive SAE
directions rather than segmenting the queried object. In Qwen and LLaVA,
causal image-prefix states cannot read the later question; question
conditioning enters through the downstream answer objective.

These case diagnostics reveal objective-dependent SAE directions, not a
method ranking. Attribution establishes neither accurate generation nor
faithful reconstruction: both LLaVA variants and PaliGemma answer ``brown''
to the jacket question, while the fixed contrast scores ``black'' against
``white.'' PaliGemma's transferred dictionary also reconstructs image states
poorly (FVU greater than one). Scoring and attribution definitions,
feature-deletion results and reconstruction/pixel controls appear in
Appendix~\ref{sec:sae-decoder-details}.

\FloatBarrier
\clearpage
\begingroup
\renewcommand{\sectionreserve}{}
\renewcommand{\subsectionreserve}{}
\printbibliography[keyword=paper,heading=reportrefs,title={References}]
\printbibliography[keyword=data,heading=subbibliography,title={Data resources}]
\endgroup

\clearpage
\appendix
\small

\hypertarget{experimental-details}{%
\section{Experimental details}\label{experimental-details}}

\label{sec:details}

\hypertarget{checkpoints-and-input-contracts}{%
\subsection{Checkpoints and input
contracts}\label{checkpoints-and-input-contracts}}

The shared VLM runs use Torch 2.6 and Transformers 5.3, BF16 weights,
eager attention, and full-sequence uncached forwards. Pixtral uses a
256-pixel longest edge and Transformers' native Mistral regex
correction, applied after tokenizer loading to avoid a duplicate-keyword
bug in version 5.3. Pinned files are unchanged. Checkpoint identities
are listed in Table \ref{tab:checkpoints}; full revisions bind the model
artifacts used by every archived run. Weights and bulk datasets are
stored outside the repository under
\texttt{/bigdata/gexin/vlm\_probing}.

\begin{table}[htbp]
\centering\footnotesize
\caption{Pinned pretrained checkpoints for the shared VLM experiments.}
\label{tab:checkpoints}
\begin{tabular}{l l}
\toprule Checkpoint & Revision \\
\midrule
\texttt{HuggingFaceTB/SmolVLM-500M-Instruct} & \texttt{a7da5b986cb59b408707209984f360a5f4ad7e47} \\
\texttt{OpenGVLab/InternVL3-1B-hf} & \texttt{014c0583a0d4bedf29fbe2dbff4f865eb998e171} \\
\texttt{Qwen/Qwen3-VL-2B-Instruct} & \texttt{89644892e4d85e24eaac8bacfd4f463576704203} \\
\texttt{mistral-community/pixtral-12b} & \texttt{c2756cbbb9422eba9f6c5c439a214b0392dfc998} \\
\texttt{llava-hf/llava-1.5-7b-hf} & \texttt{b234b804b114d9e37bb655e11cbbb5f5e971b7a9} \\
\midrule
\texttt{Qwen/Qwen3.5-4B} (appendix) & \texttt{851bf6e806efd8d0a36b00ddf55e13ccb7b8cd0a} \\
\bottomrule
\end{tabular}
\end{table}

GQA photographs come from the official image archive; questions and
boxes come from the cross-modal information-flow authors'
\href{https://github.com/FightingFighting/cross-modal-information-flow-in-MLLM/tree/main/datasets}{choose-attribute
subset}. The image IDs are 2341576 and 2342134, with boxes
\((264,33,57,89)\) and \((76,147,124,134)\) in \((x,y,w,h)\)
coordinates. MMVP uses the first released image pair. Its HF split is
named \texttt{train}, but the images are released evaluation cases, not
lens-fitting examples. VSR uses rows 0 and 1 of the zero-shot test
split. The first row's car is partly visible as a dashboard, so it is
unsuitable for a clear external-car localization claim. VisOnlyQA uses
row 0 of the real geometry-triangle and chart-extraction splits,
retaining its official no-reasoning prompts.

Shared-matrix GQA/MMVP/VSR prompts prescribe sorted answer options and
no explanations. SmolVLM uses its native leading-space continuation;
other word answers use their recorded bare-word encoding. LLaVA's Yellow
has two answer pieces. The shared score includes all pieces. Original
images, padded images, literal chat prompts, answer IDs, expanded
positions, input hashes, and native generations are archived in
\texttt{docs/report/v10/results}; previous records remain in
\texttt{docs/report/results}.

\hypertarget{hybrid-attention-in-qwen3.5}{%
\subsection{Hybrid attention in
Qwen3.5}\label{hybrid-attention-in-qwen3.5}}

\label{sec:qwen35}

The preceding experiment also evaluated Qwen3.5-4B
(\texttt{Qwen/Qwen3.5-4B}) on the eight shared inputs; its pinned
revision is included in Table \ref{tab:checkpoints}. Its 32 decoder
blocks contain 24 Gated DeltaNet blocks and eight full-attention blocks.
Residual lenses, activation patching, attribution patching, and steering
remain defined at all block outputs. Local attention methods can inspect
the full-attention blocks when the required probability sites are
exposed. Standard attention rollout, however, assumes an attention
matrix at every propagation step. Skipping DeltaNet blocks would omit
their mixing and does not reproduce the standard method. Extending
rollout requires defining and validating propagation through those
blocks. Likewise, IOI-style path patching requires a controlled
output-freezing contract for every component in the stack, which the
current DeltaNet adapter does not supply. These structural boundaries
differ from a missing editable softmax tap in a conventional decoder.
The main comparison uses Pixtral to retain full-stack methods across all
columns; the Qwen3.5 residual results remain in the archived
supplementary experiment in \texttt{docs/report/results/qwen35}.

\hypertarget{lens-fitting-and-readout-scope}{%
\subsection{Lens fitting and readout
scope}\label{lens-fitting-and-readout-scope}}

The two VLM Tuned Lenses use 1,024 COCO2014 training images and 256
separate validation images, with eight caption prediction positions per
image. The frozen model supplies 8,192 training and 2,048 validation
positions. At layers 7, 15, and 23, FP32 affine translators minimize
teacher-distribution KL with Adam, learning rate \(3\times10^{-4}\),
batch size 64, a maximum of 12 epochs, patience 3, improvement threshold
\(10^{-4}\), and seed 20261007. Epoch zero is a valid model-selection
candidate. Shared report cases are excluded from fitting; recorded
SmolVLM overlap checks include image hashes and VSR COCO source IDs.
LLaVA artifacts were fitted previously using SDPA and image-mean
padding, then replayed under the report's eager/gray conditions.

Table \ref{tab:fit} reports independent validation fidelity. It records
calibration quality on held-out caption positions, rather than
ground-truth answer accuracy or a comparison of different lens
objectives.

\begin{table}[htbp]
\centering\small
\caption{Tuned Lens validation KL in nats per selected position. Initial and selected translators are evaluated on the same held-out caption positions.}
\label{tab:fit}
\begin{tabular}{l r r r r}
\toprule Model & Layer & Initial KL & Selected KL & Epoch \\
\midrule
SmolVLM & 7 & 14.3279 & 1.9050 & 11 \\
SmolVLM & 15 & 11.4524 & 1.4967 & 12 \\
SmolVLM & 23 & 6.0436 & 0.7889 & 11 \\
LLaVA & 7 & 7.8680 & 1.3609 & 12 \\
LLaVA & 15 & 5.9005 & 0.9137 & 12 \\
LLaVA & 23 & 4.5529 & 0.3782 & 11 \\
\bottomrule
\end{tabular}
\end{table}

The GPT-2 Tuned Lens is imported from the authors and mapped to
post-block sites. Attention Lens is fitted on 4,096 BookCorpus contexts
\autocite{bookcorpus}, with 512 WikiText-2 validation contexts for
cross-domain validation \autocite{wikitext}. BookCorpus uses the
historical HF archive's \texttt{books\_large\_p2.txt} prefix, not
BookCorpusOpen; each context has 96 tokens and supervises the final
position. WikiText uses \texttt{Salesforce/wikitext},
\texttt{wikitext-2-raw-v1}; its validation and test splits supply 512
contexts each. Jacobian Lens calibrates all twelve GPT-2 residual layers
on 256 WikiText-2 training contexts. The preceding experiment selected
only layers 3, 7, and 10; its uncalibrated rows were a configuration
choice, not a method limitation. The Jacobian readout estimates causal
transport into the final residual basis; its scores are not calibrated
next-token probabilities. The available artifacts support the text
study, not a five-VLM calibrated Jacobian/Attention Lens comparison.

EmbedLens reports lexical neighbors of projected visual input
embeddings, without category thresholds for the full paper's token
classification. EmbedLens and Patchscope retain the earlier
single-word-or-phrase GQA instruction. Images and questions are shared,
but their source residuals are not pooled with the matrix's
forced-option run. Patchscope uses source layers 7, 15, 23, and 31 at
the answer-prediction position, replacing the target placeholder after
layer 0 with an identity mapping. Its zero-vector control uses the same
target site and scoring prompt. Both are readout studies, not completed
pruning or cross-model transfer trials.

\hypertarget{attention-implementations-and-spatial-display}{%
\subsection{Attention implementations and spatial
display}\label{attention-implementations-and-spatial-display}}

Single-layer Grad-CAM and ATTATTR use the last available native
attention layer; decoder propagation methods start at layer 0 and
require a consecutive stack. Qwen3-VL propagation does not include the
vision encoder, projector, or DeepStack injections as separate graph
edges. Qwen3.5 has eight full-attention layers at indices 3, 7, 11, 15,
19, 23, 27, and 31, interleaved with 24 DeltaNet layers. Skipping the
latter is not a full-stack rollout.

TAM and Beyond input integrals use 20 right-Riemann samples at
\(\alpha=1/20,\ldots,1\). SmolVLM interprets exactly zero processed
image pixels as a padding-image sentinel; excluding that endpoint avoids
removing its image token sequence. This quadrature adaptation is applied
consistently across the VLM comparison. ATTATTR retains the
official-code left-sample convention with 20 samples. TAM uses residual
normalization by \(1/2\).

The raw native patch grid is interpolated in the recorded padded-image
frame and then cropped using the original padding offsets. Each spatial
map is normalized by its own native-grid maximum absolute value, without
renormalization after cropping. In Figure \ref{fig:attention-comparison},
heatmaps remain fully opaque and the original photograph is overlaid at
20\% opacity. Color intensity follows a monotone power transform with
exponent 0.4 to make weak spatial variation visible. The palette runs
from white at zero to blue for positive values and orange for negative
ATTATTR values. This display transform does not change the archived
scores; measured zero maps are explicitly labeled. Shirt-mass diagnostics
integrate absolute values in the cropped display coordinates and are
not benchmark localization scores. Original-grid numerical arrays
remain available in the archives.

The native vision control uses the original timm/JAX checkpoint
\texttt{jx\_vit\_base\_p16\_224-80ecf9dd.pth}, losslessly converted to a
native HF 12-layer, \(14\times14\)-patch classifier. The Google ViT
identifier supplies configuration and class labels, not the measured
weights. ViT CAM/ATTATTR explain layer 11 at CLS; ATTATTR uses 20 right
samples, TAM/Beyond use 20 endpoint-inclusive samples, and Beyond-H/C
start at layers 3/5. Its
\href{https://github.com/hila-chefer/Transformer-Explainability/blob/c3e578f76b954e8528afeaaee26de3f07e3fe559/samples/catdog.png}{cat--dog}
and
\href{https://github.com/hila-chefer/Transformer-Explainability/blob/c3e578f76b954e8528afeaaee26de3f07e3fe559/samples/el2.png}{elephant--zebra}
images are author-released illustrative assets, not a benchmark subset.
Deletion sets selected processed patches to zero, equivalent to
training-mean RGB; ten seeded random orders define the displayed
envelope. Original ViT paper-specific attribution settings remain in
their source records.

\hypertarget{intervention-contracts-and-validation}{%
\subsection{Intervention contracts and
validation}\label{intervention-contracts-and-validation}}

Residual patching uses an RGB\((128,128,128)\) reference image with
identical dimensions and text. Sampled exact layers are 0, 5, 10, 16,
21, 26, and 31 on SmolVLM/LLaVA; attribution estimates use all 32
layers. Visual slots and answer-prediction slots are separate
intervention sets. Token sweeps select question subwords and the final
four assistant-boundary subwords. The public attribution API currently
replays a gradient forward per layer; within that layer one gradient
covers every token. This report does not claim the optimized all-sites
two-forward/one-backward cache. Four matched runs require 658 forwards
and 132 activation backwards, with 18.17 GiB peak allocation.

The knockout study uses the LLaVA chat instruction ``Answer the question
using a single word or phrase.'' Its question is shown in Figure
\ref{fig:knockout}. Each center \(c\) jointly masks all heads in
\([\max(0,c-4),\min(31,c+4)]\). Image keys, question queries, and the
final prediction position retain native expanded coordinates. Its metric
is \(P(Y)\) for token ID 612, not full-word Yellow probability.
Shirt/other masks contain 20/556 keys. Blocking connections already
forbidden by the causal mask is a null control. The second case has
baseline \(P(Black)=0.9982\) and a maximum image-to-question decrease of
9.43\%.

Path patching uses the paired MMVP images with the same question and
text layout. Sender outputs at the prediction position are replaced from
the reference input; every other attention-head output is fixed to its
base cache. MLP and LayerNorm recompute, the final receiver residual is
cached, and a fresh base run injects only that receiver. Complete sweeps
cover every declared sender head in the five models; SmolVLM, InternVL,
Qwen3-VL, Pixtral, and LLaVA expose 480, 336, 448, 1,280, and 1,024
paths, respectively. Full head-output freezing is unavailable for
Qwen3.5's DeltaNet stack. Selected paths are configured by the user, not
automatically assigned semantic component names.

The GPT-2 EAP graph includes input, head, and MLP output messages into
editable pre-LayerNorm Q/K/V branches, MLP inputs, and final readout.
EAP uses the clean gradient; EAP-IG uses five left input-embedding
samples \(0,0.2,0.4,0.6,0.8\), preserving position embeddings and
endpoint message differences. The author IOI CSV row is 475; its
reference replaces the final Andrew occurrence with Jeremy. This graph
experiment prepends BOS and reports its own original/reference margins,
6.1643/3.2137. It should not be numerically pooled with the separately
prepared text-lens readout. Every retained-edge curve point is an actual
forward with the recorded post-pruning edge count.

The VSV comparison uses BF16 LLaVA-1.5-7B, SmolVLM-500M and Qwen3-VL-2B
on the original COCO validation image 310196. Its common user request is
``Please help me describe the image in detail.'' Each model uses its
native chat template. Directions are the image-bearing minus image-free
last-context block residuals. The no-image input retains exactly the
positive input's text IDs after removing its complete visual span and
formatting. This preserves LLaVA's boundary whitespace instead of
retokenizing a deleted image marker. Qwen's no-image input has explicit
zero-based three-axis positions, avoiding reuse of an earlier image's
RoPE offset.

All decoder block outputs are configured. Joint intervention adds the
fixed direction at the original context-end and generated positions,
then restores the original vector norm. The retained scales are 0,
0.005, 0.01, 0.02, 0.05, 0.10 and 0.17. Every generation uses greedy
uncached full-prefix replay, a 128-token limit, and no repetition
penalty. Native uncached baselines use the same full-sequence logits
readout; the default HF last-position readout can change a BF16 argmax
tie through matrix rounding. Smol's native parity reference retains its
unmodified projected image features across every prefix; HF 5.3
otherwise removes its pixel inputs after prefill. The library generation
itself replays the image at every step. Complete outputs, IDs, direction
magnitudes, input contracts and zero/identity controls are archived.
Figure excerpts retain at most 40 original words per cell with exact
source offsets; selected visual checks are recorded separately. COCO
reference captions use both skiing and snowboarding for this image, so
that activity-label difference is not marked as an error.

Figure \ref{fig:vsv-ranks} reuses this LLaVA baseline's first 80
generated prefixes in every ranking condition. Its strength curves fix
the prefix ``The image captures a snowboarder skillfully riding'';
snowboard/couch use their first pieces, rather than complete-word
probabilities. The earlier FP16, manually formatted VSV runs remain
historical records. The current study does not include the released
implementation's PCA, normalized MLP-output rule or SLA. Repetition at
large raw-residual scales characterizes this intervention configuration,
rather than complete VISTA.

Same-input replacement and post-hook baseline restoration are exact in
the retained shared and new paired runs. Final-site readouts differ from
native BF16 logits by at most 0.125 because of selected-position matrix
rounding; FP32 graph parity is within \(1.26\times10^{-4}\). These
numerical checks verify execution, not the uniqueness of the proposed
interpretation. Newly added attribution/activation measurements preserve
the archived input and baseline scores; plotting changes leave all
source measurements unchanged.

\subsection{SAE checkpoints, splits and evaluation}
\label{sec:sae-details}

\subsubsection{Pinned dictionaries and inference functions}

The Pythia backbone is
\path{EleutherAI/pythia-160m-deduped}, revision
\nolinkurl{582159a2dfe3e712a8d47ae83dec95ae3bde8e7e}. Its safetensors SHA256 is
\nolinkurl{bbaae5c00917b163baa499fc8eb64859ee0c850c5fdecfc32f4d70dc07213575}.
Language dictionaries come from
\url{https://huggingface.co/adamkarvonen/saebench_pythia-160m-deduped_width-2pow12_date-0108},
revision \nolinkurl{ede0e8b18119ee7d77b0f21d928383a00c73843d}.
For each of \texttt{Standard}, \texttt{TopK}, \texttt{GatedSAE},
\texttt{JumpRelu}, \texttt{BatchTopK} and \texttt{MatryoshkaBatchTopK}, we use
\path{resid_post_layer_8/trainer_0}, \path{trainer_2} and
\path{trainer_4}. The full family-directory suffix is
\path{_pythia-160m-deduped__0108}. These are sparsity settings, not
independent training-seed repetitions; the selected source configurations
record seed 0. Table~\ref{tab:sae-language-all} gives the source settings and
all measured results.

The CLIP ViT-L/14-336 backbone is
\path{openai/clip-vit-large-patch14-336}, revision
\nolinkurl{55ac000197133e202aa4e8bafcd01bf5a7da6e49}, with weight SHA256
\nolinkurl{f51cb7b04a2afb04828b1b8e0edc34bbf082080d098b9874f5756e7b7ba8c807}.
Visual dictionaries come from
\url{https://huggingface.co/mateuszpach/sae-for-vlm}, revision
\nolinkurl{270d13c7cc53e2ab8e887f938b0bdd22560e9f1d}. The CLS releases are
\path{batch_top_k_20_x16} and
\path{matroyshka_batch_top_k_20_x16}, each using
\path{trainer_0/checkpoints/ae_100000.pt}. The patch release is
\path{matroyshka_batch_top_k_20_x64/random_k_2}, using
\path{trainer_0/checkpoints/ae_200000.pt}. Their SHA256 values are,
respectively,
\nolinkurl{7a17406c5199443745d2aeb76dbadb5e0a9c6083fb3163df2b5a5e8983585bf2},
\nolinkurl{105d21a4db64c70064a54ffc12ce14869a668980e048c8a777753f48a0dc9ee5}
and
\nolinkurl{e251e47c376a531b822c3c146d94b2a1e6a098f6edc8180ac60df205478420cc}.
The visual source does not supply a complete original training configuration
for these snapshots; the loading metadata specifies a minimal inference
configuration from the pinned tensor schema and author hook definition,
code revision \nolinkurl{39dff5bd6dea67fc3ef350bc7b2312e5fcfc1493}.

Conversion preserves feature order, decoder-column scales, biases and saved
thresholds. Pythia TopK uses per-token selection; BatchTopK and Matryoshka use
their saved inference thresholds. Matryoshka group sizes are
\((128,256,512,1024,2176)\) for the Pythia dictionaries,
\((1024,2048,4096,9216)\) for CLS and
\((4096,8192,16384,36864)\) for the patch dictionary. They are read from the
source rather than replaced by a default grouping.

The 18 Pythia and three CLIP ViT-L/14-336 dictionary conversions are checked
on 64 real native activations per checkpoint against independent
implementations of the saved author tensor
equations, pinned at
\nolinkurl{52b0c54ba92630cfb2ae007f020ed447d4a5ba9f}. The largest relative L2
error is \(2.45\times10^{-7}\) for encoding and
\(2.29\times10^{-7}\) for reconstruction; the largest absolute error is
\(2.44\times10^{-4}\). These comparisons retain raw saved feature magnitudes,
before optional loader-side decoder renormalization. They establish numerical
agreement for the declared functions, not a semantic interpretation of their
features.

Language and CLIP ViT-L/14-336 forwards use float32 weights, Torch 2.6,
Transformers 5.3, eager attention and uncached full-sequence execution.
Native hook captures equal the corresponding cached Hugging Face hidden
states on 32 real inputs: block 8 is \path{hidden_states[9]} in Pythia,
and block 22 is \path{hidden_states[23]} in CLIP. The activation caches,
checkpoint files and dataset selections have recorded SHA256 values.

\subsubsection{Cross-backbone patch dictionaries}

The first four spatial checkpoints use block-11 residual outputs:
ReLU and TopK for CLIP ViT-B/32, and ReLU and TopK for DINO ViT-B/16.
These four have input dimension 768 and dictionary width 49,152. The CLIP backbone
is \path{laion/CLIP-ViT-B-32-DataComp.XL-s13B-b90K}, revision
\nolinkurl{f0e2ffa09cbadab3db6a261ec1ec56407ce42912}; its source
\path{open_clip_model.safetensors} has SHA256
\nolinkurl{3c00043509d2e3f35ec62bd85a643a393883dcda624d8208d034cc390c708297}.
The DINO backbone is \path{facebook/dino-vitb16}, revision
\nolinkurl{f205d5d8e640a89a2b8ef0369670dfc37cc07fc2}, with
\path{pytorch_model.bin} SHA256
\nolinkurl{a064e36c67289caaa5c949c0b3f7f31a0fcbcba5721f5fa12419933ec1f4fe6e}.
OpenCLIP visual tensors are mapped strictly into the equivalent native CLIP
architecture, retaining the source GELU function and all visual parameters.

These four SAEs are author-published ViT-Prisma ImageNet1k releases
\autocite{prisma,datacomp,dino}.
Table~\ref{tab:sae-visual-settings} places their native observations beside
the other spatial checkpoints. CLIP ReLU, DINO ReLU and DINO TopK were
trained on CLS plus patch tokens; CLIP TopK used patch tokens only.
The ReLU L1 coefficients are $10^{-5}$ for CLIP and
$1.288194820853984\times10^{-6}$ for DINO. Their TopK budgets are 64 and
1,024, respectively. All four are evaluated on spatial
patches only; CLS tokens are excluded from their feature maps. Neither
inference preprocessing nor the published sparsity budget is retuned on COCO.

\begin{table}[!htbp]
\centering\small
\caption{Published spatial checkpoints and native observation scopes.
\(d\) is activation width and \(m\) dictionary width. All layer numbers are
zero-based source block indices. ``Raw'' means no extra SAE input
normalization; DINOv2's backbone final normalization is part of its native
observation. SAE++ L1 is a level of a jointly trained hierarchy; RA-SAE
retains its nonlinear encoder and relaxed archetypal dictionary.}
\label{tab:sae-visual-settings}
\begin{tabularx}{\linewidth}{@{}llrrX@{}}
\toprule
Backbone & Recipe & \(d\) & \(m\) & Native observation / input normalization \\
\midrule
CLIP B/32 & ReLU & 768 & 49,152 & Block 11 residual; raw \\
CLIP B/32 & TopK & 768 & 49,152 & Block 11 residual; per-token sample normalization \\
CLIP B/32 & Gated & 768 & 49,152 & Block 9 MLP output; raw \\
DINO B/16 & ReLU & 768 & 49,152 & Block 11 residual; per-token sample normalization \\
DINO B/16 & TopK & 768 & 49,152 & Block 11 residual; per-token sample normalization \\
CLIP L/14-336 & Matryoshka & 1,024 & 65,536 & Block 22 residual; raw \\
CLIP B/16 & BatchTopK & 768 & 24,576 & Block 11 residual; raw \\
SigLIP B/16 & BatchTopK & 768 & 24,576 & Block 11 residual; raw \\
DINOv2 B/14 & BatchTopK & 768 & 24,576 & Final-normalized block 11 patches \\
Qwen2.5-VL-3B & TopK & 2,048 & 65,536 & Decoder block 34 visual tokens; raw \\
Qwen3-VL-4B & SAE++ L1 & 1,024 & 20,000 & Vision block 23 pre-merge patches; raw \\
\shortstack[l]{DINOv2 B/14\\+ registers} & RA-SAE & 768 & 32,000 & Final-normalized block 11 patches; nonlinear encoder \\
\bottomrule
\end{tabularx}
\end{table}

For the three normalized ViT-Prisma dictionaries, the source function uses
per-token mean \(\mu_h\) and
sample standard deviation \(s_h\), with correction 1:
\[
u=\frac{h-\mu_h}{s_h+10^{-5}},\qquad
\widehat h=s_h D(E(u))+\mu_h.
\]
The decoder rescales by \(s_h\), without adding the encoder's epsilon.
The CLIP ReLU checkpoint uses the unnormalized activation instead.
These functions follow ViT-Prisma code revision
\nolinkurl{64daf1425f7e5311844213fd0fb6f4127a30cd16}; feature order,
biases and decoder scales are preserved.

Each source checkpoint is named \path{weights.pt}. The pinned
\href{https://huggingface.co/Prisma-Multimodal/sparse-autoencoder-clip-b-32-sae-vanilla-x64-layer-11-hook_resid_post-l1-1e-05}{CLIP ReLU release}
uses revision \nolinkurl{2b1cb17ffb8b06e7f45a699dfc97d417f9879be4} and weight
SHA256 \nolinkurl{c4a92171b0503059a9aafc43a0d0006f2522602c70f95f71648ccd9c6aeb2136}.
The \href{https://huggingface.co/Prisma-Multimodal/imagenet-sae-top_k-64-patches_only-layer_11-hook_resid_post-64-78}{CLIP TopK release}
uses revision \nolinkurl{310b1d1679c60f792d721f9605c957a5cb4b764f} and weight
SHA256 \nolinkurl{daee1b639c478d630825e025a8bc0b77be341dafa9525b32746c66c8e3599119}.
The \href{https://huggingface.co/Prisma-Multimodal/DINO-vanilla-x64-all_patches_11-resid_post-1085-99}{DINO ReLU release}
uses revision \nolinkurl{3e3ea0cdcde4186db8069f4fe0ce219e8693155a} and weight
SHA256 \nolinkurl{ba04901a22b0a8b37255246fec6eb8cdb8e0592e844754527ab6d9e7f679f8fd}.
The \href{https://huggingface.co/Prisma-Multimodal/DINO-topk-x64-all_patches_11-resid_post-1024-99}{DINO TopK release}
uses revision \nolinkurl{3a6493789ca56c3fc8e85c5c437c9d704321c95c} and weight
SHA256 \nolinkurl{8cfb15d2106fdb2e7e0c408a798b34a201a1ca5d4d5bded2bf24130adfd691ad}.

Backbone verification compares all 12 native block outputs on four real
COCO images against independent source tensor equations, both before and
after the source's layer-normalization and value-bias folding. Relative L2
differences are below \(1.55\times10^{-6}\), with absolute differences below
\(1.17\times10^{-4}\). The native capture is separately checked
against Hugging Face block-11 hidden states on 32 images.
Each SAE conversion is then checked on 64 real native activations against
its pinned source equations: feature relative L2 errors are below
\(6.93\times10^{-8}\), reconstruction relative L2 errors below
\(7.10\times10^{-7}\), and absolute reconstruction errors below
\(1.15\times10^{-5}\). These checks establish the declared backbone binding
and inference function before evaluating spatial behavior.

The Gated CLIP-B/32 dictionary is
\href{https://huggingface.co/Prisma-Multimodal/8e32860c-clip-b-sae-gated-all-tokens-x64-layer-9-mlp-out-v1}{the published Prisma block-9 MLP release},
revision \nolinkurl{2b72983a021743c1b10ef3eba2b0123e2bbb4f36}.
Its \path{n_images_2600058.pt} has SHA256
\nolinkurl{64f38e1b82776607af893690a6beb7efa9711814679432af0f6813c0f1b10a4e}.
The source uses 768-dimensional MLP outputs, 49,152 features, an L1
coefficient of 0.7, and no extra activation normalization. Its encoder is
\[
\begin{aligned}
a&=(h-b_d)W_e+b_g,\\
m&=\max\!\left((h-b_d)(W_e\odot\exp(r))+b_m,0\right),
\qquad z=\mathbf{1}[a>0]\odot m.
\end{aligned}
\]
with \(\widehat h=zW_d+b_d\). Here \(\exp(r)\) scales encoder columns;
the gate bias and magnitude bias remain distinct. The source was trained
on CLS plus patches; this evaluation excludes CLS and captures the actual
\path{vision_model.encoder.layers.9.mlp} output before residual addition.
The release-era Prisma OpenCLIP loader, revision
\nolinkurl{dedf59894c55fd843df1aa95388979b91a7a10e4}, explicitly disables
LayerNorm folding and centering of writing weights. The native backbone
retains those source coordinates. Independent original OpenCLIP block
calculations on four real COCO images agree with the native MLP observation
to relative L2 error \(2.18\times10^{-6}\). The 196 patch features agree
exactly with the author's Gated equations; reconstruction relative L2 error
is \(7.76\times10^{-8}\). The public model-bound SAE API also reproduces
those feature values exactly. These checks bind the additional dictionary
to its MLP observation rather than substituting a residual activation.

The three ViSAEBench dictionaries use the published
\path{batchtopk_32x_k64/model.safetensors} snapshots
\autocite{visaebench}. Their encoder is
\(a=\max(W_e h+b_e,0)\), with inference
\(z_j=a_j\mathbf{1}[a_j\geq\tau]\). The decoder dictionary retains its
published learned multiplier and has no output bias. The saved threshold
is held fixed, so training target \(k=64\) does not force 64 active features
on every COCO patch. The source thresholds are 0.54186058 for CLIP,
0.63923049 for DINOv2 and 0.66128719 for SigLIP. An inclusive float32 source
threshold is represented by its immediately preceding float32 value when
the native implementation uses a strict inequality. This preserves the
source boundary rather than estimating a new threshold from COCO.

The corresponding backbones \autocite{dinov2,siglip} are
\path{openai/clip-vit-base-patch16}, revision
\nolinkurl{57c216476eefef5ab752ec549e440a49ae4ae5f3};
\path{facebook/dinov2-base}, revision
\nolinkurl{f9e44c814b77203eaa57a6bdbbd535f21ede1415}; and
\path{google/siglip-base-patch16-224}, revision
\nolinkurl{7fd15f0689c79d79e38b1c2e2e2370a7bf2761ed}.
CLIP and SigLIP expose the raw output of vision block 11; DINOv2 applies
its final backbone LayerNorm to that block's patch residual. CLS is
excluded for CLIP and DINOv2; the selected SigLIP encoder has no CLS token.
The DINOv2 release has no register tokens. Source extraction follows
ViSAEBench code revision
\nolinkurl{deed5c8b485186c75e2fe45d1b0a37a1f2402a17}.

The CLIP SAE is
\href{https://huggingface.co/visaebench/clip-vitb16-saes}{\texttt{visaebench/clip-vitb16-saes}},
revision \nolinkurl{db1a83e91af8e0c8066491f2ccd24ebf5e20b174}, with weight
SHA256 \nolinkurl{7f23b7cf40ad46b073853a9db64ec8fbca49ea9aa64e509d368a3fa3063f13e4}.
The DINOv2 SAE is
\href{https://huggingface.co/visaebench/dinov2-vitb14-saes}{\texttt{visaebench/dinov2-vitb14-saes}},
revision \nolinkurl{fe01c6294a6c7ae8725427c651820ecf0202317d}, with weight
SHA256 \nolinkurl{bf51f18cc39affd25c170b04a09ddd537880505cb7a62e06fdc9c44d136e6625}.
The SigLIP SAE is
\href{https://huggingface.co/visaebench/siglip-vitb16-saes}{\texttt{visaebench/siglip-vitb16-saes}},
revision \nolinkurl{229ff1e539047bcf5a5ae079721dd3842a694b5b}, with weight
SHA256 \nolinkurl{817f62c0af22adfd49dda3c4babb03b1b05a5b0e5e7e8a4c4d0f2b234241b3f4}.
Checks on real native patches compare the converted encoder and decoder
against the independent source equations, including saved multipliers and
threshold boundaries; actual processor pixels verify the annotation transform.

The Qwen2.5 dictionary is
\href{https://huggingface.co/BeEnough/SAE-Res-Qwen2.5-VL-3B-Instruct-W64K-L34-K100}{\texttt{BeEnough/SAE-Res-Qwen2.5-VL-3B-Instruct-W64K-L34-K100}}
\autocite{qwen25_sae}, revision
\nolinkurl{1a138768b0b74ea6844399f43f7a7ad1a69c0bff}, with
\path{adapter_model.safetensors} SHA256
\nolinkurl{a6dd50d69cb91846bf3bb65f89fa72f62609bd3f84af3916268ed64c0818685b}.
Its backbone is \path{Qwen/Qwen2.5-VL-3B-Instruct}, revision
\nolinkurl{66285546d2b821cf421d4f5eb2576359d3770cd3}.
The native observation is the output of
\path{model.language_model.layers.34}, restricted to image-token positions.
For \(q=W_e(h-b_d)+b_e\), the author retains the 100 largest values of
\(q\) without rectification and sets all other entries to zero:
\[
z_j=q_j\,\mathbf{1}[j\in\mathrm{TopK}_{100}(q)],
\qquad \widehat h=W_d z+b_d.
\]
This is value-ranked selection, not selection by absolute magnitude.
Selected negative coefficients would remain negative in native inference.
On the captured COCO calibration and validation visual positions, no
negative retained coefficients are observed, including the selected
category-feature maps. The
source reports training on 498,000 LLaVA-NeXT-Data examples but does not
specify its training token mask; the evaluation restricts observations to
visual positions and does not infer a training mask from that choice.
The backbone uses source bfloat16 precision; SAE inference and source
conversion checks use float32 representations of the captured activations.
The fixed prompt is ``Describe the main objects in this image.'' Native
chat formatting and image/token merging are retained.

The Qwen3 dictionary is the level-1 component of
\href{https://huggingface.co/YusongZhao666/csae-ckpt}{\texttt{YusongZhao666/csae-ckpt}}
\autocite{saepp}, revision
\nolinkurl{a5a4e40f150bda1de46a22bf045e8b8d8fd92a1f}, with
\path{ae.pt} SHA256
\nolinkurl{03ffd54714141d06612a617f02cf5367258bf9381b17fc151eee73c76b6b39ba}.
It binds to \path{model.visual.blocks.23} of
\path{Qwen/Qwen3-VL-4B-Instruct}, revision
\nolinkurl{ebb281ec70b05090aa6165b016eac8ec08e71b17}.
The source SAE++ implementation is pinned at
\nolinkurl{0dd9056affeb72d06b6af865f2fff4de103a3832}.
Its level-1 encoder uses the learned threshold 9.30319691 after rectification,
with training target \(k_1=6\). The published level-1 weights and biases,
already rescaled by the author at saving, are used directly:
\[
a=\max(W_e(h-b_d)+b_e,0),\quad
z_j=a_j\mathbf{1}[a_j>\tau],\quad
\widehat h=W_dz+b_d.
\]
Although the frozen level-1 inference is represented by the native thresholded
SAE primitive, its weights were trained jointly within the SAE++ hierarchy.
The experiment does not treat level 2 as a second independent patch method
or reproduce the complete hierarchical training objective. All 315 original
vision tensors are loaded strictly from the pinned source shard; the language
model is not instantiated. Source vision computation uses float16, with
float32 SAE encoding of captured block-23 activations. The observation is
before spatial merging; grouped native patches are inverted to raster order
for image-grid analysis without changing feature values.

The RA-SAE release \autocite{sae_archetypal} is
\href{https://huggingface.co/matybohacek/RA-SAE-DINOv2-32k}{\texttt{matybohacek/RA-SAE-DINOv2-32k}},
revision \nolinkurl{1e10a216938e112302b31e3bb2f69818e59a12a9}.
Its \path{RA-SAE-DINOv2-32k.pth} has SHA256
\nolinkurl{3cf20c7e8a97e063273e5116a049f965cb74799e2abded0c5e77f2ec30e27faa}.
This ImageNet1k-trained dictionary binds to
\path{facebook/dinov2-with-registers-base}, revision
\nolinkurl{a1d738ccfa7ae170945f210395d99dde8adb1805}, whose
\path{model.safetensors} has SHA256
\nolinkurl{7a6f7b3b9fa4b8732e707476a03cd6cdce210048582f21aafb7991c17d98e362}.
It is a different backbone from the register-free ViSAEBench model.

For the final-normalized 768-dimensional observation, the saved encoder is
\[
a=\max\!\left(\mathrm{LN}_{32{,}000}(Ah+b),0\right),
\qquad z_j=a_j\mathbf{1}[a_j\geq0.829].
\]
LayerNorm acts across the 32,000 feature pre-activations, with its saved
affine parameters and epsilon \(10^{-5}\). The fixed threshold is the
author inference wrapper's value, not a COCO-fitted sparsity target.
The nonlinear encoder is preserved rather than replaced by a single
affine projection. The saved relaxed archetypal decoder is materialized
using its inference projection:
\[
\begin{aligned}
U_j&=\frac{\max(W_j,0)}{\sum_i\max(W_{ji},0)+10^{-8}},
\qquad R'_j=R_j\min\!\left(1,\frac{1}{\lVert R_j\rVert_2}\right),\\
D&=e^s(UC+R'),\qquad \widehat h=D^\top z.
\end{aligned}
\]
Here \(C\) is the saved set of activation anchors, \(R\) is the relaxation
with unit row-norm bound, and \(s\) is the saved scalar multiplier.
Feature order and dictionary scales are unchanged. Independent row-wise
checks of the materialized dictionary differ by at most
\(8.94\times10^{-8}\); native and source encoder feature values agree
exactly. This frozen nonlinear dictionary supplies the same public
\texttt{BaseSAE.encode/decode} evaluation contract as the other recipes.

The backbone binding is independently checked against Meta's original
\path{dinov2_vitb14_reg4_pretrain.pth}, SHA256
\nolinkurl{73182a088cf94833c94b1666d1c99e02fe87e2007bff57b564fb6206e25dba71},
and source revision
\nolinkurl{7764ea0f912e53c92e82eb78a2a1631e92725fc8}.
All 224 mapped Hugging Face tensors match the original checkpoint exactly;
real-image patch outputs agree to relative L2 error
\(2.83\times10^{-6}\). The native sequence contains 261 positions:
CLS, four registers and 256 raster-ordered patches. The evaluation selects
positions 5--260 after final LayerNorm, so neither CLS nor register
activations enter the spatial maps or patch reconstruction metrics.

\subsubsection{Text selection, metrics and probe controls}

Pile evaluation uses the explicit \texttt{val.jsonl.zst} and
\texttt{test.jsonl.zst} files from
\path{monology/pile-uncopyrighted}, revision
\nolinkurl{3be90335b66f24456a5d6659d9c8d208c0357119}. Stripped-text SHA256
values exclude repeated selected documents across calibration and test.
From the first 4,000 selected unique documents per file, the first sufficient
records with at least 1,024 tokens supply 128 calibration contexts and
512 test contexts. Each context consists of
the first 1,024 tokens of one document; documents are not concatenated.
Tokenization uses no added special tokens. Historical overlap with backbone
or SAE training documents is not established, so this is same-corpus
held-out-file evaluation, rather than a verified unseen-training-data test.

The calibration mean is calculated over all 131,072 calibration positions.
Each test reconstruction uses exactly the same 524,288 native positions.
Squared errors and the fixed-reference denominator are summed globally in
float64. Feature frequencies count \(z_j\ne0\), including signed values;
a feature unused on this split is reported as inactive on this split.
The 64-context fidelity subset is fixed to the first eight test batches of
eight contexts. We score every shifted next-token target, yielding
\(64\times1023=65{,}472\) predictions. If \(L_{\mathrm{orig}}\), \(L_{\mathrm{rec}}\) and \(L_{\mathrm{zero}}\)
are original, reconstructed and native-zero-residual NLL, loss recovered is
\[
R_{\mathrm{loss}}=\frac{L_{\mathrm{zero}}-L_{\mathrm{rec}}}
{L_{\mathrm{zero}}-L_{\mathrm{orig}}}.
\]
The original NLL is 2.3461 nats per prediction and the zero-residual NLL is
11.9161. The zero control sets the selected native residual vectors to zero;
it does not merely zero SAE features while retaining a decoder bias.

AG News uses \path{fancyzhx/ag_news}, revision
\nolinkurl{eb185aade064a813bc0b7f42de02595523103ca4}. For each of four classes,
the official training file supplies 1,000 training and 250 development
documents; the official test file supplies 250 test documents. Selection
uses seed 17 and excludes equal stripped-text hashes across all three
splits. Residual caches use right EOS padding to 128 tokens, with padded
positions excluded from pooling and evaluation. On 32 checked documents,
replacing masked padding values leaves valid activations unchanged.
Singleton versus padded-batch float32 execution has maximum absolute
residual difference 0.00217 and relative L2 difference
\(2.39\times10^{-5}\); these kernel-rounding differences are retained in
the provenance.

Each concept probe averages nonpadding token features within a document.
For a given one-versus-rest class, features are ranked by absolute
point-biserial correlation using training labels only. Training mean and
population standard deviation define scaling. Validation AUROC selects
\(C\in\{0.01,0.1,1,10\}\) for a liblinear logistic classifier; smaller \(C\)
breaks ties. No development or test observation influences feature selection
or scaling. The development split tunes regularization; this is not a
cross-validation estimate. The final metrics use the 1,000 untouched test
documents. Table~\ref{tab:sae-probe-controls} gives the corresponding
controls: 768 raw neurons, 128 training-fitted principal components and a
seed-17 Gaussian projection into 4,096 column-normalized directions.
Each representation receives the same per-class feature budget.

\begin{table}[!htbp]
\centering\footnotesize
\caption{All 18 pretrained language dictionaries. Rows within each family
use source trainer indices 0, 2 and 4. \(\lambda\) is the published sparsity
penalty; JumpReLU settings are training targets, while reported
\(\overline L_0\) is measured on Pile. FVU uses the independent calibration
mean, and \(\Delta\mathrm{NLL}\) is in nats per prediction. AUROC uses one or
five selected features separately for each AG News one-versus-rest class.}
\label{tab:sae-language-all}
\begin{tabular}{llrrrrr}
\toprule
Method & Source setting & \(\overline L_0\) & FVU &
\(\Delta\mathrm{NLL}\) & AUROC@1 & AUROC@5 \\
\midrule
ReLU + L1 & $\lambda=0.012$ & 468.80 & 0.0399 & 0.1145 & 0.9120 & 0.9329 \\
ReLU + L1 & $\lambda=0.020$ & 227.06 & 0.0690 & 0.2493 & 0.9073 & 0.9362 \\
ReLU + L1 & $\lambda=0.040$ & 77.71 & 0.1078 & 0.5796 & 0.8869 & 0.9270 \\
\addlinespace[2pt]
TopK & $k=20$ & 20.00 & 0.1081 & 0.5751 & 0.8412 & 0.9247 \\
TopK & $k=80$ & 80.00 & 0.0659 & 0.2080 & 0.9002 & 0.9318 \\
TopK & $k=320$ & 320.00 & 0.0182 & 0.0341 & 0.9239 & 0.9370 \\
\addlinespace[2pt]
Gated & $\lambda=0.012$ & 464.43 & 0.0063 & 0.0106 & 0.9050 & 0.9351 \\
Gated & $\lambda=0.024$ & 215.58 & 0.0375 & 0.0804 & 0.9050 & 0.9351 \\
Gated & $\lambda=0.060$ & 46.33 & 0.0843 & 0.3219 & 0.9038 & 0.9232 \\
\addlinespace[2pt]
JumpReLU & target $L_0=20$ & 20.25 & 0.1148 & 0.6236 & 0.8685 & 0.9216 \\
JumpReLU & target $L_0=80$ & 80.81 & 0.0657 & 0.1998 & 0.9088 & 0.9294 \\
JumpReLU & target $L_0=320$ & 321.67 & 0.0192 & 0.0331 & 0.9029 & 0.9337 \\
\addlinespace[2pt]
BatchTopK & $k=20$ & 19.84 & 0.1074 & 0.5512 & 0.8671 & 0.9254 \\
BatchTopK & $k=80$ & 79.37 & 0.0648 & 0.1945 & 0.9106 & 0.9289 \\
BatchTopK & $k=320$ & 319.13 & 0.0218 & 0.0380 & 0.9208 & 0.9380 \\
\addlinespace[2pt]
Matryoshka & $k=20$ & 19.88 & 0.1166 & 0.6178 & 0.9065 & 0.9230 \\
Matryoshka & $k=80$ & 79.80 & 0.0736 & 0.2196 & 0.9060 & 0.9294 \\
Matryoshka & $k=320$ & 320.62 & 0.0262 & 0.0455 & 0.8962 & 0.9349 \\
\bottomrule
\end{tabular}
\end{table}

\begin{table}[!htbp]
\centering\footnotesize
\caption{Matched concept-probe controls on the same AG News documents.
The three AUROC columns use one, two or five selected features per class;
F1 uses five. PCA and standardization are fitted on training documents only.}
\label{tab:sae-probe-controls}
\begin{tabular}{lrrrr}
\toprule
Representation & AUROC@1 & AUROC@2 & AUROC@5 & Macro F1@5 \\
\midrule
Raw neurons (768) & 0.8506 & 0.8929 & 0.9249 & 0.8038 \\
PCA (128) & 0.8511 & 0.9105 & 0.9481 & 0.8363 \\
Random projection (4,096) & 0.8773 & 0.9105 & 0.9286 & 0.7952 \\
\bottomrule
\end{tabular}
\end{table}

\subsubsection{Image geometry and feature selection}

The COCO subset contains 500 training and 1,000 validation images selected
by a seed-17 shuffle of sorted official image IDs, before any model outputs
are observed. The annotations archive SHA256 is
\nolinkurl{113a836d90195ee1f884e704da6304dfaaecff1f023f49b6ca93c4aaae470268}.
The four block-11 dictionaries and the block-9 Gated MLP dictionary use
the source ViT-Prisma transform:
bicubic resize of the shorter edge to 224 pixels, with antialiasing, followed
by a \(224\times224\) center crop. Crop offsets use torchvision's rounded
half-difference, including round-half-to-even behavior. CLIP uses its source
RGB channel normalization; DINO uses ImageNet means and standard deviations.
Their patch grids are \(7\times7\) and \(14\times14\), respectively.
Reconstruction excludes CLS and uses all patches: 24,500 CLIP or 98,000 DINO
calibration vectors define each observation's reference mean; 49,000 CLIP or
196,000 DINO validation vectors define the reconstruction metrics. The Gated
MLP dictionary uses its own block-9 reference mean and feature RMS, with
the same 24,500 calibration and 49,000 validation patch positions. Feature
selection and localization also use every patch. Raw maps are bilinearly
resized to the native \(224\times224\) category-box grid for localization,
without per-image contrast normalization.

For the additional CLIP ViT-L/14-336 cohort, the slow processor resizes the
shorter edge to 336 pixels, then applies
a \(336\times336\) center crop with offset
\(\lfloor(s-336)/2\rfloor\) along a resized dimension of size \(s\).
Boxes undergo the identical resize and crop, are clipped to the input
bounds, and are unioned by category. Crowd boxes and boxes lying entirely
outside the crop are excluded. This geometry is verified against processor
pixels, including odd-sized resized dimensions; no square padding is used.

For this cohort, the selected activation has 577 positions: CLS followed by a
\(24\times24\) patch grid. CLS reconstruction uses position 0. Patch
reconstruction uses all 576 spatial positions: 288,000 calibration vectors
define the fixed reference mean and 576,000 validation vectors define FVU
and activity. The spatial evaluation also uses every patch and bilinearly
resizes the measured map to the \(336\times336\) annotation grid,
without per-image min--max normalization. CLS and patch reconstruction
remain separate observations. Table~\ref{tab:sae-native-positions} gives
the native position counts for every backbone.

The ViSAEBench CLIP-B/16 processor uses bicubic shortest-edge resize to
224 and a centered \(224\times224\) crop with floored offsets, producing
\(14\times14\) spatial tokens. DINOv2 uses the author's bicubic
shortest-edge resize to 256, then torchvision's rounded center crop to
224, producing \(16\times16\) patches. Its RGB normalization is ImageNet's.
SigLIP directly resizes to \(224\times224\), without a center crop, with
channel mean and standard deviation 0.5; its grid is \(14\times14\).
Every category box receives the same resize and crop as the actual pixels.
RA-SAE uses the same bicubic resize-to-256, rounded center crop to 224
and ImageNet normalization, with the register backbone's \(16\times16\)
spatial grid. Source crop differences can change the visible region and
eligible pair membership even though the raw COCO image order is shared.

Both Qwen routes use their native fast image processors with minimum and
maximum pixel budgets set to 65,536. Dimensions retain the source aspect
ratio subject to the processor's patch-multiple rounding; no center crop is
added. Qwen2.5 uses 14-pixel patches and spatial merge factor 2. Its
\(h\times w\) processor patch grid becomes an \((h/2)\times(w/2)\)
visual-token grid in decoder block 34. Those positions are identified by
native image-token IDs after chat and image/text merging. Qwen3 uses
16-pixel patches. Block 23 is observed before merge factor 2, and its grouped
patch order is inverted to the native \(h\times w\) raster grid. A small
indexed-grid fixture checks this inverse for scalar and feature-valued
maps. Recovered RGB images are reconstructed from the actual normalized
processor tensors, so plotted grids, pixels and scaled boxes share the same
coordinate frame. Variable image shapes are grouped by native grid and pixel
shape when calling the shared localization pipeline.

For each additional checkpoint, the reference mean and feature RMS
use every captured calibration visual position; reconstruction evaluates
every captured validation visual position. Position counts therefore depend
on native geometry and are retained with the per-model results. These
all-patch measurements use the same reconstruction definition as the
CLIP-L/14-336 patch evaluation above, while retaining each source grid
and observation width.

\begin{table}[!htbp]
\centering\small
\caption{Native spatial observation counts on the fixed 500 calibration and
1,000 validation images. CLS is excluded. The Qwen rows retain their actual
image-dependent grids and observe different stages.}
\label{tab:sae-native-positions}
\begin{tabular}{llrr}
\toprule
Backbone / scope & Native grid & Calibration positions & Validation positions \\
\midrule
CLIP B/32 & $7\times7$ & 24,500 & 49,000 \\
DINO B/16 & $14\times14$ & 98,000 & 196,000 \\
CLIP L/14-336 & $24\times24$ & 288,000 & 576,000 \\
CLIP B/16 & $14\times14$ & 98,000 & 196,000 \\
SigLIP B/16 & $14\times14$ & 98,000 & 196,000 \\
DINOv2 B/14 & $16\times16$ & 128,000 & 256,000 \\
DINOv2 B/14 + registers & $16\times16$ & 128,000 & 256,000 \\
Qwen2.5-VL decoder L34 & Variable, post-merge & 36,921 & 73,965 \\
Qwen3-VL vision L23 & Variable, pre-merge & 109,468 & 219,592 \\
\bottomrule
\end{tabular}
\end{table}

For every dictionary, category \(c\) and feature \(j\), calibration selection maximizes
\[
A_{cj}=\frac{\overline {z^+}_{cj}^{\mathrm{inside}}
                 -\overline {z^+}_{cj}^{\mathrm{outside}}}
 {\sqrt{\mathbb E_{\mathrm{cal}}[z_j^2]}}.
\]
Here \(z_j^+=\max(z_j,0)\) defines positive evidence; the denominator
retains the raw feature RMS, including signed values. Inside and outside
means use category-positive calibration images, with
area-weighted box coverage at patch resolution. The RMS uses all
calibration patches. A category needs at least ten calibration images;
a candidate feature must have positive evidence in at least
\(\max(3,0.1n_c)\) category-positive images and have positive contrast.
The 20 most frequent eligible categories from the CLIP-B/32 calibration
selection define a fixed category list for the expanded cohort. Each new
dictionary selects its own feature on calibration data; categories that do
not meet the same visibility, positive-activity and contrast conditions are
recorded as ineligible rather than filled using test data. Features are
frozen before validation. The four block-11
dictionaries select the same category set and are scored on the same
1,534 validation pairs from 774 source images. The figures use the fixed
person, car and dog categories from this selection. Image--category pairs are pooled for the
headline metrics, so common categories contribute more pairs. The same
feature may associate with more than one category. Table~\ref{tab:sae-localization-all}
reports every selected category for the CLIP ViT-L/14-336 cohort.

Within-image permutation controls independently shuffle each native-grid
feature map using seed 17, preserving feature values and activity while
removing their measured spatial arrangement. Point@1 includes inactive
positive-evidence maps as misses. Activation mass uses \(M_j^+\), is
undefined when that positive channel is zero, and is averaged only over
positive-active pairs, with coverage reported separately. Raw signed
feature activity used for \(\overline L_0\) can therefore differ from
positive spatial response. The target is
a category-box union, not a pixel-level segmentation mask.

For the cross-backbone figures, every source photograph is processed
independently by each backbone's native image transform. Ground-truth boxes
undergo the identical resize and crop, and feature maps retain their native
patch grid. The same category feature is reused across images within each
dictionary. Let \(r_j=\sqrt{\mathbb E_{\mathrm{cal}}[z_j^2]}\) be its
calibration RMS. All displayed maps in this cohort are nonnegative and use \(z_j/r_j\)
on the fixed linear range 0--6, saturating larger values. The raw Qwen2.5
encoder remains unrectified even though its retained COCO coefficients
are nonnegative. No individual test panel is rescaled to its own maximum. The heatmap
is opaque and uses white at zero, then cyan and blue as activation increases.
Its native RGB image is superimposed at opacity 0.40, followed by aligned
orange GT boxes. Feature IDs, raw peaks and calibration values remain in
the rendering records rather than as per-image labels. PDF export keeps the two
image layers separate so image opacity is applied once. Inactive maps remain
inactive.

The method-comparison figure uses the same fixed person photograph,
COCO 9378, for both backbone rows: CLIP-B/32 DataComp and DINO-B/16. Its
columns show the uncropped original followed by ReLU and TopK maps. All
four dictionaries read block-11 residuals. The dictionaries retain their
source-specific activation normalization and sparsity settings, rather than
imposing an equal activity budget.
The compact atlas has four rows: person 9378, person 14439, car 263463 and
dog 532575. Its first column shows uncropped, full-color originals and
original category boxes. Six subsequent columns retain one fixed
visual-encoder dictionary per backbone: Matryoshka for CLIP-L/14-336;
ReLU for DINO-B/16; BatchTopK for SigLIP and DINOv2 without registers;
RA-SAE for DINOv2 with four registers; and SAE++ level-1 for Qwen3-VL-4B
vision layer 23. Qwen3 is observed before patch merging and language
decoding. The two DINOv2 columns use different backbone weights and token
layouts, with register tokens excluded from the RA-SAE raster.
The other evaluated checkpoints, including the distinct Qwen2.5 decoder
observation, remain in the quantitative tables. That inventory spans
observation stages; neither it nor the encoder-only atlas isolates an SAE
algorithm while holding backbone weights and activation hooks fixed.
Native crops can show different extents of the same source photograph.
All four cases and their category features are reused from the completed
evaluation; compact rendering does not rerun inference or choose new features.
RA-SAE's two person maps and its dog map retain their measured zero response
without a per-image text label; the selected car feature remains active.
Rendering records retain
original-image, native-image and map hashes, feature IDs, RMS values,
native grids, aligned boxes and saturated-value fractions.

Sampling intervals use 2,000 paired percentile-bootstrap resamples with seed
17. Language comparisons resample the 64 source documents. The additional
CLIP ViT-L/14-336 localization intervals resample the 774 source images
contributing its 1,534 image--category pairs,
retaining every paired row of a sampled image; category pairs from one image
are not treated as independent observations. The BatchTopK-minus-ReLU NLL
increase has interval \([-0.4192,-0.3569]\) nats, and the near-20-activity
Matryoshka-minus-TopK difference is 0.0427 nats with interval
\([0.0216,0.0635]\). Real-minus-shuffled point@1 differs by 44.0 percentage
points with interval 41.5--46.4; active-only mass differs by 42.1 points with
interval 40.5--43.6. These intervals condition on fixed dictionaries,
feature selection, splits and the recorded shuffle realization. They do not
estimate SAE training-seed variability.

\begin{table}[!htbp]
\centering\footnotesize
\caption{Complete held-out localization results for the CLIP ViT-L/14-336
Matryoshka patch dictionary and its train-selected categories.
\(N\) counts image--category pairs. Coverage, point@1 and active-only mass
inside category-box unions are percentages; inactive maps count as point@1
misses. Feature IDs preserve the author's dictionary order.}
\label{tab:sae-localization-all}
\begin{tabular}{lrrrrr}
\toprule
Category & Feature & \(N\) & Coverage & Point@1 & Mass (active) \\
\midrule
Backpack & 370 & 40 & 92.5 & 62.5 & 42.4 \\
Bench & 3276 & 45 & 68.9 & 60.0 & 77.8 \\
Bottle & 2167 & 79 & 64.6 & 41.8 & 56.8 \\
Bowl & 574 & 56 & 89.3 & 71.4 & 67.4 \\
Car & 2499 & 97 & 100.0 & 75.3 & 52.5 \\
Cell phone & 1195 & 36 & 72.2 & 50.0 & 59.2 \\
Chair & 1848 & 113 & 100.0 & 70.8 & 51.2 \\
Clock & 2105 & 37 & 86.5 & 64.9 & 56.5 \\
Cup & 553 & 71 & 71.8 & 52.1 & 60.8 \\
Dining table & 2557 & 92 & 65.2 & 42.4 & 61.6 \\
Dog & 119 & 45 & 97.8 & 88.9 & 82.2 \\
Handbag & 6720 & 47 & 91.5 & 57.4 & 48.9 \\
Knife & 266 & 35 & 88.6 & 34.3 & 45.5 \\
Motorcycle & 2456 & 25 & 96.0 & 88.0 & 77.3 \\
Person & 420 & 533 & 84.2 & 68.9 & 77.4 \\
Pizza & 115 & 26 & 96.2 & 96.2 & 92.6 \\
Sink & 1420 & 44 & 100.0 & 77.3 & 49.1 \\
Surfboard & 2507 & 31 & 96.8 & 54.8 & 49.5 \\
Tie & 4301 & 35 & 8.6 & 0.0 & 0.0 \\
Truck & 2606 & 47 & 70.2 & 40.4 & 51.5 \\
\bottomrule
\end{tabular}
\end{table}

\subsubsection{Additional CLIP ViT-L/14-336 CLS dictionaries}

Two 16,384-feature CLS dictionaries, BatchTopK and Matryoshka, use the
same CLIP-L/14-336 block-22 backbone and fixed COCO image selections as
the patch study. Their source dictionaries were trained on ImageNet.
These single-vector observations assess representation retention beyond
the spatial patch maps.

\begin{table}[!htbp]
\centering\small
\caption{CLS dictionaries on 1,000 held-out COCO images. FVU uses a
reference mean from 500 calibration images. Cosine is measured at the
final projected image embedding after native CLS reconstruction.}
\label{tab:sae-vision}
\begin{tabular}{llrrrr}
\toprule
Dictionary & Scope & Width & \(\overline L_0\) & FVU & Cosine \\
\midrule
BatchTopK & CLS & 16,384 & 24.35 & 0.2816 & 0.9709 \\
Matryoshka & CLS & 16,384 & 23.76 & 0.2956 & 0.9696 \\
\bottomrule
\end{tabular}
\end{table}

CLS reconstruction retains final projected image-embedding cosine
similarity at 0.9709 for BatchTopK and 0.9696 for Matryoshka. This measures
representation retention rather than classification accuracy.
The all-patch Matryoshka results appear in
Table~\ref{tab:sae-vision-comparison}; its complete category-level
localization results remain in Table~\ref{tab:sae-localization-all}.
Across 1,534 held-out image--category pairs, patch-feature coverage is
83.1\%, point@1 counting inactive maps as misses is 62.5\%, and shuffled
point@1 is 18.5\%. Conditional on positive activity, mass inside the
target boxes is 64.9\%, versus 22.8\% after spatial shuffling.
The dictionary contributes the CLIP-L/14-336 maps in
Figure~\ref{fig:sae-atlas}, using the same
independently selected feature IDs as in this quantitative evaluation.

\subsubsection{Reproducibility}

The experiment scripts in \path{experiments/sae_report_v1} use the public
\texttt{vlmprobe.evaluation.sae} API. Given the pinned raw files and weights,
the staged replay is:
\begin{Verbatim}[fontsize=\scriptsize]
python experiments/sae_report_v1/prepare_data.py --text --coco
python experiments/sae_report_v1/capture_language.py
python experiments/sae_report_v1/verify_language_cache.py
python experiments/sae_report_v1/evaluate_language.py
python experiments/sae_report_v1/capture_vision.py
python experiments/sae_report_v1/evaluate_vision.py
python experiments/sae_report_v1/summarize_results.py
python experiments/sae_report_v1/estimate_uncertainty.py --resamples 2000 --seed 17
\end{Verbatim}

The cross-backbone comparison is maintained in
\path{experiments/sae_visual_comparison_v2}. Its commands preserve the same
raw COCO selection and consume pinned weights:
\begin{Verbatim}[fontsize=\scriptsize]
python experiments/sae_visual_comparison_v2/download_sources.py
python experiments/sae_visual_comparison_v2/run_experiment.py --model clip
python experiments/sae_visual_comparison_v2/run_experiment.py --model dino
python experiments/sae_visual_comparison_v2/collect_results.py
python experiments/sae_visual_comparison_v2/render_figures.py
\end{Verbatim}
The checkpoint verifier additionally accepts real native activation tensors
through \path{--activation-dir}; the backbone preflight and the real-input
conversion checks are retained with the measured results. The rendering
stage reads measured maps and predeclared cases only; it does not load a
model, select a feature or reorder images by a test score.

The expanded spatial evaluation uses
\path{experiments/sae_visual_comparison_v3}, retaining the same raw COCO
selection and fixed categories. With the pinned source weights in place,
the additional model capture, calibration and evaluation stages are:
\begin{Verbatim}[fontsize=\scriptsize]
python experiments/sae_visual_comparison_v3/run_new_models.py --model all --phase all
python experiments/sae_visual_comparison_v3/evaluate_matryoshka_allpatch.py --stage all --device cuda:0
python experiments/sae_visual_comparison_v3/run_new_models.py --phase collect
\end{Verbatim}
Source equations and input geometry are verified before full capture. The
native artifacts preserve each source recipe, including signed value-ranked
TopK and the frozen SAE++ level-1 threshold. The additional Gated dictionary uses its independently verified MLP hook:
\begin{Verbatim}[fontsize=\scriptsize]
python experiments/sae_visual_comparison_v4/run_gated.py --phase all --device cuda:0
\end{Verbatim}
The RA-SAE route preserves the nonlinear encoder and register-token scope:
\begin{Verbatim}[fontsize=\scriptsize]
python experiments/sae_visual_comparison_v4/ra_sae.py download
python experiments/sae_visual_comparison_v4/ra_sae.py convert --device cuda:0
python experiments/sae_visual_comparison_v4/ra_sae.py preflight --device cuda:0
python experiments/sae_visual_comparison_v4/run_rasae.py all --device cuda:0
\end{Verbatim}
The completed source-bound results are collected once. Compact rendering
reads the existing cache without rerunning inference or selecting features.
The matched-backbone figure retains the version-6 artifact; the
visual-encoder-only atlas uses version-7 selection and rendering:
\begin{Verbatim}[fontsize=\scriptsize]
python experiments/sae_visual_comparison_v4/collect_results.py
SAE_V4_OUTPUT=/bigdata/gexin/vlm_probing/outputs/sae_visual_comparison_v4
python experiments/sae_visual_comparison_v6/render_figures.py \
  --input "$SAE_V4_OUTPUT/visual_cases.pt" \
  --layout experiments/sae_visual_comparison_v6/layout.json \
  --case-selection /bigdata/gexin/vlm_probing/outputs/sae_visual_comparison_v2/display_selection.json \
  --output-dir /bigdata/gexin/vlm_probing/outputs/sae_visual_comparison_v6/figures
python experiments/sae_visual_comparison_v7/render_figures.py \
  --input "$SAE_V4_OUTPUT/visual_cases.pt" \
  --layout experiments/sae_visual_comparison_v7/layout.json \
  --case-selection /bigdata/gexin/vlm_probing/outputs/sae_visual_comparison_v2/display_selection.json \
  --output-dir /bigdata/gexin/vlm_probing/outputs/sae_visual_comparison_v7/figures
python experiments/sae_visual_comparison_v7/validate_figures.py
\end{Verbatim}
Final rendering reads only measured maps and predeclared display cases,
rejects missing selected maps, and retains inactive measured responses.
The atlas validator verifies all six visual-encoder bindings against the
frozen source configurations, excludes the Qwen2.5 decoder dictionary and
checks every map, image, aligned box and original photograph against the
recorded payload.

Data, weights, native artifacts and generated outputs are stored under the
existing \path{/bigdata/gexin/vlm_probing} hierarchy. The saved records
include source revisions, checksums, raw sample IDs, selected masks and
features, configuration, per-example scores and aggregate CSV/JSON results.
The evaluation guide and live notebooks provide complete download-to-plot
examples with smaller teaching subsets. Those tutorial scores are separate
from the report measurements above. All reported checkpoints are fixed;
comparisons do not estimate SAE training-seed variability.

\subsection{Pretrained multimodal decoder SAE protocol}
\label{sec:sae-decoder-details}

The decoder case study uses five published dictionaries with their original
weights and inference equations. The LLaVA-Llama3 dictionary accompanies
\textcite{multimodal_sae}; its training pipeline encodes complete multimodal
conversations and flattens all decoder positions. The Qwen2.5 author reports
498,000 LLaVA-NeXT training samples without specifying an exact token mask.
The Mistral dictionary comes from \textcite{line_of_sight}. Qwen3.5's
checkpoint-specific configuration records LMSYS-Chat-1M text conversations,
700 million training tokens and exclusion of special tokens
\autocite{qwen35_sae}. PaliGemma's MMDiff release trains on text positions
of VQAv2 validation image/question forwards \autocite{mmdiff}. The latter
two dictionaries are transferred to visual positions here. In particular,
the PaliGemma cases are not an independent held-out VQAv2 evaluation.
Its released training script uses an \texttt{answer en} task prefix;
this study uses the native plain question plus the common instruction below.

\begin{table}[!htbp]
\centering\small
\caption{Native pretrained decoder observations. All widths and sparsity
budgets come from the published weights/configurations, not new training.}
\begin{tabularx}{\linewidth}{@{}Xlrrl@{}}
\toprule
VLM & Boundary & Hidden & Features & Source activation \\
\midrule
Qwen2.5-VL-3B & L34 output & 2,048 & 65,536 & Raw TopK, $k=100$ \\
Qwen3.5-4B & L27 output & 2,560 & 65,536 & JumpReLU export \\
LLaVA-NeXT Llama3-8B & L24 output & 4,096 & 131,072 & ReLU + TopK, $k=256$ \\
LLaVA-v1.6-Mistral-7B & L24 input & 4,096 & 32,768 & ReLU, scale 8 \\
PaliGemma2-3B mix448 & L23 output & 2,304 & 16,384 & JumpReLU \\
\bottomrule
\end{tabularx}
\end{table}

All layer indices are zero-based. The Mistral observation is the input to
block 24, implemented at block 23's residual output; the LLaVA-Llama3
dictionary instead observes block 24's output. These leave 8 and 7 complete
decoder blocks downstream, respectively; Qwen2.5, Qwen3.5 and PaliGemma
leave 1, 4 and 2. No common layer number or relative depth is imposed.

Each run binds the following pinned backbone and dictionary releases:
\begin{itemize}
\item Qwen2.5-VL:
\href{https://huggingface.co/Qwen/Qwen2.5-VL-3B-Instruct/tree/66285546d2b821cf421d4f5eb2576359d3770cd3}{Qwen2.5-VL-3B-Instruct}
and \href{https://huggingface.co/BeEnough/SAE-Res-Qwen2.5-VL-3B-Instruct-W64K-L34-K100/tree/1a138768b0b74ea6844399f43f7a7ad1a69c0bff}{BeEnough's L34 TopK dictionary}.
\item Qwen3.5:
\href{https://huggingface.co/Qwen/Qwen3.5-4B/tree/851bf6e806efd8d0a36b00ddf55e13ccb7b8cd0a}{Qwen3.5-4B}
and \href{https://huggingface.co/decoderesearch/qwen-3.5-saes/tree/3a67187f0082a25634aa959092bb834957aeda02/qwen-3.5-4b/btk-mat-layer-27-k-100}{Decode Research's L27 Matryoshka BatchTopK dictionary}.
\item LLaVA-Llama3:
\href{https://huggingface.co/llava-hf/llama3-llava-next-8b-hf/tree/b041c0d0ea0dd0196d147206c210c8d1752fc2da}{LLaVA-NeXT Llama3-8B}
and \href{https://huggingface.co/lmms-lab/llama3-llava-next-8b-hf-sae-131k/tree/8c9f77711303b585e7ea589b6ccedb0fd22712b0}{LMMs-Lab's L24 dictionary}.
\item LLaVA-Mistral:
\href{https://huggingface.co/llava-hf/llava-v1.6-mistral-7b-hf/tree/2424fdd47412fccc66d91719126b420e9fbd7065}{LLaVA-v1.6-Mistral-7B}
and \href{https://huggingface.co/flybamboo/vlm-saes/tree/5bede0c48d5212a71b70140f4ce26f96449d647d}{Line-of-Sight weights released in vlm-saes}, file
\path{sae_final_layer24_8x_norm13.996978.pkl}.
\item PaliGemma:
\href{https://huggingface.co/google/paligemma2-3b-mix-448/tree/1406c92ec87d32cc6b983239278901b904ba7a51}{PaliGemma2-3B mix448}
and \href{https://huggingface.co/OX-PIXL/MMDiff_SAEs/tree/c3c816f225488e2bf5e015516281d4330015a23a/paligemma2/jumprelu_text_only}{MMDiff's text-position JumpReLU dictionary}, file
\path{text-only_layer_23.pt}.
\end{itemize}
Download manifests verify file sizes and source LFS SHA256 values. Frozen
contracts bind the local weight files, code snapshots, questions, images and
answer-format policies; checkpoints are never silently substituted.

\paragraph*{Native encoding and scale.}
Qwen2.5 retains signed coefficients selected by raw-value TopK; LLaVA-Llama3
centers by decoder bias, applies ReLU and retains its largest 256 values.
For Qwen3.5, the saved inference export computes
\(a=(h-b_d)W_e+b_e\), \(z=\max(a,0)\mathbf{1}[a>\theta]\),
then \(D(z)=zW_d+b_d\). It uses learned per-feature thresholds rather
than reapplying the training budget of 100. PaliGemma uses the same
thresholded form with \(a=hW_e+b_e\): no decoder-bias centering or
activation normalization is introduced.

Mistral follows the original
\href{https://github.com/multimodal-interpretability/multimodal-saes/blob/8a466949994700be3eb67e3364375083d7697562/sae/evaluation/eval_sae_damage.py}{published reconstruction-evaluation recipe}:
\(z=\mathrm{ReLU}(\mathrm{fc1}(8h))\) and
\(D(z)=\mathrm{fc2}(z)/8\), retaining every positive coordinate.
The later SAVE code uses a different capture boundary and a hard TopK
filter; those choices are not substituted for this source recipe. Alternative
public scale/boundary conventions are retained as preflight diagnostics,
without choosing among them using QA outcomes.

\paragraph*{Question and box binding.}
The eight fixed questions and canonical/foil strings appear in
Table~\ref{tab:sae-decoder-questions}. Original VQAv2 validation questions
join COCO2017 images by integer image ID to their COCO2014 counterparts.
The dog-tag shape question is 532575002; the dog-breed question is
532575001. The tag-shape question has ten of ten answers ``heart.'' The
breed question has eight ``german shepherd'' answers, one ``shepherd''
and one ``doberman''; that disagreement is retained. Official question
and annotation ZIP archives are downloaded
with byte-count, SHA256 and ZIP CRC checks. Constructed questions are
visually checked investigator references rather than crowd annotations.
Every model input appends the same instruction to the unmodified question:
\begin{quote}\small
Answer with a short phrase in lowercase only. Do not write a full sentence.
\end{quote}
The image remains unmarked. No reference answer, foil or box coordinate is
given to free generation. Greedy generation has a 32-new-token ceiling.
The canonical and foil are separately appended as answer content to the
same scoring prefix; neither sequence includes an EOS/EOT
in the scored content.

The score uses a fixed answer surface established from all eight native raw
generations per model. Both Qwen models and PaliGemma start directly with
lowercase content. Both LLaVA variants start with an initial capital; only
LLaVA-Llama3 emits one preceding newline (token 198), which is shared context
and is not scored. Mistral has no extra structural prefix. A three-surface
audit retains lowercase, capitalized and leading-space lowercase scores
before one rule is frozen for each route. This is a format check, not a
selection of the highest-scoring spelling or a separate rule per question.
Scores refer to these declared continuations rather than an invariant
semantic answer posterior.

PaliGemma's native processor constructs a bidirectional image/question
prefix with token type 0. Appended answer tokens have type 1 and use its
causal suffix mask. A native tiny-model check verifies that changing a later
answer leaves prefix states and earlier answer predictions unchanged;
incorrectly retaining type 0 leaks that answer into image states. Changing
the question can legitimately alter PaliGemma's image-prefix features. Both
candidates use the identical question and attention convention.

\begin{table}[!htbp]
\centering\small
\caption{Fixed decoder questions. C denotes an investigator-constructed
question; V denotes original VQAv2. The selected COCO instance ID binds
all feature and pixel controls for that question.}
\label{tab:sae-decoder-questions}
\begin{tabularx}{\linewidth}{rXlll}
\toprule
Image & Question & Canonical / foil & Instance & Source \\
\midrule
14439 & What color is the jacket worn by the foreground child to the right of the large kite? & black / white & 1733380 & C \\
14439 & What color are the trousers worn by the foreground child to the right of the large kite? & blue / red & 1733380 & C \\
263463 & What part of the car surrounds the dog's reflection? & side mirror / windshield & 145584 & C \\
263463 & What color is the frame of the car's side mirror in the foreground? & black / white & 145584 & C \\
263463 & What animal is reflected in the car's side mirror? & dog / cat & 10416 & C \\
532575 & What shape is the dog tag? & heart / circle & 4318 & V \\
532575 & What kind of dog is this? & german shepherd / poodle & 4318 & V \\
532575 & What color is the heart-shaped tag on the dog collar? & red / blue & 4318 & C \\
\bottomrule
\end{tabularx}
\end{table}

\paragraph*{Answer score and sparse attribution.}
For image \(I\) and question \(Q\), let \(A\) and \(F\) denote the fixed
reference and foil answer-token sequences. For any candidate \(B\),
\(|B|\) counts its answer-content tokens.
For the common image/question scoring prefix \(P(I,Q)\), the candidate
answer score and reference-versus-foil contrast are
\[
\ell(B)=\frac{1}{|B|}\sum_{t=1}^{|B|}
 \log p\bigl(b_t\mid P(I,Q),b_{<t}\bigr),
\qquad S=\ell(A)-\ell(F).
\]
Only answer-content tokens enter the score; role markers, padding and
end-of-turn tokens are excluded. Each answer token is scored at its
preceding predictor position. Token counts, tokenwise log probabilities
and summed scores are retained alongside the mean-token contrast.
Length normalization makes \(S\) a contrast of mean log probabilities,
rather than complete-answer log odds. Reference continuations used for
scoring are not evidence that the model freely generated that answer.

Let \(z^B_{pf}=E(h^B_p)_f\) be the frozen source encoding at visual patch
position \(p\) in the pass for candidate \(B\). An identity-valued proxy
allows differentiation along dictionary directions without changing the
unmodified forward. The signed feature-direction contrast and displayed
map are
\[
 C^B_{pf}=z^B_{pf}\frac{\partial\ell(B)}{\partial z^B_{pf}},
 \qquad C_{pf}=C^A_{pf}-C^F_{pf},
 \qquad M_{p,\mathcal J}^{+}=\max\!\left(\sum_{f\in\mathcal J} C_{pf},0\right).
\]
For each model and case, \(\mathcal J\) contains up to eight feature IDs
with the largest positive global contributions \(\sum_p C_{pf}\).
This same-example, objective-selected group is exploratory; object boxes
never select it. Figure~\ref{fig:sae-decoder-qa} displays this exact group,
which is also used for the measured deletion below. Canonical and foil
continuations share the same scoring prefix but are differentiated
separately. Signed contributions, negative evidence and the full-dictionary
aggregate remain archived. With an affine decoder, the full-dictionary
aggregate reduces to attribution along the reconstruction vector excluding
decoder bias; the selected-group map retains its sparse decomposition.

These are local gradient attributions along SAE directions, rather than raw
feature activations or attention weights. In Qwen and LLaVA, causal
image-prefix states cannot read the later question; the downstream answer
objective provides question conditioning. PaliGemma's bidirectional prefix
can already encode the question, with answer leakage prevented by the
causal suffix mask described above. Contextualized grid positions mix
visual and decoder information: a peak is a token location, not an isolated
pixel receptive field or an object-segmentation mask.

\paragraph*{Geometry and numerical validation.}
Both Qwen models use their native fast image processor with minimum and maximum pixel
budgets both set to 65,536. The post-merge decoder grid is reconstructed
from \texttt{image\_grid\_thw}; displayed pixels are recovered by inverting
native patch packing and normalization, using their respective patch and
merge sizes. PaliGemma uses its native $448\times448$ RGB processor and
$32\times32$ patch grid. Both LLaVA variants use native AnyRes
processors. The retained cases produce a $30\times48$ or $32\times48$
high-resolution spatial grid; the 576-token low-resolution base crop and
30 or 32 newline tokens are excluded only from the displayed/intervened
spatial branch. Original pixel coordinates pass through the actual
resize, padding and unpadding affine. Native source packing is independently
checked on five aspect ratios using coordinate-valued features. For display,
RGB images, maps and boxes share a final affine resize to uniform tiles;
native forward inputs and measurements remain unchanged.

The software environment uses PyTorch 2.6.0 (CUDA 12.4), Transformers 5.3.0
and Accelerate 1.15.0. All computations use frozen FP32 models with eager attention. LLaVA-Llama3 streams
the weights of decoder blocks 0--23 from CPU storage to the GPU for execution;
the observation block and downstream blocks remain on the GPU. This reduces
resident weight memory without changing computation precision or the SAE
equation. The three added routes run fully resident on separate GPUs.
Native hook-output device checks verify GPU execution.
The derivative
is computed along the affine dictionary directions at active source
coordinates, keeping inactive coordinates fixed. Canonical and foil passes
retain their own native sparse coordinates. Their signed contributions are
computed separately and subtracted after aggregation, preserving occasional
TopK boundary changes from FP32 sequence-shape rounding. A separate equal-length
masked-padding check yields identical visual-prefix states; the numerical
guard records both maximum absolute difference and relative L2 difference.
Its fixed acceptance bounds are $10^{-2}$ and $10^{-5}$, respectively;
exceeding either rejects the case rather than replacing its source values.
An identity-valued gradient
proxy leaves forward hidden states bitwise unchanged. Source encoder/decoder
equations, a sparse decoder's dense-equivalent forward and VJP, candidate
token shifts and the first question's central finite difference are checked
before accepting results. No weights, thresholds or normalization statistics
are updated.
Canonical/foil prefix consistency, exact no-op candidate scores and the first
question's feature-scaling finite difference with $\epsilon=0.01$ are
checked for every route. All forty accepted cases and the five directional
checks pass. Full per-case reconstruction and perturbation controls remain
separate from these numerical execution checks.

Table~\ref{tab:sae-decoder-fidelity} evaluates reconstruction at the actual
formal visual positions. Its FVU denominator centers hidden states across
positions within that example, whereas the earlier aggregate SAE evaluation
uses a training-only reference mean. Neither statistic measures the size of
a dictionary-direction intervention.

\begin{table}[!htbp]
\centering\fontsize{8.5}{10.5}\selectfont
\setlength{\tabcolsep}{4pt}
\caption{Dictionary fidelity at the image positions used in the illustrated decoder analyses. Values use the same formal image, question and answer-format instruction as the interventions. FVU is position-centered within the canonical pass; relative error is \(\|D(z)-h\|_2/\|h\|_2\), and mean \(L_0\) counts active coordinates per image position.}
\label{tab:sae-decoder-fidelity}
\begin{tabularx}{\linewidth}{@{}lXrrr@{}}
\toprule
Model & Question & FVU & Relative error & Mean \(L_0\) \\
\midrule
Qwen2.5-VL-3B & Child jacket & 0.225 & 0.253 & 100.0 \\
Qwen2.5-VL-3B & Mirror part & 0.228 & 0.256 & 100.0 \\
Qwen2.5-VL-3B & Tag shape\(^{\dagger}\) & 0.221 & 0.254 & 100.0 \\
Qwen3.5-4B & Child jacket & 0.455 & 0.549 & 124.7 \\
Qwen3.5-4B & Mirror part & 0.489 & 0.563 & 118.0 \\
Qwen3.5-4B & Tag shape\(^{\dagger}\) & 0.407 & 0.534 & 124.9 \\
LLaVA-NeXT Llama3-8B & Child jacket & 0.106 & 0.273 & 256.0 \\
LLaVA-NeXT Llama3-8B & Mirror part & 0.105 & 0.275 & 256.0 \\
LLaVA-NeXT Llama3-8B & Tag shape\(^{\dagger}\) & 0.111 & 0.281 & 256.0 \\
LLaVA-NeXT Mistral-7B & Child jacket & 0.357 & 0.496 & 36.9 \\
LLaVA-NeXT Mistral-7B & Mirror part & 0.373 & 0.516 & 39.3 \\
LLaVA-NeXT Mistral-7B & Tag shape\(^{\dagger}\) & 0.385 & 0.518 & 36.0 \\
PaliGemma2-3B Mix-448 & Child jacket & 2.341 & 1.336 & 176.0 \\
PaliGemma2-3B Mix-448 & Mirror part & 2.364 & 1.363 & 181.4 \\
PaliGemma2-3B Mix-448 & Tag shape\(^{\dagger}\) & 3.040 & 1.488 & 183.1 \\
\bottomrule
\end{tabularx}
\end{table}

\paragraph*{Feature and pixel intervention protocol.}
For an edited sparse vector \(z'\), the residual-preserving intervention is
\[
 h'=h+D(z')-D(z).
\]
The native decoder \(D\) returns raw activation units. This update retains
the original reconstruction error. We remove group \(\mathcal J\) at
every valid spatial image-token position. The targeted effect
\(\Delta_8=S_{\rm base}-S_{\rm ablated}\) measures the resulting change
in the fixed answer contrast. At these late hooks, visual information
already transferred into text positions remains intact; deletion tests
the remaining downstream routes from image-token states.

Three seeded random controls sample active global feature IDs outside the
targeted set when enough exist, using seed \(17+\)question index. Their
mean score drop is \(\overline{\Delta}_{\rm rnd}\). The groups match the
number of feature IDs, not their edited-coordinate counts or hidden-state
perturbation norms. No-op and full-reconstruction controls are retained,
and all coordinate counts and perturbation norms are recorded.
Table~\ref{tab:sae-decoder-all-deletions} reports \(S\), \(\Delta_8\)
and \(\overline{\Delta}_{\rm rnd}\) for all eight questions and five
configurations.

Pixel controls fill the target COCO box with the rounded mean RGB of the
original image and compare three same-area translations. The translated
masks retain the integer-pixel size of the clipped target box; disjoint
placements are preferred when feasible. The large car and dog boxes cover
approximately 70.8\% and 65.1\% of their images, so equal-area translations
necessarily overlap the target box. These are not background-only controls
and cannot establish object-exclusive causal dependence. The dog tag has
no separate COCO annotation and uses the whole-dog box.

\paragraph*{Observed local sensitivity.}
For the jacket question, deleting the selected group changes the contrast
from 0.649 to \(-0.252\) in LLaVA-Mistral and from 0.083 to
\(-0.703\) in PaliGemma. Both nevertheless freely generate ``brown''
(with native capitalization), rather than the scored reference ``black.''
Conversely, Qwen2.5 generates ``black'' for the trousers question despite
a positive blue-versus-red contrast. These outcomes separate sensitivity
of a fixed answer pair from unrestricted generation
(Table~\ref{tab:sae-decoder-generated}).

Local gradient attribution and finite deletion can differ in magnitude:
Qwen2.5's tag-shape group has signed first-order contribution
0.261, compared with a measured score drop of 0.572.

The transferred dictionaries also differ in visual reconstruction. Qwen3.5
has within-image FVU 0.407--0.489 on the three pictured inputs, whereas
PaliGemma's FVU is 1.814--3.040 across the eight questions. The latter
exceeds the within-image variance baseline, despite passing the native
source-equation checks (Table~\ref{tab:sae-decoder-fidelity}). Its nonzero attribution and deletion effects
therefore do not establish a faithful sparse decomposition of image states.

Random-ID controls match feature count but have much smaller perturbation
budgets. For LLaVA-Llama3's jacket case, selected deletion has
\(\|\delta h\|_2=160.62\), versus a random-set mean of 5.82; their small
score drops therefore do not establish semantic specificity.
Reconstruction and pixel controls appear in
Table~\ref{tab:sae-decoder-controls}; coordinate counts and perturbation
norms appear in Table~\ref{tab:sae-decoder-budgets}.

\paragraph*{Reproduction and saved evidence.}
The original two routes reside in
\texttt{experiments/sae\_decoder\_qa\_v1/}; the three added routes are in
\texttt{experiments/sae\_decoder\_qa\_v2/}. The latter reuse the frozen v1
scoring and intervention engine through an explicit answer-format adapter.
Native route modules load the pinned checkpoints. Completed questions resume
only after the full source/data/options contract is checked. For example:
\begin{Shaded}
\begin{Highlighting}[]
python experiments/sae_decoder_qa_v1/prepare_cases.py \
  --coco-dir /path/to/coco/val2017 \
  --output outputs/decoder_qa/cases.json
python experiments/sae_decoder_qa_v1/run_decoder_qa.py \
  --route qwen --cases outputs/decoder_qa/cases.json \
  --device cuda:0 --output outputs/decoder_qa/runs/qwen_native_format_v2
python experiments/sae_decoder_qa_v1/run_decoder_qa.py \
  --route llava --cases outputs/decoder_qa/cases.json \
  --device cuda:0 --output outputs/decoder_qa/runs/llava_native_format_v2
PYTHONPATH=src python experiments/sae_decoder_qa_v2/score_format_probe.py \
  --route paligemma --cases outputs/decoder_qa/cases.json \
  --device cuda:0 --output outputs/decoder_qa/format/paligemma
PYTHONPATH=src python experiments/sae_decoder_qa_v2/run_decoder_qa.py \
  --route paligemma --cases outputs/decoder_qa/cases.json \
  --scoring-policy experiments/sae_decoder_qa_v2/policies/paligemma.json \
  --device cuda:0 --output outputs/decoder_qa/runs/paligemma_native_v2
\end{Highlighting}
\end{Shaded}
Each question saves the complete native prompt, tokenwise scores, generated
answer, per-coordinate signed attribution, predictor-position features,
reconstruction and feature-intervention effects, mask geometry and source
hashes. Weak or negative effects are retained. Spatial plots normalize each
positive net map separately and therefore compare arrangement rather than
absolute evidence magnitude. The complete signed values remain in the
saved tensors.

\begingroup
\fontsize{8}{10}\selectfont
\setlength{\tabcolsep}{3pt}
\renewcommand{\arraystretch}{1.00}
\begin{longtable}{@{}p{.25\linewidth}p{.24\linewidth}rrr@{}}
\caption{Decoder feature deletions across eight fixed questions and five model configurations. \(\dagger\) marks original VQAv2 questions; other questions are constructed.}\label{tab:sae-decoder-all-deletions}\\
\toprule
Model & Question & \(S\) & \(\Delta_8\) & \(\overline{\Delta}_{\rm rnd}\) \\
\midrule
\endfirsthead
\toprule
Model & Question & \(S\) & \(\Delta_8\) & \(\overline{\Delta}_{\rm rnd}\) \\
\midrule
\endhead
\midrule
\multicolumn{5}{r}{Continued on next page}\\
\endfoot
\bottomrule
\endlastfoot
Qwen2.5-VL-3B & Child jacket & 4.972 & 0.008 & -0.0001 \\
Qwen2.5-VL-3B & Child trousers & 4.314 & 0.787 & 0.0006 \\
Qwen2.5-VL-3B & Mirror part & 1.982 & 0.027 & -0.0002 \\
Qwen2.5-VL-3B & Mirror frame & 5.274 & 0.713 & -0.0001 \\
Qwen2.5-VL-3B & Reflected animal & 5.725 & 0.229 & 0.0005 \\
Qwen2.5-VL-3B & Tag shape\(^{\dagger}\) & 2.452 & 0.572 & 0.0004 \\
Qwen2.5-VL-3B & Dog breed\(^{\dagger}\) & 5.820 & 0.107 & 0.0001 \\
Qwen2.5-VL-3B & Tag color & 5.177 & 0.024 & -0.0001 \\
Qwen3.5-4B & Child jacket & 5.601 & 0.217 & -0.0020 \\
Qwen3.5-4B & Child trousers & 4.981 & 0.126 & -0.0018 \\
Qwen3.5-4B & Mirror part & 0.621 & 0.209 & -0.0009 \\
Qwen3.5-4B & Mirror frame & 3.951 & 0.067 & 0.0008 \\
Qwen3.5-4B & Reflected animal & 3.093 & 0.201 & -0.0017 \\
Qwen3.5-4B & Tag shape\(^{\dagger}\) & 4.294 & 0.084 & -0.0018 \\
Qwen3.5-4B & Dog breed\(^{\dagger}\) & 5.037 & 0.215 & 0.0033 \\
Qwen3.5-4B & Tag color & 7.379 & 0.122 & -0.0022 \\
LLaVA-NeXT Llama3-8B & Child jacket & 0.334 & 0.551 & 0.0006 \\
LLaVA-NeXT Llama3-8B & Child trousers & 7.405 & 4.868 & 0.0003 \\
LLaVA-NeXT Llama3-8B & Mirror part & 2.132 & 0.230 & 0.0002 \\
LLaVA-NeXT Llama3-8B & Mirror frame & 5.843 & 0.278 & -0.0016 \\
LLaVA-NeXT Llama3-8B & Reflected animal & 8.339 & 0.419 & 0.0001 \\
LLaVA-NeXT Llama3-8B & Tag shape\(^{\dagger}\) & 7.207 & 0.243 & -0.0005 \\
LLaVA-NeXT Llama3-8B & Dog breed\(^{\dagger}\) & 5.811 & 0.114 & -0.0020 \\
LLaVA-NeXT Llama3-8B & Tag color & 9.743 & 0.604 & 0.0066 \\
LLaVA-NeXT Mistral-7B & Child jacket & 0.649 & 0.901 & -0.0001 \\
LLaVA-NeXT Mistral-7B & Child trousers & 7.208 & 0.362 & -0.0002 \\
LLaVA-NeXT Mistral-7B & Mirror part & 1.233 & 0.037 & 0.0000 \\
LLaVA-NeXT Mistral-7B & Mirror frame & 5.902 & 0.104 & 0.0000 \\
LLaVA-NeXT Mistral-7B & Reflected animal & 7.742 & 0.372 & 0.0001 \\
LLaVA-NeXT Mistral-7B & Tag shape\(^{\dagger}\) & 6.936 & 0.164 & -0.0027 \\
LLaVA-NeXT Mistral-7B & Dog breed\(^{\dagger}\) & 4.908 & 0.135 & 0.0000 \\
LLaVA-NeXT Mistral-7B & Tag color & 7.776 & 0.534 & -0.0019 \\
PaliGemma2-3B Mix-448 & Child jacket & 0.083 & 0.786 & -0.0017 \\
PaliGemma2-3B Mix-448 & Child trousers & 6.306 & 1.352 & -0.0010 \\
PaliGemma2-3B Mix-448 & Mirror part & 1.979 & 0.609 & 0.0000 \\
PaliGemma2-3B Mix-448 & Mirror frame & 5.261 & 1.247 & 0.0013 \\
PaliGemma2-3B Mix-448 & Reflected animal & 8.481 & 0.957 & 0.0013 \\
PaliGemma2-3B Mix-448 & Tag shape\(^{\dagger}\) & 8.743 & 1.086 & 0.0036 \\
PaliGemma2-3B Mix-448 & Dog breed\(^{\dagger}\) & 4.390 & 0.359 & 0.0006 \\
PaliGemma2-3B Mix-448 & Tag color & 8.524 & 1.484 & 0.0011 \\
\end{longtable}
\endgroup
\begingroup
\fontsize{8}{10}\selectfont
\setlength{\tabcolsep}{3pt}
\renewcommand{\arraystretch}{1.00}
\begin{longtable}{@{}p{.19\linewidth}*{5}{p{.14\linewidth}}@{}}
\caption{Greedy generated content from the original image and question. Display whitespace is trimmed; candidate scoring is separate.}\label{tab:sae-decoder-generated}\\
\toprule
Question & Qwen2.5-VL & Qwen3.5 & LLaVA Llama3 & LLaVA Mistral & PaliGemma \\
\midrule
\endfirsthead
\toprule
Question & Qwen2.5-VL & Qwen3.5 & LLaVA Llama3 & LLaVA Mistral & PaliGemma \\
\midrule
\endhead
\midrule
\multicolumn{6}{r}{Continued on next page}\\
\endfoot
\bottomrule
\endlastfoot
Child jacket & black & black & Brown & Brown & brown \\
Child trousers & black & blue & Blue & Blue & blue \\
Mirror part & mirror & mirror & Mirror & Mirror & mirror \\
Mirror frame & black & black & Black & Black & black \\
Reflected animal & dog & dog & Dog & Dog & dog \\
Tag shape\(^{\dagger}\) & heart & heart & Heart & Heart & heart \\
Dog breed\(^{\dagger}\) & german shepherd & german shepherd & German shepherd & German shepherd & german shepherd \\
Tag color & red & red & Red & Red & red \\
\end{longtable}
\endgroup
\begingroup
\fontsize{8}{10}\selectfont
\setlength{\tabcolsep}{3pt}
\renewcommand{\arraystretch}{1.00}
\begin{longtable}{@{}p{.22\linewidth}p{.17\linewidth}rrrrr@{}}
\caption{Reconstruction and pixel controls. FVU is centered over canonical-pass image positions within each example. Score drops compare full reconstruction, target-box masking and the mean of three same-area translations. Max overlap is the largest translated-mask fraction overlapping the target.}\label{tab:sae-decoder-controls}\\
\toprule
Model & Question & FVU & \(\Delta_{\rm rec}\) & \(\Delta_{\rm box}\) & \(\overline{\Delta}_{\rm px}\) & Max overlap \\
\midrule
\endfirsthead
\toprule
Model & Question & FVU & \(\Delta_{\rm rec}\) & \(\Delta_{\rm box}\) & \(\overline{\Delta}_{\rm px}\) & Max overlap \\
\midrule
\endhead
\midrule
\multicolumn{7}{r}{Continued on next page}\\
\endfoot
\bottomrule
\endlastfoot
Qwen2.5-VL-3B & Child jacket & 0.225 & 0.001 & 2.766 & -0.200 & 0.000 \\
Qwen2.5-VL-3B & Child trousers & 0.225 & 0.006 & 0.789 & 0.468 & 0.000 \\
Qwen2.5-VL-3B & Mirror part & 0.228 & 0.010 & 2.169 & 2.234 & 0.926 \\
Qwen2.5-VL-3B & Mirror frame & 0.228 & -0.011 & 1.808 & 1.386 & 0.854 \\
Qwen2.5-VL-3B & Reflected animal & 0.228 & -0.021 & 4.627 & 0.229 & 0.000 \\
Qwen2.5-VL-3B & Tag shape\(^{\dagger}\) & 0.221 & -0.002 & 3.238 & 0.314 & 0.690 \\
Qwen2.5-VL-3B & Dog breed\(^{\dagger}\) & 0.221 & 0.010 & 5.174 & 5.016 & 0.942 \\
Qwen2.5-VL-3B & Tag color & 0.221 & 0.001 & 5.175 & 1.747 & 0.783 \\
Qwen3.5-4B & Child jacket & 0.455 & 0.013 & 6.490 & 0.464 & 0.000 \\
Qwen3.5-4B & Child trousers & 0.455 & -0.047 & -0.246 & 0.026 & 0.000 \\
Qwen3.5-4B & Mirror part & 0.489 & 0.102 & 1.793 & 1.821 & 0.926 \\
Qwen3.5-4B & Mirror frame & 0.489 & 0.096 & 1.529 & 0.731 & 0.854 \\
Qwen3.5-4B & Reflected animal & 0.489 & 0.036 & 0.889 & 0.074 & 0.000 \\
Qwen3.5-4B & Tag shape\(^{\dagger}\) & 0.407 & 0.105 & 6.382 & 0.371 & 0.690 \\
Qwen3.5-4B & Dog breed\(^{\dagger}\) & 0.407 & 0.114 & 4.693 & 4.617 & 0.942 \\
Qwen3.5-4B & Tag color & 0.407 & 0.133 & 6.852 & 1.278 & 0.783 \\
LLaVA-NeXT Llama3-8B & Child jacket & 0.106 & -0.077 & 2.385 & -0.054 & 0.000 \\
LLaVA-NeXT Llama3-8B & Child trousers & 0.106 & 0.159 & 1.592 & -0.217 & 0.000 \\
LLaVA-NeXT Llama3-8B & Mirror part & 0.105 & 0.006 & 2.200 & 2.068 & 0.926 \\
LLaVA-NeXT Llama3-8B & Mirror frame & 0.105 & -0.012 & 0.730 & 1.302 & 0.854 \\
LLaVA-NeXT Llama3-8B & Reflected animal & 0.105 & -0.050 & 0.150 & 0.012 & 0.000 \\
LLaVA-NeXT Llama3-8B & Tag shape\(^{\dagger}\) & 0.111 & 0.067 & 9.099 & 0.221 & 0.690 \\
LLaVA-NeXT Llama3-8B & Dog breed\(^{\dagger}\) & 0.111 & -0.001 & 5.029 & 5.424 & 0.942 \\
LLaVA-NeXT Llama3-8B & Tag color & 0.111 & 0.168 & 9.215 & 1.280 & 0.783 \\
LLaVA-NeXT Mistral-7B & Child jacket & 0.357 & -0.104 & 3.475 & -0.005 & 0.000 \\
LLaVA-NeXT Mistral-7B & Child trousers & 0.357 & 0.479 & -0.188 & -0.251 & 0.000 \\
LLaVA-NeXT Mistral-7B & Mirror part & 0.373 & 0.007 & 1.135 & 1.502 & 0.926 \\
LLaVA-NeXT Mistral-7B & Mirror frame & 0.373 & 0.202 & 0.059 & 0.463 & 0.854 \\
LLaVA-NeXT Mistral-7B & Reflected animal & 0.373 & 0.595 & 1.266 & 0.339 & 0.000 \\
LLaVA-NeXT Mistral-7B & Tag shape\(^{\dagger}\) & 0.385 & 0.010 & 9.467 & 0.777 & 0.690 \\
LLaVA-NeXT Mistral-7B & Dog breed\(^{\dagger}\) & 0.385 & 0.015 & 3.922 & 3.861 & 0.942 \\
LLaVA-NeXT Mistral-7B & Tag color & 0.385 & 0.414 & 7.531 & 0.655 & 0.783 \\
PaliGemma2-3B Mix-448 & Child jacket & 2.341 & -0.032 & 3.445 & -0.057 & 0.000 \\
PaliGemma2-3B Mix-448 & Child trousers & 2.157 & -0.100 & 1.616 & -0.069 & 0.000 \\
PaliGemma2-3B Mix-448 & Mirror part & 2.364 & -0.406 & 1.189 & 1.293 & 0.926 \\
PaliGemma2-3B Mix-448 & Mirror frame & 1.814 & -0.185 & 3.815 & 2.202 & 0.854 \\
PaliGemma2-3B Mix-448 & Reflected animal & 2.538 & -0.361 & 1.808 & -0.014 & 0.000 \\
PaliGemma2-3B Mix-448 & Tag shape\(^{\dagger}\) & 3.040 & 0.127 & 7.655 & 0.503 & 0.690 \\
PaliGemma2-3B Mix-448 & Dog breed\(^{\dagger}\) & 2.118 & 0.144 & 3.980 & 3.795 & 0.942 \\
PaliGemma2-3B Mix-448 & Tag color & 2.523 & 0.370 & 7.654 & 1.055 & 0.783 \\
\end{longtable}
\endgroup
\begingroup
\fontsize{8}{10}\selectfont
\setlength{\tabcolsep}{3pt}
\renewcommand{\arraystretch}{1.00}
\begin{longtable}{@{}p{.22\linewidth}p{.17\linewidth}rrrr@{}}
\caption{Actual deletion budgets. Each entry reports canonical/foil passes. \(n\) counts edited nonzero coordinates across spatial positions; random columns average three groups. ID-count matching does not match coordinate count or perturbation energy.}\label{tab:sae-decoder-budgets}\\
\toprule
Model & Question & \(n_8\) & \(\bar n_{\rm rnd}\) & \(\|\delta h_8\|_2\) & \(\overline{\|\delta h_{\rm rnd}\|_2}\) \\
\midrule
\endfirsthead
\toprule
Model & Question & \(n_8\) & \(\bar n_{\rm rnd}\) & \(\|\delta h_8\|_2\) & \(\overline{\|\delta h_{\rm rnd}\|_2}\) \\
\midrule
\endhead
\midrule
\multicolumn{6}{r}{Continued on next page}\\
\endfoot
\bottomrule
\endlastfoot
Qwen2.5-VL-3B & Child jacket & 207/207 & 14.3/14.3 & 652.70/652.70 & 46.65/46.65 \\
Qwen2.5-VL-3B & Child trousers & 295/295 & 16.7/16.7 & 1767.97/1767.97 & 54.76/54.76 \\
Qwen2.5-VL-3B & Mirror part & 296/296 & 16.0/16.0 & 901.80/901.80 & 41.48/41.48 \\
Qwen2.5-VL-3B & Mirror frame & 204/204 & 19.3/19.3 & 1825.23/1825.23 & 32.23/32.23 \\
Qwen2.5-VL-3B & Reflected animal & 119/119 & 16.7/16.7 & 1022.65/1022.65 & 32.66/32.66 \\
Qwen2.5-VL-3B & Tag shape\(^{\dagger}\) & 282/282 & 14.0/14.0 & 2148.47/2148.47 & 22.29/22.29 \\
Qwen2.5-VL-3B & Dog breed\(^{\dagger}\) & 74/74 & 13.7/13.7 & 220.78/220.78 & 25.55/25.55 \\
Qwen2.5-VL-3B & Tag color & 239/239 & 15.7/15.7 & 935.88/935.88 & 38.56/38.56 \\
Qwen3.5-4B & Child jacket & 170/170 & 26.0/26.0 & 41.15/41.15 & 7.75/7.75 \\
Qwen3.5-4B & Child trousers & 249/249 & 14.0/14.0 & 104.73/104.73 & 4.03/4.03 \\
Qwen3.5-4B & Mirror part & 101/101 & 18.3/18.3 & 36.60/36.60 & 7.60/7.60 \\
Qwen3.5-4B & Mirror frame & 139/139 & 14.0/14.0 & 42.56/42.56 & 7.20/7.20 \\
Qwen3.5-4B & Reflected animal & 187/187 & 18.3/18.3 & 112.95/112.95 & 7.29/7.29 \\
Qwen3.5-4B & Tag shape\(^{\dagger}\) & 203/203 & 21.7/21.7 & 51.81/51.81 & 8.05/8.05 \\
Qwen3.5-4B & Dog breed\(^{\dagger}\) & 56/56 & 23.7/23.7 & 20.37/20.37 & 10.51/10.51 \\
Qwen3.5-4B & Tag color & 130/130 & 20.0/20.0 & 80.71/80.71 & 6.78/6.78 \\
LLaVA-NeXT Llama3-8B & Child jacket & 5588/5588 & 152.7/152.7 & 160.62/160.62 & 5.82/5.82 \\
LLaVA-NeXT Llama3-8B & Child trousers & 9514/9514 & 90.3/90.3 & 268.38/268.38 & 3.76/3.76 \\
LLaVA-NeXT Llama3-8B & Mirror part & 9256/9256 & 71.0/71.0 & 239.33/239.33 & 3.42/3.42 \\
LLaVA-NeXT Llama3-8B & Mirror frame & 8577/8577 & 127.3/127.3 & 270.28/270.28 & 6.01/6.01 \\
LLaVA-NeXT Llama3-8B & Reflected animal & 5790/5790 & 112.0/112.0 & 248.22/248.22 & 4.08/4.08 \\
LLaVA-NeXT Llama3-8B & Tag shape\(^{\dagger}\) & 8087/8087 & 89.7/89.7 & 249.95/249.95 & 2.15/2.15 \\
LLaVA-NeXT Llama3-8B & Dog breed\(^{\dagger}\) & 5709/5709 & 112.0/112.0 & 230.22/230.22 & 5.28/5.28 \\
LLaVA-NeXT Llama3-8B & Tag color & 6860/6860 & 299.0/299.0 & 172.82/172.82 & 8.40/8.40 \\
LLaVA-NeXT Mistral-7B & Child jacket & 1165/1165 & 61.7/61.7 & 107.33/107.33 & 3.94/3.94 \\
LLaVA-NeXT Mistral-7B & Child trousers & 1811/1811 & 47.3/47.3 & 114.37/114.37 & 3.14/3.14 \\
LLaVA-NeXT Mistral-7B & Mirror part & 958/958 & 45.0/45.0 & 34.99/34.99 & 2.43/2.43 \\
LLaVA-NeXT Mistral-7B & Mirror frame & 513/513 & 50.7/50.7 & 19.31/19.31 & 3.93/3.93 \\
LLaVA-NeXT Mistral-7B & Reflected animal & 375/375 & 50.3/50.3 & 50.46/50.46 & 3.64/3.64 \\
LLaVA-NeXT Mistral-7B & Tag shape\(^{\dagger}\) & 923/923 & 54.3/54.3 & 33.65/33.65 & 3.13/3.13 \\
LLaVA-NeXT Mistral-7B & Dog breed\(^{\dagger}\) & 2059/2059 & 110.0/110.0 & 127.82/127.82 & 5.29/5.29 \\
LLaVA-NeXT Mistral-7B & Tag color & 433/433 & 55.3/55.3 & 21.84/21.84 & 3.46/3.46 \\
PaliGemma2-3B Mix-448 & Child jacket & 1731/1731 & 98.7/98.7 & 9448.96/9448.96 & 269.98/269.98 \\
PaliGemma2-3B Mix-448 & Child trousers & 1798/1798 & 100.7/100.7 & 9603.61/9603.61 & 275.84/275.84 \\
PaliGemma2-3B Mix-448 & Mirror part & 1073/1073 & 156.0/156.0 & 4625.68/4625.68 & 399.84/399.84 \\
PaliGemma2-3B Mix-448 & Mirror frame & 1607/1607 & 96.7/96.7 & 8144.84/8144.84 & 256.72/256.72 \\
PaliGemma2-3B Mix-448 & Reflected animal & 1430/1430 & 180.3/180.3 & 6554.28/6554.28 & 363.05/363.05 \\
PaliGemma2-3B Mix-448 & Tag shape\(^{\dagger}\) & 1445/1445 & 87.3/87.3 & 2289.34/2289.34 & 255.29/255.29 \\
PaliGemma2-3B Mix-448 & Dog breed\(^{\dagger}\) & 1137/1137 & 127.7/127.7 & 4885.53/4885.53 & 331.54/331.54 \\
PaliGemma2-3B Mix-448 & Tag color & 2049/2049 & 173.0/173.0 & 9603.87/9603.87 & 372.35/372.35 \\
\end{longtable}
\endgroup

\clearpage
\hypertarget{replaying-figures-and-measurements}{%
\subsection{Replaying figures and
measurements}\label{replaying-figures-and-measurements}}

The archived lens, attention and causal report notebook replays bundled
results and figures without weights or a GPU. The live SAE evaluation
notebooks instead acquire raw inputs, load published checkpoints and compute
new measurements; their smaller teaching subsets are separate from the report
results. The following archive commands regenerate the earlier comparisons;
dependencies include Pandoc, XeLaTeX, and Biber:

\begin{Shaded}
\begin{Highlighting}[]
\ExtensionTok{python}\NormalTok{ examples/render\_crossmodel\_report.py }\KeywordTok{\textbackslash{}}
  \ExtensionTok{{-}{-}source{-}dir}\NormalTok{ docs/report/v10/results {-}{-}output{-}dir docs/report/v11/lens/v2 }\KeywordTok{\textbackslash{}}
  \ExtensionTok{{-}{-}case{-}ids}\NormalTok{ gqa\_07302654 {-}{-}figure{-}types logit{-}lens}
\ExtensionTok{python}\NormalTok{ examples/render\_report\_lenses\_v09.py}
\ExtensionTok{python}\NormalTok{ examples/render\_report\_attention\_v09.py}
\ExtensionTok{python}\NormalTok{ examples/render\_report\_attention\_v09.py }\KeywordTok{\textbackslash{}}
  \ExtensionTok{{-}{-}output}\NormalTok{ docs/report/v11/attention/v2 {-}{-}groups local {-}{-}overlay{-}style contrast}
\ExtensionTok{python}\NormalTok{ examples/render\_report\_causal\_comparison.py }\KeywordTok{\textbackslash{}}
  \ExtensionTok{{-}{-}output{-}dir}\NormalTok{ docs/report/v11/causal/v2}
\ExtensionTok{python}\NormalTok{ examples/render\_report\_vsv.py}
\ExtensionTok{python}\NormalTok{ examples/render\_causal\_report.py}
\ExtensionTok{python}\NormalTok{ examples/run\_report\_path.py {-}{-}render{-}only }\KeywordTok{\textbackslash{}}
  \ExtensionTok{{-}{-}source}\NormalTok{ docs/report/v10/results/path {-}{-}report{-}dir docs/report/v10/causal}
\ExtensionTok{python}\NormalTok{ examples/build\_technical\_report.py}
\end{Highlighting}
\end{Shaded}

The measurement runners handle the shared matrix, attention comparison,
paired residual replacements, and path sweeps. Their filenames and calls
are linked in the report notebook; dedicated method notebooks retain
data acquisition and inference calls. Calibrated artifacts bind model,
tokenizer, activation site, and data provenance. Portable JSON and
figure manifests record source hashes and exported-result equality.
These artifacts support replay and new-input analysis while keeping bulk
checkpoints outside the code repository.

\FloatBarrier

\clearpage
\hypertarget{additional-comparisons}{%
\section{Additional comparisons}\label{additional-comparisons}}

\hypertarget{a-second-vision-classifier-comparison}{%
\subsection{A second vision-classifier comparison}\label{a-second-vision-classifier-comparison}}

\begin{figure}[htbp]
\centering
\includegraphics[width=\linewidth,height=0.64\textheight,keepaspectratio]{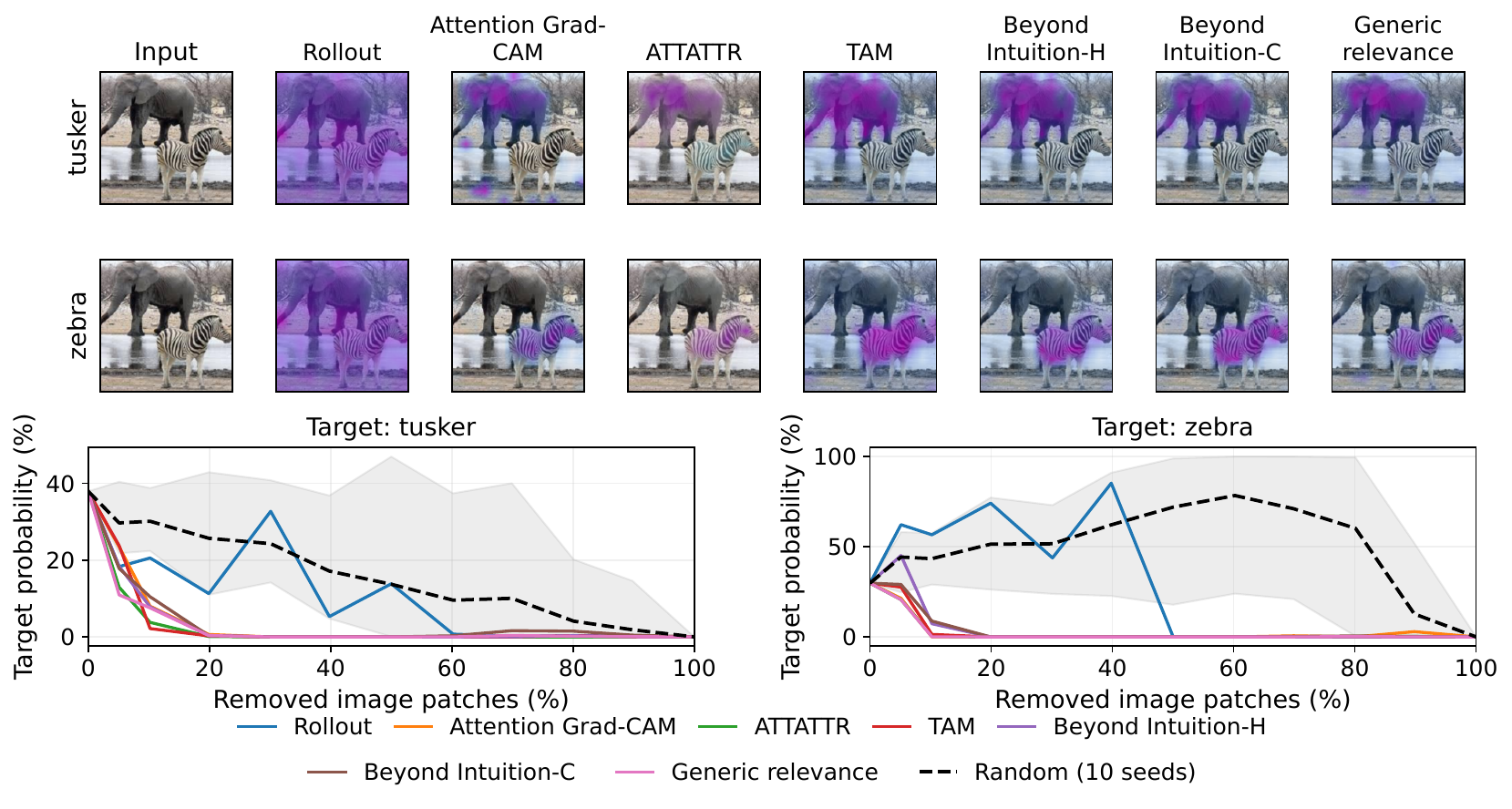}
\caption{A second native ViT control separates elephant and zebra attribution. Identical image preprocessing and deletion budgets allow map differences to be checked against the effect of removing selected patches.}
\label{fig:vit-extra}
\end{figure}

Figure \ref{fig:vit-extra} repeats the native ViT comparison on the
author's elephant--zebra image with the same deletion protocol as the
cat--dog control in Figure \ref{fig:vit}.

\FloatBarrier

\hypertarget{residual-replacement-on-the-black-shirt-input}{%
\subsection{Residual replacement on the black-shirt
input}\label{residual-replacement-on-the-black-shirt-input}}

\begin{figure}[htbp]
\centering
\includegraphics[width=\linewidth,height=0.80\textheight,keepaspectratio]{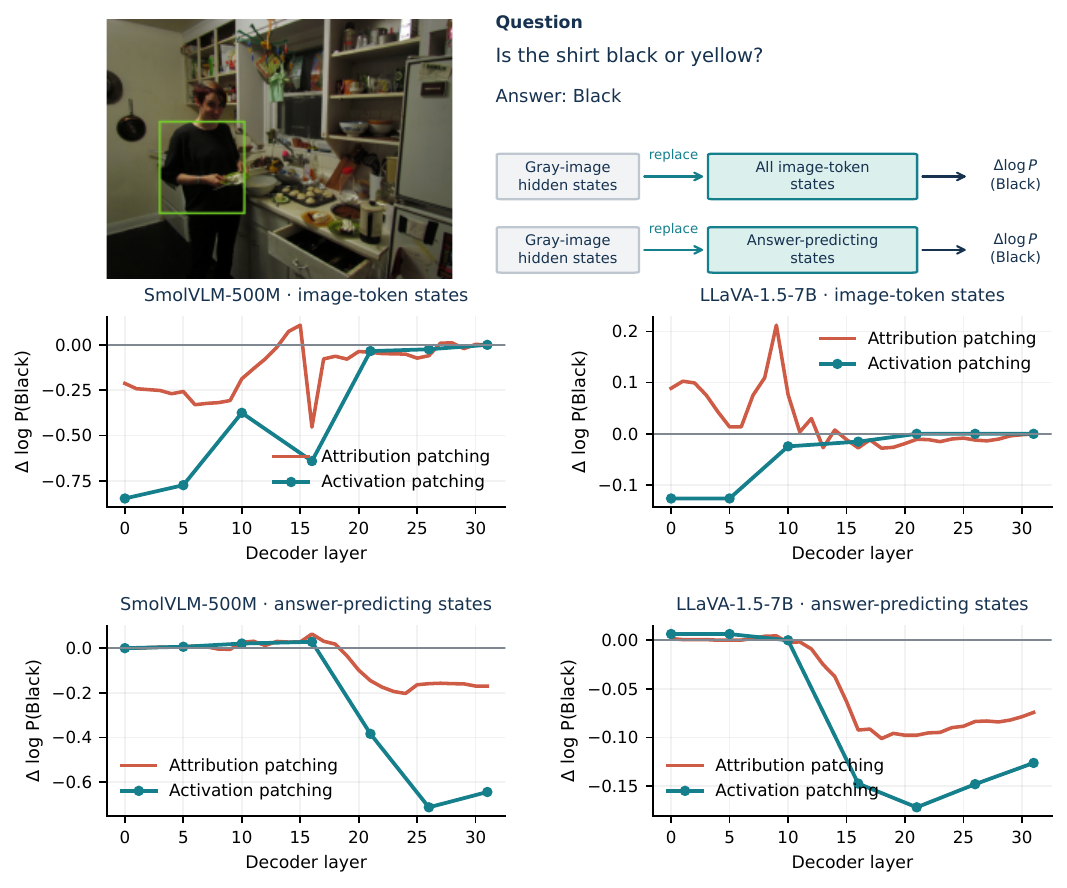}
\caption{Exact and first-order residual replacement for the black-shirt question. LLaVA's measured effects are markedly smaller than on Yellow, whereas SmolVLM remains sensitive to the gray-image reference. Both methods use the same native sites and complete Black score.}
\label{fig:patch-black}
\end{figure}

\begin{figure}[htbp]
\centering
\includegraphics[width=\linewidth]{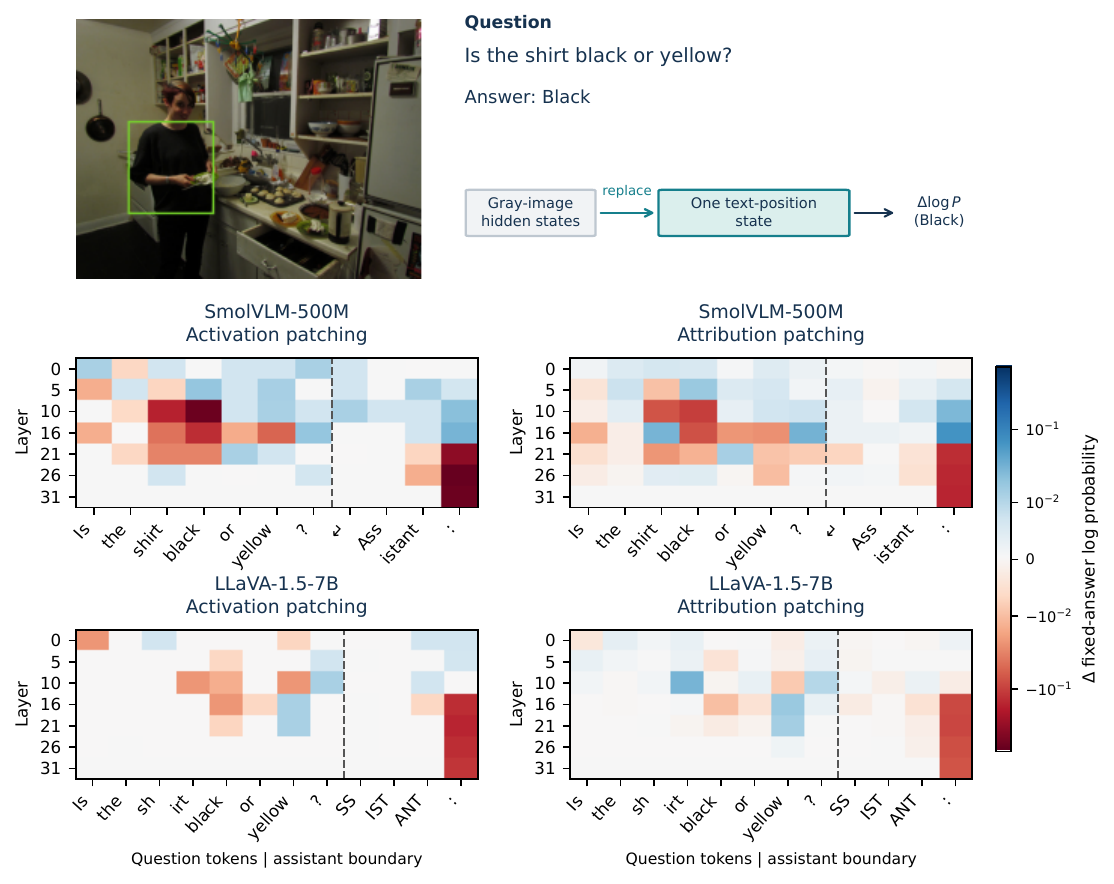}
\caption{Text-position patching for the black-shirt input. Exact replacements identify intermediate question-word and later prediction-position effects; the aligned first-order maps provide a direct approximation check rather than a separate attribution target.}
\label{fig:tokens-black}
\end{figure}

Figures \ref{fig:patch-black} and \ref{fig:tokens-black} retain the
second GQA case under the same reference input, sites, and metrics as
the yellow-shirt study. The smaller LLaVA effect and stronger SmolVLM
question-word sensitivity illustrate why the main comparison is
interpreted per case rather than as a universal pathway ranking.

\FloatBarrier

\end{document}